\documentclass[Afour,sageh,times]{sagej}

\usepackage{moreverb,url}

\usepackage[colorlinks,bookmarksopen,bookmarksnumbered,citecolor=red,urlcolor=red]{hyperref}

\usepackage{multirow}
\usepackage{mathtools}
\usepackage{amsmath} % assumes amsmath package installed
\usepackage{amssymb}  % assumes amsmath package installed
\usepackage{amsthm}
\usepackage{nccmath}
\usepackage{graphicx}
\usepackage{gensymb}
\usepackage{commath}
\usepackage{float}
\usepackage[table,xcdraw]{xcolor}
\usepackage{makecell}
\usepackage{algorithm}
\usepackage{algpseudocode}
\usepackage{bm}
\usepackage{hhline}
\usepackage{booktabs}
\usepackage{lscape}
\usepackage{flushend}

\usepackage{subfloat}

\usepackage[caption=false,font=footnotesize,labelfont=sf,textfont=sf]{subfig}
\DeclareMathOperator*{\argmin}{arg\,min}
\DeclareMathOperator*{\argmax}{arg\,max}

\newtheorem{defn}{Definition}[section]

\newtheorem{prob}{Problem}[section]

\newcommand\BibTeX{{\rmfamily B\kern-.05em \textsc{i\kern-.025em b}\kern-.08em
T\kern-.1667em\lower.7ex\hbox{E}\kern-.125emX}}

\def\volumeyear{2023}
\begin{document}

\runninghead{Lim et al.}

\title{MOB-Net: Limb-modularized Uncertainty Torque Learning of Humanoids for Sensorless External Torque Estimation}

\author{Daegyu Lim\affilnum{1}, Myeong-Ju Kim\affilnum{2}, Junhyeok Cha\affilnum{3} and Jaeheung Park\affilnum{3}\affilnum{4}\affilnum{5}}

\affiliation{\affilnum{1}ROBROS Inc., Seoul, Republic of Korea\\
\affilnum{2}Hyundai Motor Group Robotics Lab, Uiwang, Republic of Korea\\
\affilnum{3}Department of Intelligence and Information, Seoul National University, Seoul, Republic of Korea\\
\affilnum{4}Advanced Institutes of Convergence Technology, Suwon, Republic of Korea\\
\affilnum{5}ASRI, RICS, Seoul National University, Seoul, Republic of Korea}

\corrauth{Jaeheung Park, Dynamic Robotic Systems Lab, Department of Intelligence and Information, Seoul National University, Gwanak-ro 1, Gwanak-gu, Seoul 08826, Republic of Korea.}

\email{park73@snu.ac.kr}

\begin{abstract}
Momentum observer (MOB) can estimate external joint torque without requiring additional sensors, such as force/torque or joint torque sensors. However, the estimation performance of MOB deteriorates due to the model uncertainty which encompasses the modeling errors and the joint friction. Moreover, the estimation error is significant when MOB is applied to high-dimensional floating-base humanoids, which prevents the estimated external joint torque from being used for force control or collision detection in the real humanoid robot.
In this paper, the pure external joint torque estimation method named \emph{MOB-Net}, is proposed for humanoids. 
MOB-Net learns the model uncertainty torque and calibrates the estimated signal of MOB\textcolor{black}{, substantially reducing the estimation errors of MOB}.
The external joint torque can be estimated in the generalized coordinate including whole-body and virtual joints of the floating-base robot with only internal sensors (an IMU on the pelvis and encoders in the joints).
\textcolor{black}{Furthermore, MOB-Net shows more robust performance for the unseen data compared to the end-to-end learning method, and the robustness of MOB-Net is validated through extensive simulations, real robot experiments, and ablation studies.}
Finally, various collision handling scenarios are presented \textcolor{black}{to show the versatility of MOB-Net}: contact wrench feedback control for locomotion, collision detection, and collision reaction for safety. 
\end{abstract}


\maketitle

\section{Introduction}
\label{Section/Introduction}
\begin{figure}[!t]
\centering
\vspace*{0.3cm}
\includegraphics[width=1.0\linewidth]{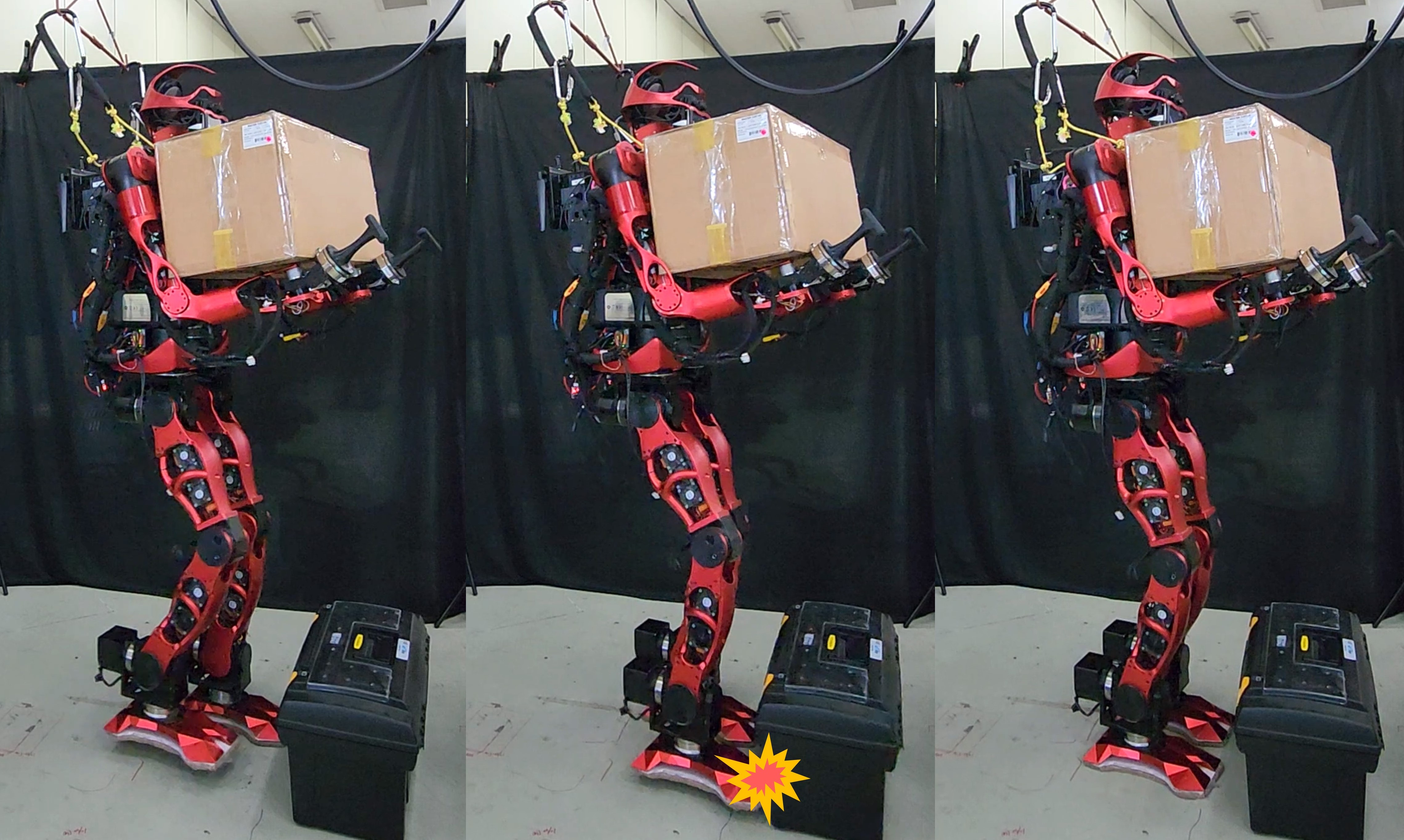}
\caption{\textbf{An example of collision detection on the foot of a humanoid robot with a heavy box due to the occlusion of the camera sensor on the head and collision reaction by back-stepping.} See Scenario \#3 in \href{https://youtu.be/ZpnMEjvGsaQ}{Extension 2} for the video.}
\label{fig/humanoid_collide_with_obstacle_while_carrying_box}
\end{figure}

Safety is the most essential requirement for the collaborative robot that shares the workspace with humans.
To avoid undesirable contact with obstacles or humans, collision avoidance algorithms have been developed using vision sensors. However, it is difficult to prevent all the possible collisions using the vision sensors because blind spots can exist where the sensor can not detect due to occlusions as shown in Figure~\ref{fig/humanoid_collide_with_obstacle_while_carrying_box}.
To detect and cope with unexpected collisions, additional force/torque sensors (FTS) can be equipped on the end-effectors or artificial tactile/skin sensors can be attached to the surface of the robot as in \cite{park2007force,cirillo2015conformable,kobayashi2022whole}.
However, incorporating these additional sensors raises the overall cost of the robot system, making it unaffordable.
Moreover, attaching these sensors increases the inertia of the robot, system complexity, and probability of system failure limiting the allowable number of sensors.  

To cope with unexpected collisions with minimal increase in cost, external torque estimation and collision handling methods using the generalized momentum observer with proprioceptive sensors have also been actively studied in recent decades \cite{de2003actuator, de2005sensorless, de2006collision, de2007acceleration, haddadin2008collision,  haddadin2010new, haddadin2017robot}.
As a result of those studies, not only the collaborative manipulators but also legged robots can estimate the external joint torque and can reliably detect and react to unexpected collisions without exteroceptive sensors. 

In the case of bipedal robots, such as humanoid robots, maintaining the robot's balance is directly related to safety because the robot's base is not fixed, and a fall could cause severe injury to nearby humans. 
With awareness of this problem, many studies have been conducted on the walking controller to enhance the stability and robustness of humanoid robots in \cite{griffin2017walking, joe2018balance, jeong2019robust, khadiv2020walking, daneshmand2021variable,  mesesan2021online, kim2022humanoid, kim2023foot}. 
In these studies, the balancing control performances against external pushes are demonstrated in an open space.
However, such external pushes by an experimenter or released weight can not represent all collisions in the real-world environment. For example, as shown in Figure~\ref{fig/humanoid_collide_with_obstacle_while_carrying_box}, collision sources in the real world would not disappear causing repetitive large impacts to the robot (e.g., heavy box, desk, stair, wall, etc.). Therefore, only with passive balancing controls, humanoid robots could eventually lose balance in these collisions with heavy obstacles.
However, if an active collision detection and reaction system is implemented in humanoids, they would not fall by avoiding further collisions with obstacles. 

Nonetheless, studies on the collision handling capacity of the humanoid robot are spotlighted less compared to the balancing control problem. 
There were several studies on the collision avoidance of quadrupedal robots in \cite{gaertner2021collision, chiu2022collision} and the collision/contact detection of quadrupedal robots in \cite{bledt2018contact, lin2021legged, narukawa2017real, van2022collision}. 
For humanoid robots, there were several studies for external torque estimation for contact detection or contact wrench control using model-based algorithms and FTS in \cite{kaneko2012disturbance, flacco2016residual, vorndamme2017collision, benallegue2018model, lee2018contact, ito2019experimental}, but those methods require FTS and are validated only in the simulation or only for the quasi-static conditions. 
Moreover, to the best of our knowledge, research on the active collision detection and reaction strategy of humanoids while walking has not been conducted yet.
In fact, during DARPA Robotics Challenge in 2015, \cite{krotkov2018darpa}, and ANA Avatar Xprize in 2022, \cite{ackerman2023human}, several humanoid robots fell while walking due to a collision with an obstacle even though advanced balancing controllers were implemented, vision sensors were equipped, and the robot was operated remotely. 
The collision handling method of humanoids can be implemented in parallel with the balancing controller to enhance the safety of the robot in real scenarios.

One of the major reasons for the absence of the collision handling approach (from collision detection to reaction) for humanoids is the low accuracy of the external joint torque estimation. The estimated external joint torque is commonly used as a collision signal for the first step of the collision handling process, i.e., collision detection. 
Unlike the fixed-base manipulator, it is challenging to estimate the external joint torque of humanoids. The floating-base humanoid suffers from modeling errors in the complex system of high degrees of freedom (DoF), large friction torques due to heavy load on the leg, and the sensing noise that occurs from the repetitive collision on the foot while walking. 
Such model uncertainty causes large errors in the external joint torque estimation. 
Due to its significant errors, sensitive collision detection and reaction are difficult for humanoid robots. 
\textcolor{black}{Even with accurate joint torque measurement using a joint torque sensor (JTS), series-elastic-actuator (SEA), or quasi-direct drive (QDD) actuator in the advanced humanoids, dealing with modeling error and sensor noise remains a challenging problem for external joint torque estimation.}

Therefore, in this paper, we propose an uncertainty torque learning method for humanoids to estimate the pure external torque of the whole body without additional sensors. First, the model uncertainty torque is learned using the proposed limb-modularized neural network architecture, \emph{MOB-Net}, which combines a model-based momentum observer (MOB) and a data-driven deep learning method. Then, the estimated external torque from MOB is calibrated using the learned uncertainty torque for accurate external torque estimation. The proposed method is validated through extensive simulations, real robot experiments, and ablation studies. Additionally, various collision handling scenarios using the estimated external joint torque are presented.
\textcolor{black}{Note that, although MOB-Net is applied for a humanoid robot in this work, it can also be utilized for other types of floating-base robots (e.g. mobile manipulators, quadrupedal robots, bipedal robots, different humanoids, etc.).}

The major contributions of this work can be summarized as follows:
\begin{enumerate}
    \item Limb-modularized recurrent neural network, MOB-Net, is proposed for learning the whole-body uncertainty torque of humanoids, and the external joint torque is accurately estimated in real-time thanks to the effective and efficient limb-modularized structure of MOB-Net. The external joint torque is estimated with proprioceptive sensors (joint encoders and an Inertia Measurement Unit (IMU)) for a humanoid while FTS on foot is required for data collection. The estimation performance is significantly improved compared to the model-based methods, contributing to the realization of low-cost humanoids.
    \item Robust estimation performance of the MOB-Net for the unseen data is validated through simulations and real robot experiments compared to the end-to-end learning method (our previous work in \cite{lim2023proprioceptive}) and other model-based methods. 
    Additionally, more sensitive and consistent collision detection performance using both the estimated external joint torque and the estimated standard deviation of the uncertainty torque is presented, contributing to the safety of humanoids.
    \item To the best of our knowledge, we demonstrate the first active collision detection and reaction scenarios for humanoid robots while walking. 
\end{enumerate}

\section{Related Work}
\label{section/Introduction/Related Work}
\subsection{Model Uncertainty Handling in External Torque Estimation}
\cite{de2003actuator} introduced the concept of generalized momentum
and proposed the residual momentum observer for actuator failure detection. Extending from this method, collision detection to reaction methods are developed for safe human-robot interaction in \cite{de2006collision, haddadin2008collision, de2012integrated}. In \cite{haddadin2017robot}, a collision handling pipeline is introduced as a sequential process: collision detection, isolation, identification, classification, and reaction. Collision detection, the first step of the collision handling pipeline, should be performed sensitively to cope with unexpected collision, but a model-based method fundamentally suffers from model uncertainty which deteriorates not only the collision detection performance but also the other collision handling methods following the collision detection.

% filtering
To enhance the collision detection performance with the existence of the uncertainty torque, a band-pass filter is used in \cite{song2013collision}. A band-pass filter suppresses both the high-frequency errors from the sensor noise and the low-frequency errors from the modeling error resulting in more sensitive collision detection.
\cite{caldas2013adaptive, guo2018manipulator} introduced dynamics threshold methods for sensitive collision detection while preventing false positives due to modeling errors.
Although such methods improve collision detection sensitivity, they do not fundamentally improve the estimation performance of external joint torque, and are limited to collision detection. 

% friction modeling
\cite{lee2015sensorless} suggested a friction model that calibrates the residual signal from the momentum observer for more accurate external joint torque estimation, and a more sensitive collision detection performance was shown using the friction model.
\cite{birjandi2020model} proposed an online model adaptation method based on the regressor. 
The regressor-based method calibrates the inertia parameter adaptively and improves the collision detection performance, but requires joint acceleration and joint torque measurement. 
\cite{lee2018geometric} also proposed an inertia parameter identification method formulating the problem in the Riemannian manifold. They validated the accuracy and the robustness of the proposed method for a multi-body system, humanoid.
However, these methods only cover the joint friction or errors of the inertial parameter, which is a part of the model uncertainty.  
Although these methods tried to solve the model uncertainty problem, it is difficult to design an appropriate model to represent the nonlinear friction or an effective algorithm to correct the modeling errors.

% learning method
Conversely, many data-driven collision detection methods have been developed to circumvent the model uncertainty problem and directly infer the collision. Using the deep learning technique, end-to-end collision detection methods showed superior detection performance compared to the model-based method in \cite{heo2019collision, park2020learning, park2021collision, kim2021transferable}. However, the end-to-end collision detection method can not be applied to other tasks such as external force estimation or force control. 

Meanwhile, deep learning is also used for an external wrench or external joint torque estimation which can be applied to various problems: collision detection to reaction, contact force control, and human-robot interaction. \cite{el2018force} proposed an observer based on a recurrent neural network (RNNOB) which infers the pattern of the FTS during contact-free motion and calibrates the sensor measurement. Although the force during free motion is estimated using only joint encoders and IMU, this method still requires FTS for external wrench estimation after the training. \cite{tran2020deep} also trained a multilayer perceptron (MLP) network to directly estimate the external linear force of the surgical robot, da Vinci Research Kit, with joint torques and velocities. A similar approach is presented for the industrial manipulator in \cite{shan2023fine}. They performed the peg-in-whole task to validate the contact wrench estimation performance. \cite{yilmaz2020neural} proposed the residual dynamics torque learning during the free motion for external joint torque estimation on the da Vinci Research Kit, and the estimated external torque is transformed to the external wrench by using Jacobian inverse mapping. We also proposed uncertainty torque learning based on the momentum observer for external torque estimation without additional sensors in \cite{lim2021momentum}. The estimated external torque is used for collision detection in the 2 DoF planar manipulator. However, research on the external torque learning for the legged robot has not been performed except for our previous work in \cite{lim2023proprioceptive}.

\subsection{Collision Handling for Legged Robot}
% collision handling for legged robot
Momentum observer introduced in \cite{de2006collision} is also applied to humanoid robots in various works. \cite{flacco2016residual} presented a momentum observer framework for floating-base humanoids based on the work in \cite{de2006collision} and validated the external torque estimation and contact point estimation performance in simulation. \cite{vorndamme2017collision} utilized the momentum observer for multi-contact collision detection, isolation, and identification of the humanoid robot, and the proposed method is validated and analyzed in simulation. In \cite{benallegue2018model}, the extended Kalman filter-based external wrench estimation algorithm is suggested. The proposed method performs better than their previous sensor-based method in \cite{kaneko2012disturbance} for the hand contact wrench estimation through simulation and the real robot experiment. However, all the aforementioned methods require FTS and the estimation task is only validated in simulation or in the real robot but during quasi-static motion. 

For quadrupedal robots, \cite{hwangbo2016probabilistic} proposed a probabilistic foot contact estimation method combined with the dynamics and kinematics of the robot, and they demonstrated faster contact detection performance compared with the momentum observer-based method.
\cite{benallegue2018model} proposed a discretized version of the momentum observer for more accurate estimation performance, and it is used for foot contact estimation. \cite{van2022collision} also utilized the momentum observer combined with a band-pass filter for collision detection of the legged manipulator. The band-pass filter relieves the modeling error and signal noise simultaneously, but the estimation error increases when the legged robot walks. \cite{barasuol2019detection} proposed a kinematic-based collision detection method and reaction control method for the collision on the shin link using the implemented trunk controller, but it does not produce additional reaction motion at the planning level. Moreover, all model-based methods for the legged robot do not account for the model uncertainty.

Foot contact detection methods based on deep learning are investigated for the humanoid robot in \cite{piperakis2022robust} and the quadrupedal robot in \cite{lin2021legged}. In \cite{piperakis2022robust}, the contact states of each foot are estimated directly using both FTS and IMU on the foot. \cite{lin2021legged} also proposed a deep learning classification method for the contact state estimation of a quadrupedal robot, and the estimated contact states are used for the SLAM algorithm. 
An SVM-based collision detection and reaction method in the crawling humanoid is proposed in \cite{narukawa2017real} even though the collision handling is performed during quadrupedal walking using both arms and legs at a slow speed. 
However, as mentioned in the previous section, the application of end-to-end contact estimation is limited to contact detection.

In summary, there were many works for collision/contact detection and identification problems using model-based methods or data-driven methods. However, existing studies rarely cover the collision reaction problem against unexpected collisions on the leg of the legged robot system while walking. Additionally, model uncertainty is indirectly handled by threshold, signal filtering, and probabilistic model, or ignored. In our approach, the external torque is estimated accurately in the generalized coordinate considering the model uncertainty, and the entire collision handling scenarios of the humanoid robot are presented from collision detection to reaction without FTS.

\subsection{Comparison to our previous work}
This paper is expanded from our previous work in \cite{lim2021momentum} as follows. 1) We expand the external torque estimation method validated in the 2 DoF fixed-based test bed to the whole body of the floating-base humanoid robot with 37 DoF except the head and hand joints.
2) The architecture of the neural network is advanced to train and infer the network efficiently for humanoids. Specifically, a limb-modularized network structure is proposed in this paper using different networks, using different input/output features, and introducing a different data collection method compared to the method in \cite{lim2021momentum}. The general performance of the proposed method for the unseen data is extensively validated through simulations, experiments, and ablation studies. 
3) In this study, the humanoid robot's estimated pure external torque was utilized not only for collision detection but also for various collision reaction scenarios. This extends the work of \cite{lim2021momentum}, which focused solely on collision detection using estimated external joint torque in a 2 DoF test bed.

\section{Problem Formulation}
\label{Section/Problem Formulation}

\subsection{Rigid Body Dynamics of Floating-base Robot}
\label{Subsection/Problem Formulation/Rigid Body Dynamics of Floating-base Robot}
A humanoid can be described as a floating-base multi-body system that has $n+1$ rigid bodies and $n$ joints. In the case of a floating-base robot, the robot's dynamics can be expressed by linking six virtual joints to the base frame. Thus, the rigid body dynamics for an $n+6$ DoF system with six virtual joints is outlined as follows:
\begin{align}
\label{equation/dynamics equation}
    \bm{M}(\bm{q}_{v})\ddot{\bm{q}}_{v}+\bm{C}(\bm{q}_{v},\dot{{q}}_{v})\dot{\bm{q}}_{v} + \bm{g}(\bm{q}_{v}) & = \bm{\tau}_{v}+\bm{\tau}_{e}, \\
\label{equation/external torque ft}
    \bm{\tau}_{e} &= \sum_{i=1}^{k} \bm{J}^{T}_{c,i}\bm{F}_{e,i},
\end{align}
where $ \bm{M}(\bm{q}_{v}), \bm{C}(\bm{q}_{v},\dot{\bm{q}}_{v})\in \mathbb{R}^{(n+6)\times (n+6)}$, and $\bm{g}(\bm{q}_{v})\in \mathbb{R}^{n+6}$ are the inertia matrix, the Coriolis and centrifugal matrix, and the gravity vector, respectively. $\bm{q}_v, \dot{\bm{q}}_{v}, \ddot{\bm{q}}_{v}\in \mathbb{R}^{n+6}$ are the generalized position, velocity, and acceleration vectors including virtual joints, respectively. $\bm{\tau}_{v}\in \mathbb{R}^{n+6}$ is the control torque and $\bm{\tau}_{e}\in \mathbb{R}^{n+6}$ is the external torque. The external torque occurs by the external wrench in the operational space as represented in (\ref{equation/external torque ft}) where $k$, $\bm{J}_{c,i}\in \mathbb{R}^{6\times (n+6)}$, and $\bm{F}_{e,i}\in \mathbb{R}^{6}$ are the total number of contacts, the $i$-th contact Jacobian matrix, and the $i$-th external wrench, respectively. The generalized coordinate vector consists of six virtual joints, and $n$ motor angles: $\bm{q}_{v} = [\bm{x}_{fb}^{T} \ \bm{\theta}_{fb}^{T} \ \bm{q}^{T}]^{T}$, $\dot{\bm{q}}_{v} = [\bm{v}_{fb}^{T} \ \bm{\omega}_{fb}^{T} \ \dot{\bm{q}}^{T}]^{T}$, $\ddot{\bm{q}}_{v} = [\dot{\bm{v}}_{fb}^{T} \ \dot{\bm{\omega}}_{fb}^{T} \ {{\ddot{\bm{q}}}}^{T}]^{T}$. $\bm{x}_{fb} \in \mathbb{R}^{3}$ and $\bm{\theta}_{fb}\in \mathbb{R}^{3}$ are the position vector and the Euler angles of the floating-base. The orientation of the floating base can also be expressed with the orientation matrix $\bm{R}_{fb}^{T}\in SO(3)$. $\bm{v}_{fb}\in \mathbb{R}^{3}$ and $\bm{\omega}_{fb}\in \mathbb{R}^{3}$ are the linear and angular velocity of the floating-base. The underactuated floating-base robot has zero input torque for the virtual joints, and motor torque $\bm{\tau}_{m}\in \mathbb{R}^{n}$ for the actuating joints as $\bm{\tau}_{v} = [\bm{0}^{T} \  \bm{\tau}_{m}^{T}]^{T}$.

\subsection{Momentum Observer (MOB)} % can be removed
\label{Subsection/Problem Formulation/Momentum Observer}
The external torque on the robot can be obtained by mapping the contact force ${F}_{e,c}$ measured by FTS to the joint torque as described in (\ref{equation/external torque ft}). However, in the absence of FTS in the robot, the external torque can be estimated from only internal sensors including the joint encoder, motor current sensor, and IMU. 

Using the robot's dynamics in (\ref{equation/dynamics equation}) and the nominal model, external torque can be calculated by rearranging the equation as below
\begin{equation}
\label{equation/external torque}
    \bm{\tau}_{e} = {\overline{\bm{M}}}({q}_{v})\ddot{\bm{q}}_{v}+{\overline{\bm{C}}}(\bm{q}_{v},\dot{\bm{q}}_{v})\dot{\bm{q}}_{v} + {\overline{\bm{g}}}(\bm{q}_{v}) - \bm{\tau}_{v}
\end{equation}
where $\overline{(\cdot)}$ indicates the nominal model obtained from the estimated model parameters.
Assuming ideal system identification, $\bm{\tau}_{e}$ can be obtained only with internal sensor information $\bm{q}_{v}$, $\dot{\bm{q}}_{v}$, $\ddot{\bm{q}}_{v}$ and $\bm{\tau}_{m}$.

\label{Subsection/Preliminaries/Momentum Observer}
However, the joint acceleration $\ddot{\bm{q}}$ in (\ref{equation/external torque}) is highly noisy because it is calculated by the numerical differentiation. To circumvent this problem, a conventional model-based method, momentum observer was introduced in \cite{de2003actuator} which does not use joint acceleration for external torque estimation. According to \cite{de2003actuator}, the generalized momentum of the floating-base robot is defined as
\begin{equation}
\label{equation/generalized momentum}
    \bm{p} = {\overline{\bm{M}}(\bm{q}_{v})\dot{\bm{q}}_{v}}.
\end{equation}
The derivative of $\bm{p}$ can be expressed by using floating-base dynamics (\ref{equation/dynamics equation}) with a nominal model and well known relation $\dot{{\overline{\bm{M}}}} = {\overline{\bm{C}}} + {\overline{\bm{C}}}^{T}$ as
\begin{align}
        \label{derivative of generalized momentum}
        {\dot{\bm{p}}} &=  \bm{\tau}_{v} + \bm{\tau}_{e} +{\overline{\bm{C}}}^{T}(\bm{q}_{v},\dot{\bm{q}}_{v})\dot{\bm{q}}_{v}-{\overline{\bm{g}}}(\bm{q}_{v})       \\
        \label{derivative of generalized momentum beta}
        &= \bm{\tau}_{v} + \bm{\tau}_{e} +{\overline{\bm{\beta}}}(\bm{q}_{v}, \dot{\bm{q}}_{v}), 
\end{align}
where ${\overline{\bm{\beta}}}(\bm{q}_{v}, \dot{\bm{q}}_{v}) =  {{\overline{\bm{C}}}}^{T}(\bm{q}_{v}, \dot{\bm{q}}_{v})\dot{\bm{q}}_{v}-{{\overline{\bm{g}}}}(\bm{q})$ is defined for the readability.
To estimate external torque, a residual vector $\bm{r}\in \mathbb{R}^{n+6}$ and its dynamics are defined as follows:
\begin{align}
        \label{derivative of residue}
        \dot{\bm{r}} &= \bm{K}_0(\dot{\bm{p}} - \dot{\hat{\bm{p}}}),
        \\
        \label{derivative of estimated momentum}
        {\dot{\hat{\bm{p}}}} &= \bm{\tau}_{v}+{\overline{\bm{\beta}}}(\bm{q}_{v},\dot{\bm{q}}_{v})+\bm{r},
\end{align}
where $\bm{K}_0 = diag\{k_{0,i}\}$ is the positive diagonal matrix and $\dot{\hat{\bm{p}}}$ is the derivative of the estimated momentum. 
Integrating (\ref{derivative of residue}) results in 
\begin{equation}
\label{equation/residue integration}
    \bm{r} = \bm{K}_0\left\{\bm{p}(t) - \bm{p}(0) - \int_{0}^{t}(\bm{\tau}_{v}+{\overline{\bm{\beta}}}(\bm{q}_{v}, \dot{\bm{q}}_{v})+\bm{r})dt \right\}.
\end{equation}
 Substituting (\ref{derivative of generalized momentum beta}) and (\ref{derivative of estimated momentum}) to (\ref{derivative of residue}) derives the relation between $\bm{\tau}_{e}$ and $\bm{r}$ as
\begin{equation}
\label{equation/residue dynamics}
    \dot{\bm{r}} =\bm{K}_0(\bm{\tau}_{e}-\bm{r}).
\end{equation}
Once $\bm{K}_0$ is a diagonal matrix, the residual dynamics (\ref{equation/residue dynamics}) in each joint is independent and using Laplace transformation for each joint leads to
\begin{equation}
\label{equation/residual lpf}
    \frac{r_{i}(s)}{\tau_{e,i}(s)} = \frac{k_{0,i}}{s+k_{0,i}} \ \ (i = 1, 2, \dots, n+6).
\end{equation}
This means that the MOB residual vector, $\bm{r}$, is the first-order low-pass filtered signal of the pure external torque, ${\tilde{\bm{\tau}}}_{e}$. 
% \textcolor{black}{Notably, the robot equipped with JTS or SEA also requires MOB to estimate joint external torque using the measured joint torque $\bm{\tau}_j$ instead of motor current torque $\bm{\tau}_m$ (\cite{de2006collision, kim2015design, lee2017two}).}

\subsection{Uncertainty Torque}
\label{Subsection/Problem Formulation/Uncertainty Torque}
Despite the theoretical guarantee of exponential convergence in (\ref{equation/residue dynamics}), the residual torque is prone to have large errors in practice. The estimation error of the momentum observer results from the unmodeled dynamics torques which are not considered in the nominal dynamics or sensor noise. We formally define the error of the nominal model in the torque level as an uncertainty torque.

\begin{defn}
\label{definition/uncertainty torque}
(Uncertainty Torque):\\
Given a nominal model and its dynamics as (\ref{equation/external torque}), the difference between the dynamics torque calculated using the nominal model (${\overline{\bm{M}}}(\bm{q}_{v})\ddot{\bm{q}}_{v}+{\overline{\bm{C}}}(\bm{q}_{v},\dot{\bm{q}}_{v})\dot{\bm{q}}_{v} + {\overline{\bm{g}}}(\bm{q}_{v})$) and the sum of the control torque and the external torque ($\bm{\tau}_{v}$ + $\bm{\tau}_{e}$) is {uncertainty torque} ($\bm{\tau}_{u}$).
\begin{equation}
    \label{equation/uncertainty torque}
    \bm{\tau}_{u} = {\overline{\bm{M}}}(\bm{q}_{v})\ddot{\bm{q}}_{v}+{\overline{\bm{C}}}(\bm{q}_{v},\dot{\bm{q}}_{v})\dot{\bm{q}}_{v} + {\overline{\bm{g}}}(\bm{q}_{v}) - \bm{\tau}_{v} - \bm{\tau}_{e}. 
\end{equation} 
\end{defn}

Then, the major sources of the uncertainty torque are categorized into three factors in this paper; modeling error, friction torque, and sensor noise. 

In practice, it is difficult to obtain the accurate inertia parameters of the robot $\bm{M}(\bm{q}_{v}), \bm{C}(\bm{q}_{v},\dot{\bm{q}}_{v})$, and $\bm{g}(\bm{q}_{v})$ in (\ref{equation/dynamics equation}). Instead, the nominal model is commonly obtained from a CAD model or estimated by a system identification procedure. Then, the dynamics of the floating-base robot with nominal model parameters is expressed below
\begin{equation}
\label{equation/dynamics equation with modeling error}
    {\overline{\bm{M}}}(\bm{q}_{v})\ddot{\bm{q}}_{v}+{\overline{\bm{C}}}(\bm{q}_{v},\dot{\bm{q}}_{v})\dot{\bm{q}}_{v} + {\overline{\bm{g}}}(\bm{q}_{v}) = \bm{\tau}_{v}+\bm{\tau}_{e}+\bm{\tau}_{p}.
\end{equation}
In (\ref{equation/dynamics equation with modeling error}), $\bm{\tau}_{p} = ({\overline{\bm{M}}}(\bm{q}_{v})\ddot{\bm{q}}_{v}+{\overline{\bm{C}}}(\bm{q}_{v},\dot{\bm{q}}_{v})\dot{\bm{q}}_{v} + {\overline{\bm{g}}}(\bm{q}_{v})) - (\bm{{M}}(\bm{q}_{v})\ddot{\bm{q}}_{v}+{\bm{C}}(\bm{q}_{v},\dot{\bm{q}}_{v})\dot{\bm{q}}_{v} + {\bm{g}}(\bm{q}_{v})) \in \mathbb{R}^{n+6}$ is the modeling error torque induced by the error of the parameters between the nominal model and the real robot. Therefore, because the nominal model is used for the momentum observer in (\ref{equation/residue integration}), the residual torque contains modeling error torque as well as the external torque.

Various kinds of joint friction models have been introduced to compensate for the friction including coulomb, static, viscous, and load-dependent frictions as introduced in \cite{xiao2018collision, lee2015sensorless, abeykoon2014position, ma2018investigation, quiroga2021load}.
Moreover, it is reported that the friction torque shows hysteresis in \cite{kircanski1997experimental, ruderman2015sensorless}.
In this study, we consider the joint friction torque as a function of joint position, joint velocity, and joint torque, $\bm{\tau}_{f}(\bm{q}_{v}, {\dot{\bm{q}}_{v}}, \bm{\tau}_{v})$.

Although all kinds of sensors have noise in their measurement, such noise can be modeled and interpreted using a probabilistic model and one of the most common models for sensor noise is the Gaussian model. 
Therefore, a lumped disturbance torque induced by the sensor noise, ${\tau}_{n}$, can be added to the disturbance. 
$\bm{\tau}_{n}(\bm{q}_{v}, \dot{\bm{q}}_{v}, \ddot{\bm{q}}_{v}) \sim P(\bm{\tau}_{n}|\bm{q}_{v}^{m}, \dot{\bm{q}}_{v}^{m}, \ddot{\bm{q}}_{v}^{m}, \bm{\sigma}^2_{s})$ follows the conditional probability distribution and $\bm{\sigma}^2_{s}$ is the variance of all the sensor noise.

By summing up the three components of the uncertainty torque, the uncertainty torque in (\ref{equation/uncertainty torque}) can be expressed as a stochastic function of joint position, velocity, acceleration, and joint torque:
\begin{equation}
\begin{split}
\label{equation/uncertainty torque2}
    \bm{\tau}_{u}(\bm{q}_{v}, \dot{\bm{q}}_{v}, \ddot{\bm{q}}_{v}, \bm{\tau}_{v}) = ~ &\bm{\tau}_{p}(\bm{q}_{v}, \dot{\bm{q}}_{v}, \ddot{\bm{q}}_{v})+\bm{\tau}_{f}(\bm{q}_{v}, {\dot{\bm{q}}_{v}}, \bm{\tau}_{v}) \\
    &+\bm{\tau}_{n}(\bm{q}_{v}, \dot{\bm{q}}_{v}, \ddot{\bm{q}}_{v}).
\end{split}
\end{equation}

\begin{figure*}[!ht]
\centering
\includegraphics[width=1.0\linewidth]{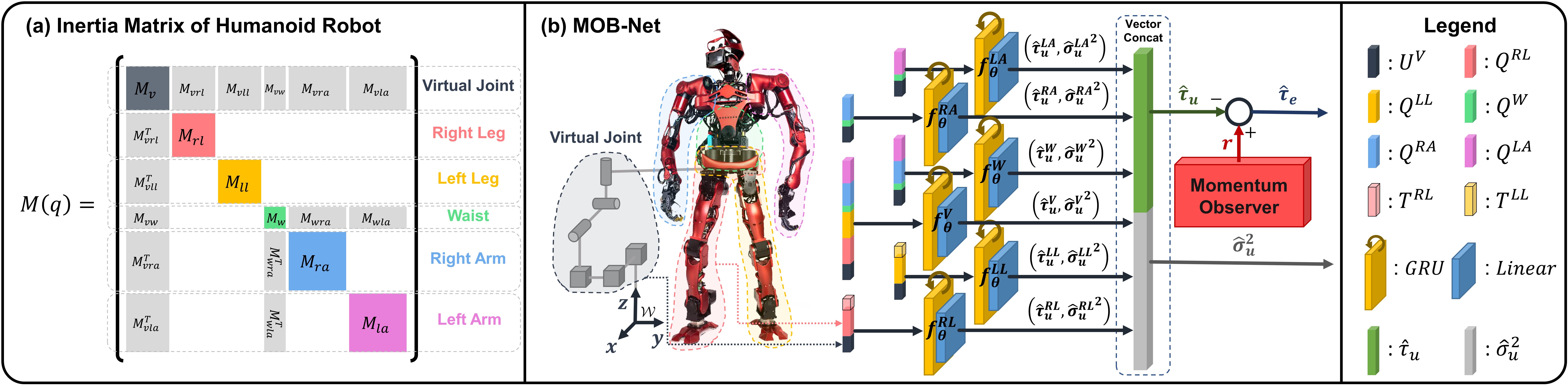}
\caption{\textbf{Diagram of the proposed method.} (a) Structure of the inertia matrix of a humanoid robot. The empty region has zero elements. (b) The architecture of the proposed MOB-Net. Each color bar represents the input data from the specific limb.}
\label{fig/M_matrix_MOB_Net}
\end{figure*}

\subsection{Problem Definition}
\label{Subsection/Problem Formulation/Problem Definition}
In this paper, the fundamental goal is to estimate the pure external joint torques of the floating-base robot using only internal sensors for the realization of both safe and low-cost robots. To achieve this goal, MOB is utilized as a baseline algorithm. However, as mentioned in the previous section, the estimated residual vector of the MOB contains not only the low-pass filtered external torque but also the low-pass filtered uncertainty torque: 
\begin{equation}
\label{equation/residual equal external torque and uncertainty torque}
    \bm{r} = \tilde{\bm{\tau}}_{e} + \tilde{\bm{\tau}}_{u}.
\end{equation}

Therefore, the core problem is defined as an estimation of the delayed uncertainty torque ($\hat{\bm{\tau}}_{u}$). After that, the delayed pure external torque ($\tilde{\bm{\tau}}_{e}$) can be obtained by subtracting $\hat{\bm{\tau}}_{u}$ from the residual $\bm{r}$ in (\ref{equation/residual equal external torque and uncertainty torque}) as $\tilde{\bm{\tau}}_{e} = \bm{r} - \hat{\bm{\tau}}_{u}$. According to the previous section, the uncertainty torque is a stochastic function conditioned by the internal states $\bm{X}_p$. An assumption is made to simplify the complex problem that the conditional probability of the uncertainty torque follows the Gaussian distribution independently for each joint as ${\tau}_{u,j}(\bm{X}_p) \sim \mathcal{N}(\hat{\tau}_{u,j}, \hat{\sigma}_{u,j}^2 | \bm{X}_p)$. Finally, the core problem is formulated as obtaining the optimal normal distribution of the uncertainty torque from proprioceptive sensing.

\begin{prob}
\label{problem/uncertainty torque estimation}
Model Uncertainty Torque Estimation:
\begin{equation}
\label{equation/uncertainty torque estimation}
\hat{\bm{\tau}}_{u}^*(\bm{X}_p), \hat{\bm{\sigma}}_{u}^{2*}(\bm{X}_p) = \argmax_{\hat{\bm{\tau}}_{u}, \hat{\bm{\sigma}}_{u}^2} \ {P}(\bm{\tau}_{u}|\hat{\bm{\tau}}_{u}, \hat{\bm{\sigma}}_{u},\bm{X}_p).
\end{equation}
{Find the optimal mean and variance functions of the normal distribution, maximizing the conditional probability of the true uncertainty torque given the observation $\bm{X}_p$. $\bm{X}_p$ is a set of internal states of the robot obtained from the proprioceptive sensors.}
\end{prob}

%%%%%%%%%%%%%%%%%% MOB-Net %%%%%%%%%%%%%%%%%%%%%%%%%%%%
\section{Model Uncertainty Torque Learning for\\Pure External Torque Estimation}
\label{Section/Proposed Method}
In this section, a data-driven method called \emph{MOB-Net} is presented which is the deep learning method for model uncertainty torque estimation to solve \textbf{Problem \ref{problem/uncertainty torque estimation}} and, for pure external torque estimation. Deep learning is powerful for nonlinear regression problems if appropriate input and output features and training data can be obtained. Based on the momentum observer in Section \ref{Subsection/Problem Formulation/Momentum Observer}, dynamics of the humanoid robot, and analysis of the uncertainty torque in Section \ref{Subsection/Problem Formulation/Uncertainty Torque}, neural networks are designed to estimate the uncertainty torque with only internal information. Details on the network architecture, data collection, and training method are introduced.

% input features for each limb

% target value

\subsection{MOB-Net Architecture}
\label{Subsection/Proposed Method/Network Architecture}

% MOB-Net structure summary
\begin{table*}[!t]
\centering
\caption{Summary of the network architecture of MOB-Net.}
\label{Table/Network Architecture}
\resizebox{\textwidth}{!}{%
\begin{tabular}{@{}cccccc@{}}
\toprule
\textbf{Network} &
  \textbf{\begin{tabular}[c]{@{}c@{}}Input State \\ ($\bm{x}^{g}$)\end{tabular}} &
  \textbf{\begin{tabular}[c]{@{}c@{}}Target Value \\ ($\bm{\tau}_{u}$)\end{tabular}} &
  \textbf{Network Architecture} &
  \textbf{Network Size} &
  \textbf{\begin{tabular}[c]{@{}c@{}}Time Horizon of TBPTT \\ ($h$)\end{tabular}} \\ \midrule
$\bm{f}_{\theta}^{RL}$  & ($\bm{Q}^{RL},\bm{T}^{RL},\bm{U}^{V}$)             & $\bm{r}^{RL}-\bm{\tau}_e^{RL}$ & GRU[30, 150]-Linear[150, 12] & 83712  & 50 steps \\
$\bm{f}_{\theta}^{LL}$ & ($\bm{Q}^{LL},\bm{T}^{LL},\bm{U}^{V}$)              & $\bm{r}^{LL}-\bm{\tau}_e^{LL}$ & GRU[30, 150]-Linear[150, 12] & 83712  & 50 steps \\
$\bm{f}_{\theta}^{RA}$  & ($\bm{Q}^{RA},\bm{Q}^{W},\bm{U}^{V}$)       & $\bm{r}^{RA}$        & GRU[34, 200]-Linear[200, 16] & 144816 & 50 steps \\
$\bm{f}_{\theta}^{LA}$ & ($\bm{Q}^{LA},\bm{Q}^{W},\bm{U}^{V}$)       & $\bm{r}^{LA}$        & GRU[34, 200]-Linear[200, 16] & 144816 & 50 steps \\
$\bm{f}_{\theta}^{W}$     & ($\bm{Q}^{RA},\bm{Q}^{LA},\bm{Q}^{W},\bm{U}^{V}$)    & $\bm{r}^{W}$     & GRU[50, 200]-Linear[200, 6]  & 152406 & 50 steps \\
$\bm{f}_{\theta}^{V}$ &
  ($\bm{Q}^{RL},\bm{Q}^{LL},\bm{Q}^{RA},\bm{Q}^{LA},\bm{Q}^{W},\bm{U}^{V}$) &
  $\bm{r}^{V}-\bm{\tau}_e^{V}$ &
  GRU[74, 200]-Linear[200, 12] &
  168012 &
  100 steps \\ \bottomrule
\end{tabular}%
}
\end{table*}

To design the architecture of MOB-Net, the dynamics of the humanoid robot is analyzed. 
According to \cite{featherstone2014rigid}, the tree structure of the humanoid robot causes sparsity in the robot's dynamics. Therefore, when the virtual joints are attached to the pelvis link of the humanoid robot, the inertia matrix of the robot has a sparse structure as shown in Figure~\ref{fig/M_matrix_MOB_Net}~(a). In the inertia matrix, the blank white area has zero components and gray blocks represent the coupled dynamics. The inertia matrix is only presented for visualizing the dynamics as it is core for analysis of dynamics and $C(q, \dot{q})$ matrix has the same structure as the inertia matrix. As the inertia matrix represents, the robot's dynamics can be divided into 6 groups: \emph{Virtual Joint} ($V$), \emph{Right Leg} ($RL$), \emph{Left Leg} ($LL$), \emph{Waist} ($W$), \emph{Right Arm} ($RA$), and \emph{Left Arm} ($LA$). Note that the neck joints are ignored for simplification in this work. 
When it comes to the dynamics of the left leg, for example, only the states of the virtual joint and the left leg's joint are involved in the dynamics of the left leg. This relationship is utilized for uncertainty torque learning. Therefore, only relevant states are provided to the network for training the uncertainty torque of the specific limb, which enhances the learning performance. This dynamics analysis is leveraged as an inductive bias for an efficient and effective learning framework resulting in six independent networks. 

As shown in Figure~\ref{fig/M_matrix_MOB_Net}~(b), MOB-Net consists of a momentum observer and six Gated Recurrent Unit (GRU)-Linear networks ($\bm{f}_{\theta}^{g}(\bm{X}_p), \ g\in\{V, RL, LL, W, RA, LA\}$) that estimate uncertainty torque for each limb group.
Each GRU-Linear network receives a relevant input vector from the robot's proprioceptive sensors for each limb's uncertainty torque estimation and infers the uncertainty torque and its variance. To ensure that the estimated variance is always positive, the softplus activation function is applied for the output vectors of the variance. Then, the output vectors are concatenated to a single whole-body uncertainty torque and its variance. Finally, the residual vector of the momentum observer ($\bm{r}$) is subtracted from the estimated uncertainty torque ($\hat{\bm{\tau}}_{u}$) leaving pure external torque ($\hat{\bm{\tau}}_{e}$). 

The reason for using the Recurrent Neural Network (RNN) structure with a GRU module is that RNN is specialized for handling time-sequence data inherently. 
The friction torque shows hysteresis requiring history information in addition to the current states for uncertainty torque estimation and RNN is suitable. 
Moreover, RNN enables to avoid using joint acceleration which is noisy. More details about input features are discussed in Section \ref{Subsection/Proposed Method/Input and Output Features}. 
The superiority of GRU compared with the other kinds of networks is presented in Section \ref{Subsection/Ablation Study/Network Structure Comparison}.

Therefore, limb-group modularized GRU is adopted as a key architecture of MOB-Net. 
Thanks to its compact size, not only the high estimation performance but also the real-time calculation of 6 networks on the embedded computer is possible within 1\,ms using a single core (Intel Core i7-10700K). \textcolor{black}{Considering our system's complexity and computational power, it is also possible for the other systems to calculate MOB-Net online.}

\subsection{Input Feature and Target Value}
\label{Subsection/Proposed Method/Input and Output Features}
The selection of the input feature ($\bm{X}_{p}$) and the target value of the uncertainty torque ($\bm{\tau}_{u}$) affect the way of data collection, the architecture of the network, and most importantly the learning performance. Therefore, the input feature and the target value should be determined carefully considering various aspects. We design the input feature and the target value of the whole body of the humanoid robot by categorizing the limb groups of a humanoid robot into legs (RL, LL), upper body (W, RA, LA), and virtual joint (V).

% Q^{g}
The input vector is defined as a sequence of the selected proprioceptive states as below:
\begin{align*}
    \bm{X}_{p}^{g} &= [\bm{x}^{g}(k-h+1), \ \dots \ , \ \bm{x}^{g}(k-1), \ \bm{x}^{g}(k)],
\end{align*} 
where $k$ is the current discrete control time, and $h$ is the length of the data sequence horizon. 
$(\cdot)^{g}$ indicates the partial vector of the joints in group $g$.
The proprioceptive input states, $\bm{x}^{g}(k)$, are composed of a combination of three sensor vectors of each limb group $g$: 
\begin{equation}
\label{equation/sensor vectors}
\begin{split}
    \bm{Q}^{g}(k) &= [\bm{q}^{g}(k),\ \dot{\bm{q}}^{g}(k)], \\
    \bm{T}^{g}(k) &= [\bm{\tau}_{d}^{g}(k)], \\
    \bm{U}^{V}(k) &= [{}^{p}\bm{R}_{fb,1:6}(k),\ {}^{p}\bm{\omega}_{fb}(k),\ {}^{p}\dot{\bm{v}}_{fb}(k)],
\end{split}
\end{equation}
where $\bm{Q}^{g}(k)$ is the concatenation of joint states, $\bm{T}^{g}(k)$ is the desired torque vector, and $\bm{U}^{V}(k)$ is the vector of IMU measurement. $\bm{\tau}_{d}^{g}(k)$ is the desired joint torque belonging to the limb group $g$. ${}^{p}\bm{R}_{fb,1:6}(k) \in \mathbb{R}^{6}$ is the first two columns of the rotation matrix of the pelvis. ${}^{p}\bm{\omega}_{fb}(k)$ and ${}^{p}\dot{\bm{v}}_{fb}(k)$ are the pelvis angular velocity and the pelvis linear acceleration in the pelvis local frame.
The specific input state and the target value for each network $\bm{f}_{\theta}^{g}$ is summarized in Table \ref{Table/Network Architecture}.
The reason for the selection of the input state and the target value of MOB-Net is addressed in the three considerations: {the quality of the available information}, characteristics of the {potential target task}, and {the cost of the data collection}.

\subsubsection{Quality of Information}
\label{Subsubsection/Proposed Method/Input and Output Features/Quality of Information}
Although all the required input features ($\bm{q}_{v}, \dot{\bm{q}}_{v}, \ddot{\bm{q}}_{v}, \bm{\tau}_{m}$) of uncertainty torque derived in (\ref{equation/uncertainty torque2}) can be measured or estimated from the internal sensors (encoder, IMU, and motor current sensor), joint acceleration is too noisy to use from the numerical differentiation. Also, measuring the motor torque requires a current sensor and it is a time-delayed signal compared to the command torque. Therefore, not all input features from the dynamics analysis are provided to the network. The sequence of the partial information alone is used as input data based on several assumptions, which is also used in our previous work, \cite{lim2023proprioceptive}:
\begin{itemize}
     \item Joint acceleration is redundant and can be estimated from the time sequence of joint velocity.
     \item Angular acceleration and linear velocity of the pelvis can be estimated with the time sequence of joint position, joint velocity, and IMU data (orientation of pelvis, pelvis angular velocity, and pelvis linear acceleration), \cite{rotella2014state}.
     \item Motor torque measured from the current sensor ${\tau}_{m}$ can be replaced with the desired torque ${\tau}_{d}$ because it has a reasonably small error to the measured torque.
 \end{itemize}
Therefore, only three sensor vectors in (\ref{equation/sensor vectors}) are used as input features.
\textcolor{black}{Note that the orientation is explicitly included in the input feature because a high-performance IMU, commonly used in legged robots, provides orientation information, negating the need for researchers to implement a Kalman filter for estimation.}
\subsubsection{Potential Target Task}
\label{Subsubsection/Proposed Method/Input and Output Features/Interesting Task}
The expected target tasks for the legs of the humanoid robot are to support the entire weight of the robot on one or two legs and walk to the target destination making repetitive impacts with the ground. During these tasks, the legs are loaded with large forces ($>1000$\,N for our robot) making the load-dependent dynamics dominant. Conversely, the upper body, especially the arm's typical target task is manipulation and their maximum expected load is below 5\,kg ($< 50$\,N), making the load-dependent dynamics minor. Therefore, load-dependent friction is considered for the legs requiring the desired joint torque $\bm{T}^{g}(k)$ as an input feature in addition to $\bm{Q}^{g}(k)$ and $\bm{U}^{V}(k)$. For the upper body, however, load-dependent friction is ignored and the friction torque becomes a function of joint velocity ($\bm{\tau}_{f}(\dot{\bm{q}})$) only requiring $\bm{Q}^{g}(k)$ and $\bm{U}^{V}(k)$ as input features. Virtual joints do not consider friction and require only $\bm{Q}^{g}(k)$ of all joints and $\bm{U}^{V}(k)$ to infer the modeling error torque.

\subsubsection{Cost of Data Collection}
\label{Subsubsection/Proposed Method/Input and Output Features/Cost of Data Collection}
For the target value of the uncertainty torque, the residual vector is used which can be calculated with only internal states. If we assume that no external force is applied, the residual vector during contact-free motion itself becomes the target uncertainty torque, which costs nothing more than the residual calculation. 
Thus, contact-free motion is used to get the training data for the upper body.
For the legs, Conversely, walking data is utilized without additional devices for data collection. 
If walking data is used, external force should be measured to obtain the model uncertainty torque.
In our robot, the FTS under the feet is used to measure the contact wrench, and the measured contact wrenches are converted to the external torque as (\ref{equation/external torque ft}). 
Alternatively, the free motion of the robot's legs can be recorded by suspending it in mid-air, neglecting the load-dependent friction and potential joint elasticity, similar to the upper body joints.
In this case, however, additional experimental devices are needed to hang the robot up in the air, to measure the lifting force for calculating the external force in the virtual joints, and to measure the linear velocity of the base link for calculating the residual of MOB. 
For our humanoid robot with a high gear ratio reducer (100:1), the load-dependent dynamics could not be ignored because of the large estimation error, and such experimental settings for hanging up the humanoid robot cost more than the FTS on the feet. 
Therefore, the residual minus the measured external torque $\bm{r}-\bm{\tau}_{e}^{m}$ from the walking data is used as the target value of the uncertainty torque learning for the networks of the legs, and only the residual $\bm{r}$ from the free motion data is used as the output for the upper body networks.
In the virtual joints, all the external forces are observed, thus $\bm{r}-\bm{\tau}_{e}^{m}$ is used for the target value.

\begin{figure}[t]
\centering
\vspace*{0.0cm}
\includegraphics[width=1.0\linewidth]{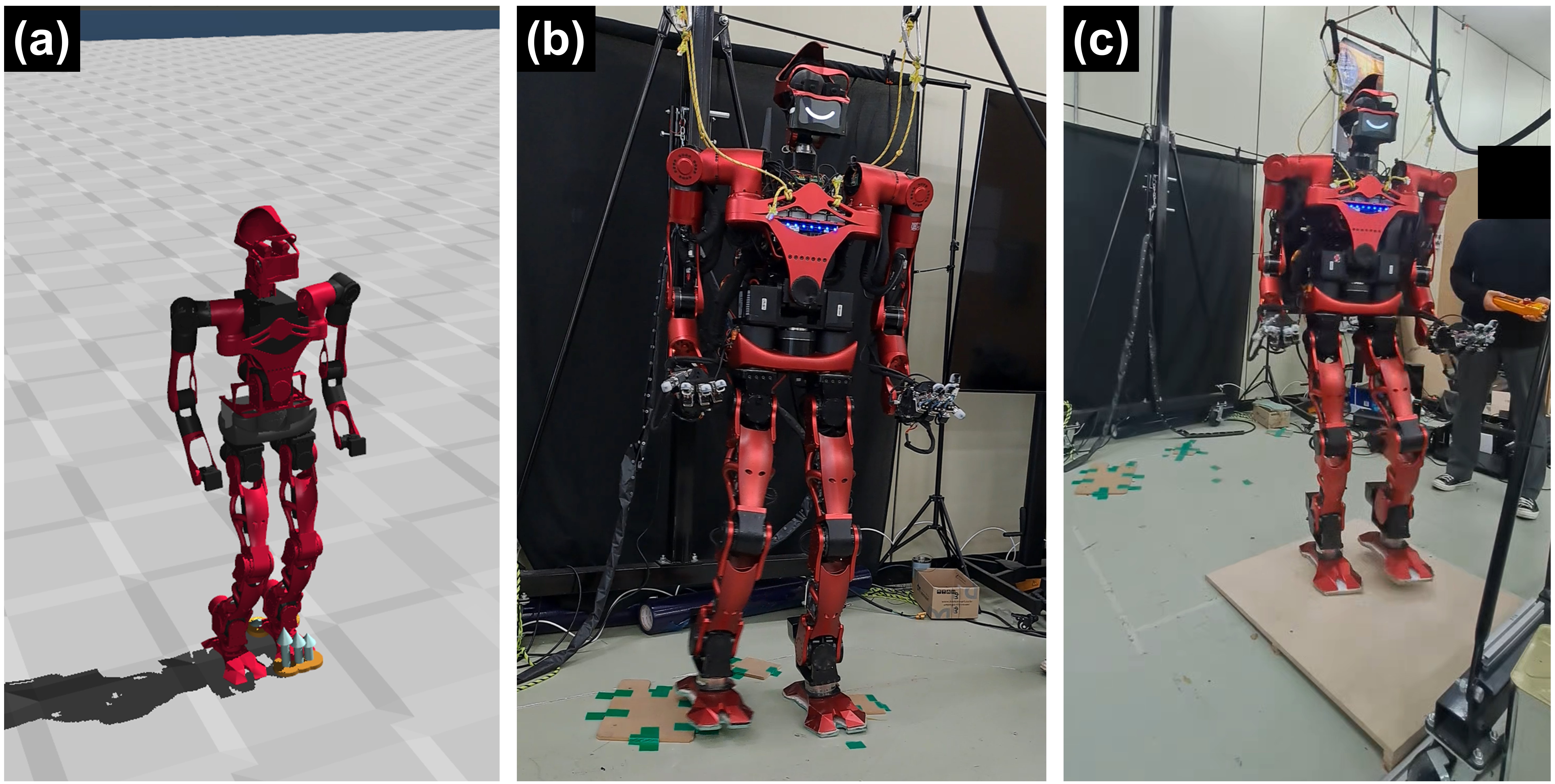}
\caption{\textbf{Random walking data collection environment for the simulation and the real robot experiment.} (a) Simulation environment with uneven terrain of check pattern. (b) Uneven terrain setting with three wooden plates of approximately 10mm thickness. (c) Slope terrain setting of approximately 3 degrees.}
\label{fig/data_collection_example}
\end{figure}

\subsection{Data Collection}
\label{Subsection/Proposed Method/Data Collection}
Two sets of data are collected using our humanoid robot, TOCABI, and the implemented controllers. The first data set (\emph{Random Walking Data}) is collected for the legs and the second set (\emph{Random Motion Data}) is collected for the upper body and the virtual joint, see \href{https://youtu.be/Z8Psm5RKFxk}{Extension 1}. Across both data sets, the data is collected with a 1000\,Hz sampling frequency which is half of the control frequency, 2000\,Hz. The data contains input and output features of all the joints including the virtual joints (39 DoF). The collected data is divided into training and validation data with a 9:1 ratio. 

\begin{table}[t]
\centering
\caption{Random walking parameters and their range}
\label{Table/walking parameters}
\resizebox{0.7\columnwidth}{!}{%
\begin{tabular}{@{}lc@{}}
\toprule
\textbf{Walking Parameters} & \textbf{Range}            \\ \midrule
Step Length X-dir          & {[}-0.15, 0.20{]}\,m        \\
Step Length Y-dir          & {[}-0.10, 0.10{]}\,m        \\
Turning Angle              & {[}-20, 20{]}\degree       \\
Step Duration              & \{0.6, 0.7, 0.8, 0.9\}\,s \\
Maximum foot height        & \{4.0, 5.0, 6.0, 7.0\}\,cm      \\ \bottomrule
\end{tabular}
}
\end{table}

\emph{Random Walking Data} is collected using the walking controller implemented in our robot (\cite{kim2023foot}). The operator can control the robot using a joystick with various walking patterns. The walking command consists of 2D step lengths and a 1D turning angle. Our robot can walk omnidirectionally given an arbitrary combination of walking commands within the predefined limitations. Additionally, the robot's step duration and the maximum foot height of the swing foot were also changed among several predefined values. The range of the random walking parameters is summarized in Table~\ref{Table/walking parameters}.
The data collection was performed in the uneven terrain environment as shown in Figure~\ref{fig/data_collection_example}.
Figure~\ref{fig/data_collection_example}~(a) shows the check-patterned uneven terrain with a depth difference of 15\,mm in the simulation.
Figure~\ref{fig/data_collection_example}~(b) and (c) show the uneven terrain and slope terrain settings for data collection for the real robot. Three wooden plates of 10\,mm thickness are placed on the floor in Figure~\ref{fig/data_collection_example}~(b), and a wooden slope of 3 degrees is placed next to the uneven terrain settings.
The robot walks over uneven terrain to produce wide-distributed and uncorrelated data. 
If the data is only collected from the flat terrain, the uncertainty torque can be correlated with the foot height. 
Additionally, an experimenter applied external disturbances to \textcolor{black}{the swing foot link and upper body of the robot for largely distributed data. The disturbances on the swing foot link are measured by FTS and considered as external forces. The disturbances on the upper body do not have to be measured because they act as disturbances to the base link. These disturbances on the upper body are measured by the IMU sensor and indirectly alter the external forces on the supporting foot, allowing for diverse data distribution.}
While the humanoid robot is walking, the other joints in the upper body are fixed. The total amount of collected data in 1 hour, resulted in approximately 3.6 million samples.

\emph{Random Motion Data} is collected using the teleoperation control framework developed in our group (\cite{lim2022online}) to randomly move the upper body of the humanoid robot. A person equips the motion trackers on the body and the human's motion data is recorded. The recorded motion data is mapped to the robot to mimic the human's motion according to the motion mapping and control framework in \cite{lim2022online}. During the data collection procedure, none of the external forces are applied to the upper body of the robot, but ground reaction forces are exerted on the feet and the contact wrenches are measured by FTS on the foot. To ensure that self-collision does not occur, a learning-based self-collision avoidance algorithm in \cite{koptev2021real} is implemented in the IK solver, and the random motion is tested in the simulation before data collection on the real robot. During the random motion of the upper body, the robot's leg was walking or standing with a 1:1 ratio. The robot's leg controls the center of mass of the robot to be at the center of the support polygon while the robot is standing. We collected 15 minutes of the random motion of one person and augmented the data 4 times by modifying the speed of the motion and adding offsets to the target positions of the two hands. Consequently, the total amount of collected data in 1 hour resulted in approximately 3.6 million samples, which is the same as the \emph{Random Walking Data}.

% add random joint torque
To obtain decent generalization capability from deep learning, largely distributed data that covers the potential target tasks is essential. This is because the data-driven method inherently shows superior generalization ability for interpolation problems, but shows poor performance for extrapolation problems. 
From this perspective, random actions are used in both data sets to generate the large distribution of the position states such as random walking commands and random motions within feasible constraints. For the velocity and desired torque states, however, the distribution of the training data can not cover all the collision data because the training data only contains collision-free data except for ground reactions on the feet. When a collision occurs on the knee link, for example, the accuracy of the estimation decreases because the velocity and torque of the collision-engaged joints abruptly surge in the swing phase, which is not observed in the training data. 

To mitigate this problem, \emph{Random Torque Exploration} (RTE) method is introduced. Random torques are added to the desired torques in every joint except the support leg joints to explore more distributed data during data collection. The random torque is designed as a step function with random timing and magnitude. The duration of the on/off period is sampled from the uniform distribution $\mathcal{U}(0.1, 0.5)$\,s, and the joint torque is also sampled from the uniform distribution $\mathcal{U}(\underline{\bm{\tau}}_{rnd}, \overline{\bm{\tau}}_{rnd})$\,Nm. This random torque sampling is performed independently for each joint. The maximum magnitude of the random torque ($\overline{\bm{\tau}}_{rnd}=-\underline{\bm{\tau}}_{rnd})$) was determined to be as large as possible until the robot maintains its stability; $\overline{\bm{\tau}}_{rnd,j}$ for leg and waist joints is 50\,Nm, and $\overline{\bm{\tau}}_{rnd,j}=[3, 15, 15, 10, 10, 5, 5, 5]\,Nm$ for the arm joints. Although this random torque can not cover all the possible collisions with limited maximum value, it is observed that applying random torque enhances the training results especially when the collision occurs. Validation for this can be found in Section \ref{Subsection/Ablation Study/Effectiveness of Random Torque Exploration}.

\subsection{Training}
\label{Subsection/Proposed Method/Training}
The network is trained in a supervised learning manner. The parameters of the GRU network are trained using Truncated Back Propagation Through Time (TBPTT) with a specific time horizon as specified in the last column of Table~\ref{Table/Network Architecture}. The training epoch and the batch size are 200 and 64, respectively. The Adam optimizer was used for the parameter update with the default betas (0.9, 0.999) because it shows the best performance among the three optimization algorithms [SGD, RMSprop, Adam]. The linear learning rate scheduler is used to decrease the learning rate from the initial value of 0.05 to 0.0005 during the first 100 epochs. It is observed that the linear learning rate scheduling shows more stable and better learning results than the constant learning rate.

Recall the Problem \ref{problem/uncertainty torque estimation} where the optimal mean and variance functions of the normal distribution should be found to estimate the true uncertainty torque. Let's interpret this problem within a deep learning framework where the optimal function of the normal distribution is a neural network $\bm{f}_{\theta}(\bm{X}_p)$:
% \begin{equation}
\begin{align}
\label{equation/uncertainty torque learning1}
&[\hat{\bm{\tau}}_{u, \theta^*}(\bm{X}_p), \hat{\bm{\sigma}}_{u, \theta^*}^2(\bm{X}_p)]= \bm{f}_{\theta^*}(\bm{X}_p), \\ 
\label{equation/uncertainty torque learning2}
&\theta^* \approx \argmax_{\theta} \ \sum_{i=1}^{n_T} \sum_{j=1}^{n_j} \mathcal{N}(\tau_{u,j}^{i}|\bm{X}_{p}^{i}, \theta),
\end{align}
% \end{equation}
where $n_T$ is the number of training data samples and $n_j$ is the number of joints for the uncertainty torque.
The conditional probability of the normal distribution is expressed in the exponential function as
\begin{align}
\label{equation/uncertainty torque learning}
\mathcal{N}({\tau}_{u,j}|\bm{X}_{p}, \theta) = \frac{1}{\sqrt{2\pi \hat{\sigma}_{u,j,\theta}^2}}\exp\Biggl(\frac{-\bigl(\hat{\tau}_{u,j,\theta}-\tau_{u,j}\bigl)^2}{2\hat{\sigma}_{u,j,\theta}^2}\Biggl).
\end{align}
So, if the natural log is taken on the right side of (\ref{equation/uncertainty torque learning2}), the Gaussian negative log-likelihood loss function $L$ is obtained as described in \cite{nix1994estimating}:
\begin{align}
\label{equation/uncertainty torque learning}
\theta^* &\approx \argmin_{\theta} \ \sum_{i=1}^{n_T} \sum_{j=1}^{n_j} -2\ln{(\mathcal{N}(\tau_{u,j}^{i}|\bm{X}_{p}^{i}, \theta))}, \\ 
L &= -2\ln{(\mathcal{N}(\tau_{u,j}^{i}|\bm{X}_{p}^{i}, \theta))} = \ln(\hat{{\sigma}}_{u,j,\theta}^2) + \frac{\bigl(\hat{\tau}_{u,j,\theta}-\tau_{u,j}\bigl)^2}{\hat{{\sigma}}_{u,j,\theta}^2}.
\end{align}
To prevent dividing by zero, the estimated standard deviation $\hat{{\sigma}}_{u}$ is clipped with the minimum value (1e-6) for the numerically stable loss function calculation.

%%%%%%%%%%%%%%%%%%%%%%%%%%%%%%%%%%%%%%%%%%%%%%%%%%%%%%%

\begin{figure}[!t]
\centering
\vspace*{0.0cm}
\includegraphics[width=1.0\linewidth]{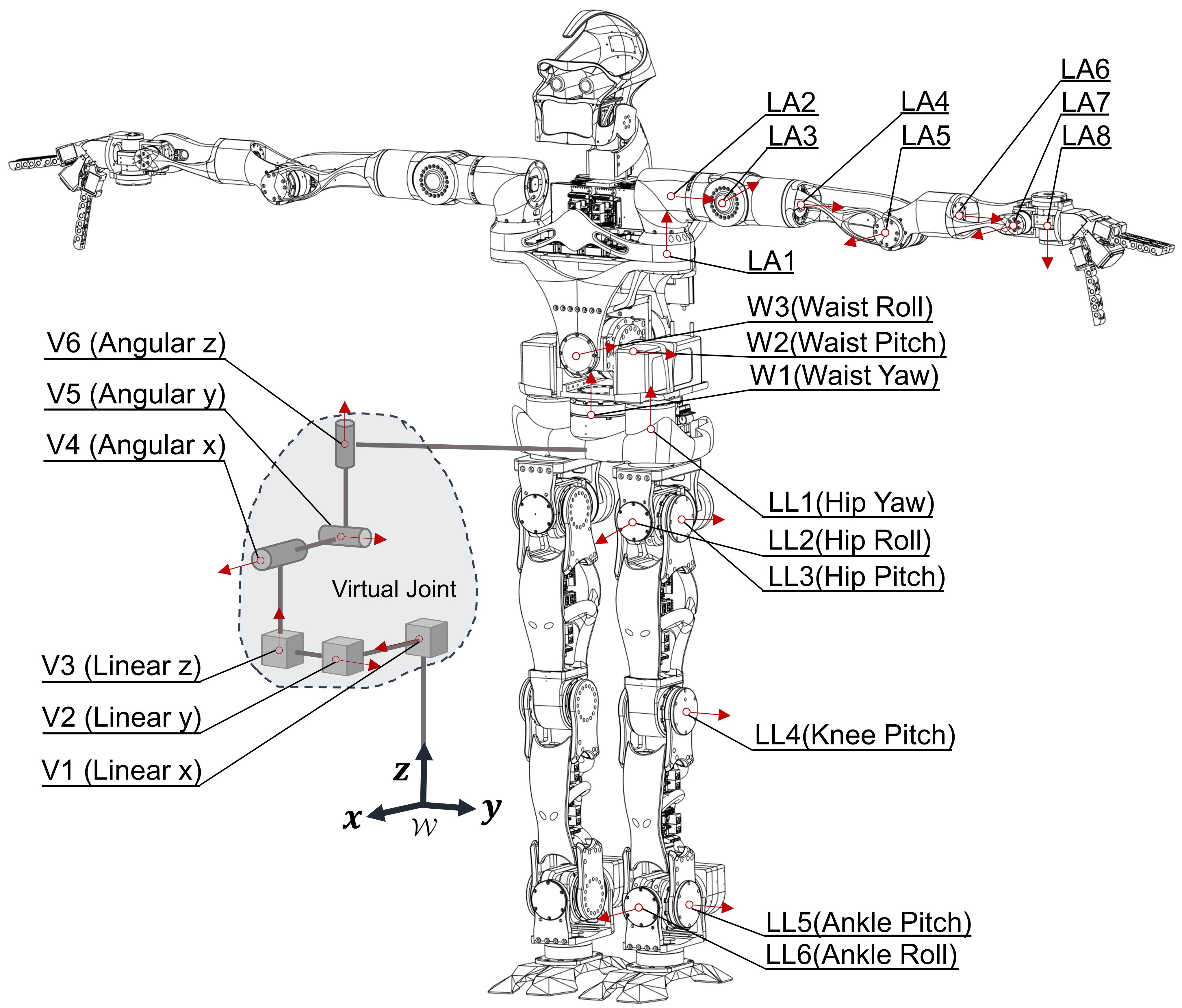}
\caption{\textbf{A humanoid robot, TOCABI, used in the experiment and its joint names}. Only the joint names in the left limbs are displayed due to their symmetry.}
\label{fig/tocabi_joint_index}
\end{figure}

\section{Experiment}
\label{Section/Experiments}

The experiments are conducted using the real robot and the simulation environment. Throughout the experiments, a full-size humanoid robot, TOCABI, is used, and only the results of one side of the leg and arm are demonstrated because the other side limbs show similar results.
The MuJoCo simulator in \cite{todorov2012mujoco} is used for the simulations.
For the robot's hardware specifications, TOCABI has 39 DoF (16 in both arms, 12 in both legs, 3 in the waist, 2 in the neck, and 6 in the virtual joint).
The joint names are displayed in Figure~\ref{fig/tocabi_joint_index}. 
The height is approximately 1.8m and the weight is approximately 100\,kg. The robot has FTS on each foot, an IMU sensor on the pelvis, and a motor side encoder in each joint. Each actuator consists of an electric brushless motor, a harmonic gear with 100:1 gear reduction, and an Elmo motor controller which controls the current of the motor. \textcolor{black}{The joint torque is controlled in the feedforward manner assuming that the joint torque is proportional to the motor current.} For the robot's software framework, ROS and real-time Linux kernel are used. The torque control frequency is 2\,kHz in both the simulation and real robot experiment while the data is collected at 1\,kHz. 
% For all experiments, the training is performed in a single desktop computer that has a single CPU (Intel Core i9-12900K), a single GPU (NVIDIA GeForce RTX 3090), and 128\,GB RAM.

%Sensor Noise
\begin{table}[!t]
\centering
\caption{The sensor noise used in the simulation is Gaussian noise with zero mean and different variance.}
\label{table/sensor_noise}
\resizebox{\columnwidth}{!}{%
\begin{tabular}{@{}lcccc@{}}
\toprule
\textbf{Variable}         & \textbf{Joint position} & \textbf{Joint velocity} & \textbf{Base linear acceleration} & \textbf{Base angular velocity} \\ \midrule
\textbf{\begin{tabular}[c]{@{}c@{}}Noise\\ variance\end{tabular}} & 1e-7            &2e-3             &1e-4                        &5e-3                     \\ \bottomrule
\end{tabular}%
}
\end{table}

 %firction Coeff
\begin{table}[!t]
\centering
\caption{The joint friction parameter used in the \emph{All uncertainty} simulation.}
\label{table/friction_parameters}
\resizebox{\columnwidth}{!}{%
\begin{tabular}{@{}lcccccc@{}}
\toprule
\textbf{Parameter} &
  \textbf{$f_{c}$} &
  \textbf{$f_{s}$} &
  \textbf{$v_{s}$} &
  \textbf{$k_{vf}$} &
  \textbf{$k_{lf}$} &
  \textbf{$\overline{\tau}_{loss}$} \\ \midrule
\textbf{Value} &
  5.0 &
  2.0 &
  1.51 &
  \begin{tabular}[c]{@{}c@{}}4.0 (legs) \\ 3.0 (otherwise)\end{tabular} &
  0.002 &
  \begin{tabular}[c]{@{}c@{}}10.0\,Nm (legs) \\ 8.0\,Nm (otherwise)\end{tabular} \\ \bottomrule
\end{tabular}%
}
\end{table}

\subsection{Comparison of External Joint Torque Estimation Performance in Simulation}
\label{Subsection/Experiment/Comparison of External Torque Estimation Performance in Simulation}
In this section, the estimation performance of MOB-Net is validated in the simulation environment where the ground truth of the estimation can be obtained easily. 
Three different levels of uncertainty are organized to measure the estimation performance of MOB, the backbone of MOB-Net, according to the uncertainty components, and to compare with the learning performance of MOB-Net. 
The three uncertainty levels are listed below.
\begin{itemize}
    \item \emph{Ideal}: The uncertainty torque is zero. All the inertia parameters of the robot are correctly identified and the actuator accurately produces the desired torque without friction.
    \item \emph{Sensor Noise}: Sensor noise is artificially added to the sensor measurements. The uncertainty is only from the sensor noise and is the baseline of the uncertainty torque learning. All the sensor noises are Gaussian noise with zero means, and the noise variances of each variable are summarized in Table \ref{table/sensor_noise}.
    \item \emph{All Uncertainty}: In this environment, three uncertainty components are intentionally added to the ideal simulation: sensor noise, modeling error, and joint friction.
    The modeling error is induced by decreasing the mass and inertia parameters of each link of the nominal model by 10\,\%. The nonlinear Stribeck joint friction torques, $\bm{\tau}_{sf}$, are applied for each joint and the load-dependent joint friction torques, $\bm{\tau}_{lf}$, are only applied to the leg joints. 
    \begin{equation}
    \begin{split}
        \tau_{f,j} &= \tau_{sf, j} + \tau_{lf, j}\\
        \tau_{sf, j} &= -\text{sgn}(\dot{q}_{j})(f_{c} + (f_{s}-f_{c})\exp({-|\dot{q}_{j}/v_{s}|}))\\
        &\ \ \ \ \ - k_{vf}\dot{q}_{j}\\
        \tau_{lf, j} &= -\text{sgn}(\dot{q}_{j})( k_{lf}|\tau_{m,j}|^{2} )
    \end{split}
    \end{equation} 
    Additionally, the friction loss option in the MuJoCo simulator is activated, which ignores the joint torques whose magnitude is below the threshold torque, $\overline{\tau}_{loss}$.
    All the parameters for the joint friction in the simulation are summarized in Table \ref{table/friction_parameters}.
\end{itemize}

% left leg forward walking results
\begin{table}[!t]
\centering
\caption{Simulation result of forward walking on uneven terrain. RMSE of the estimated external joint torque is summarized according to the uncertainty level and the estimation method. The dimension of all the values is [\emph{Nm}].}
\label{table/simulation_walking_forward_rough}
\resizebox{\columnwidth}{!}{%
\begin{tabular}{@{}ccccccccc@{}}
\toprule
\textbf{} &
  \textbf{} &
  \multicolumn{6}{c}{\textbf{Left Leg Joint}} &
  \textbf{} \\ \midrule
\textbf{Uncertainty level} &
  \textbf{Method} &
  \textbf{LL1} &
  \textbf{LL2} &
  \textbf{LL3} &
  \textbf{LL4} &
  \textbf{LL5} &
  \textbf{LL6} &
  \textbf{avg} \\ \midrule
\textbf{Ideal} &
  \textbf{MOB} &
  0.00 &
  0.01 &
  0.01 &
  0.01 &
  0.01 &
  0.01 &
  \textbf{0.01} \\ \midrule
\textbf{Sensor noise} &
  \textbf{MOB} &
  0.44 &
  1.69 &
  1.49 &
  1.21 &
  1.36 &
  1.16 &
  \textbf{1.22} \\ \midrule
 &
  \textbf{MOB} &
  7.73 &
  24.61 &
  18.67 &
  33.60 &
  17.63 &
  13.54 &
  \textbf{\begin{tabular}[c]{@{}c@{}}19.30\\ (+18.08)\end{tabular}} \\ \cmidrule(l){2-9} 
 &
  \textbf{MOB-Net} &
  \cellcolor[HTML]{C0C0C0}1.45 &
  \cellcolor[HTML]{C0C0C0}1.05 &
  \cellcolor[HTML]{C0C0C0}1.53 &
  \cellcolor[HTML]{C0C0C0}1.31 &
  \cellcolor[HTML]{C0C0C0}1.35 &
  \cellcolor[HTML]{C0C0C0}1.16 &
  \cellcolor[HTML]{C0C0C0}\textbf{\begin{tabular}[c]{@{}c@{}}1.31\\ (+0.09)\end{tabular}} \\ \cmidrule(l){2-9} 
\multirow{-6}{*}{\textbf{All uncertainty}} &
  \textbf{FTS-e2e} &
  \cellcolor[HTML]{C0C0C0}1.34 &
  \cellcolor[HTML]{C0C0C0}1.41 &
  \cellcolor[HTML]{C0C0C0}2.47 &
  \cellcolor[HTML]{C0C0C0}1.34 &
  \cellcolor[HTML]{C0C0C0}1.50 &
  \cellcolor[HTML]{C0C0C0}1.00 &
  \cellcolor[HTML]{C0C0C0}\textbf{\begin{tabular}[c]{@{}c@{}}1.51\\ (+0.29)\end{tabular}} \\ \bottomrule
\end{tabular}%
}
\end{table}

Each MOB-Net for each limb was trained using the corresponding training data that is collected in the simulation environment as shown in Figure~\ref{fig/data_collection_example}~(a).
For comparisons, two different methods are implemented as below.
\begin{itemize}
    \item \emph{MOB}: A discretized version of momentum observer introduced in \cite{bledt2018contact}. $\bm{K}_{0}=100$ is used for MOB. This MOB is also used for the uncertainty learning of MOB-Net.
    \item \emph{FTS-e2e}: Our previous method in \cite{lim2023proprioceptive} uses supervised learning in an end-to-end manner to directly learn the external joint torques of legs which are measured by FTS on each foot and transformed to the joint torques. FTS-e2e network is trained using the same training data and the same network structure as MOB-Net. FTS-e2e infers the external joint torques and the variance. 
\end{itemize}
The testing data for the leg joints contains the forward walking data for one minute on uneven terrain with a step duration of 0.7 s and a maximum foot height of 5.5 cm.
In Table \ref{table/simulation_walking_forward_rough}, the root-mean-square error (RMSE) of the external joint torque estimation is summarized according to the joint, uncertainty level, and the estimation method. 
In the \emph{Ideal} environment, the model-based method, MOB, can accurately estimate the external joint torque with approximately 0.01\,Nm errors for each joint.
When the sensor noise is added to the sensor measurements, the estimation errors increase by 1.31\,Nm on average. 
However, in \emph{All Uncertainty} environment, the average estimation error of MOB increases by 19.30\,Nm due to the additional modeling error and joint frictions. 
In the same uncertainty level, MOB-Net learns the uncertainty torque and cancels out the uncertainty torque from MOB resulting in significant improvement in the estimation by reducing the estimation error to 1.31\,Nm on average which is just 0.09\,Nm (6.7\%) larger than the MOB error in the \emph{Sensor Noise} environment.
The MOB errors in the \emph{Sensor Noise} environment are regarded as the minimum error that MOB-Net or FTS-e2e can achieve in \emph{All Uncertainty} because, unlike the other uncertainty components, the random sensor noise can not be inferred correctly.
FTS-e2e has similar estimation errors to MOB-Net for all joints except for the LL3 joint.

\begin{figure}[!t]
\centering
\vspace*{0.0cm}
\includegraphics[width=1.0\linewidth]{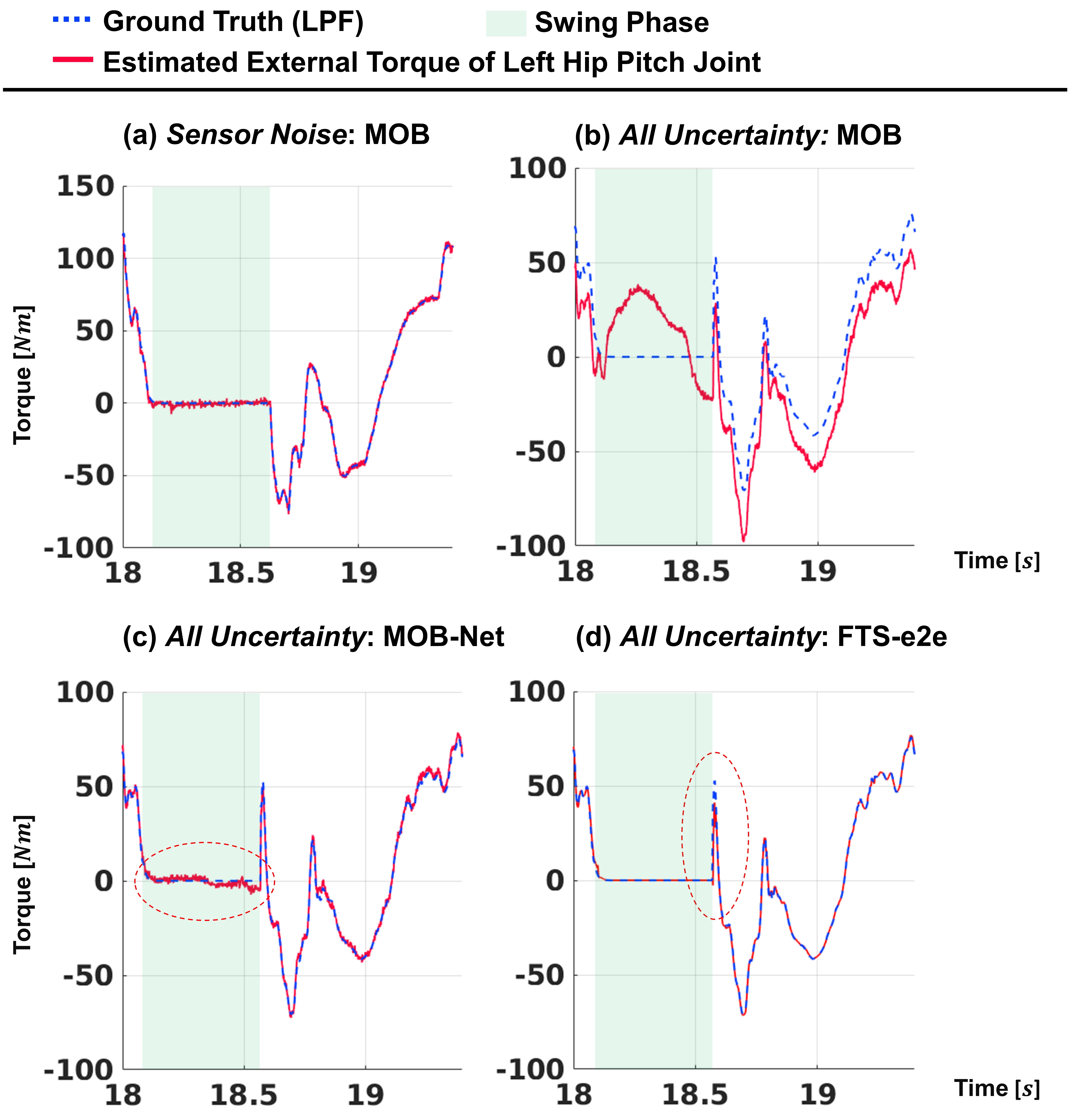}
\caption{\textbf{Simulation results of forward walking on uneven terrain.} The estimated external joint torque of the LL3 (left hip pitch) joint is plotted according to the uncertainty level and the estimation method. The robot walks forward on uneven terrain with a step duration of 0.7\,s and the plots show one cycle of the locomotion.}
\label{fig/simulation_walking_forward_rough}
\end{figure}

\begin{figure}[!t]
\centering
\vspace*{0.0cm}
\includegraphics[width=1.0\linewidth]{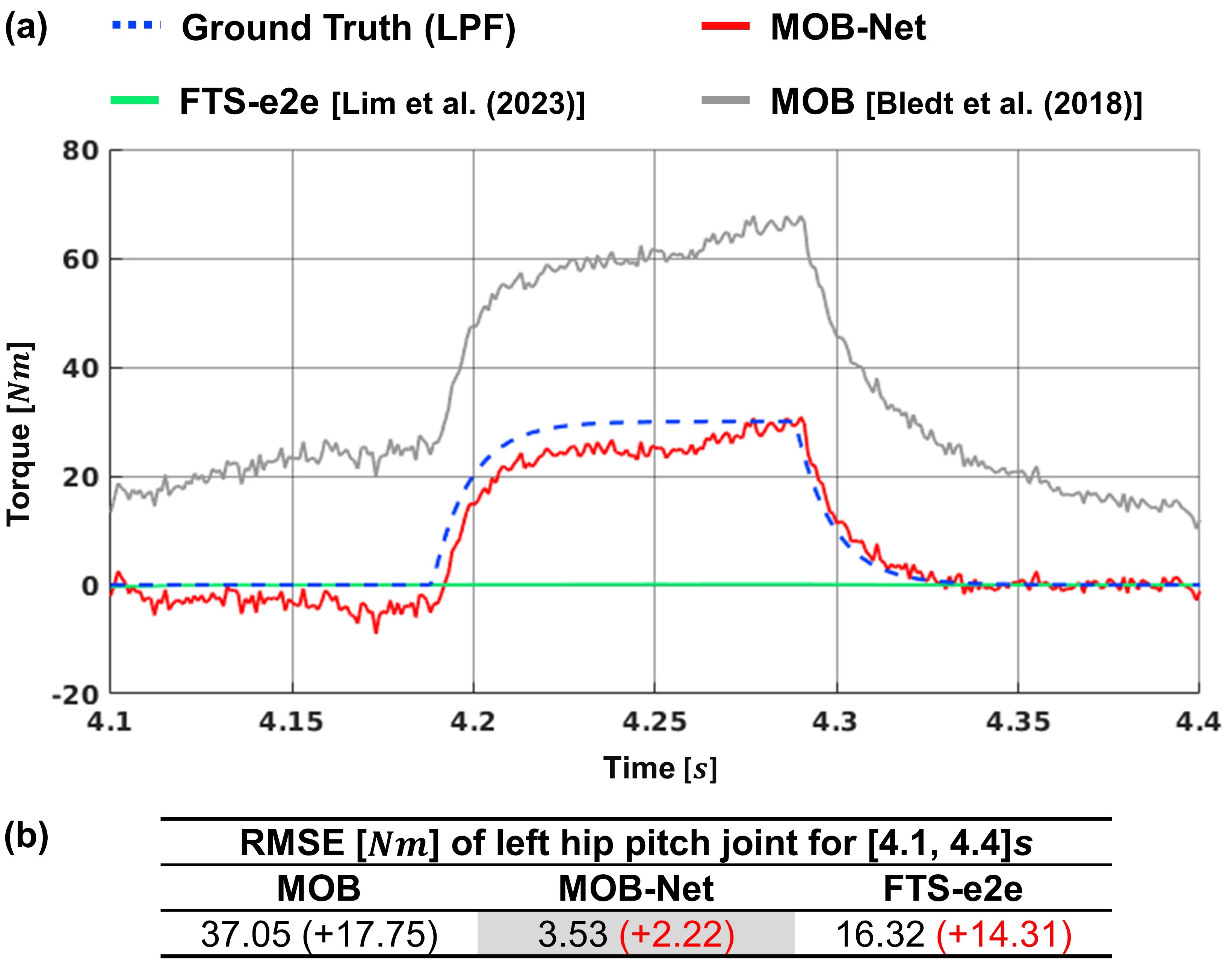}
\caption{\textbf{Simulation result of the step disturbance in the LL3 (left hip pitch) joint.} (a) The estimated external joint torque of the LL3 joint is displayed when the external joint torque of 30\,Nm is applied to the LL3 joint for 0.1\,s during the swing phase. (b) The RMSE of the external torque estimation in the LL3 joint for the plot range [4.1, 4.4]\,s.}
\label{fig/simulation_left_hip_pitch_dist_30nm}
\end{figure}

Figure~\ref{fig/simulation_walking_forward_rough} shows the ground truth and the estimated external joint torque of the hip pitch joint (LL3) for one cycle of the walking period in the test data which includes the swing and the supporting phase.
The swing phase is displayed with the green area on the plots. 
Figure~\ref{fig/simulation_walking_forward_rough}~(a) shows the MOB residual signals in the \emph{Sensor Noise} environment. The residual signal is noisy but estimates the true external joint torque closely.
Conversely, the estimated external joint torque of the MOB significantly deviates from the true values in Figure~\ref{fig/simulation_walking_forward_rough}~(b) due to the modeling errors and the joint frictions resulting in large errors (18.67\,Nm). 
However, in Figure~\ref{fig/simulation_walking_forward_rough}~(c), MOB-Net calibrates the erroneous signals of MOB by subtracting the uncertainty torque and results in a similar estimation performance to MOB with the sensor noise.
The MOB-Net shows noisy signals similar to the MOB residual as marked with a dotted red circle in Figure~\ref{fig/simulation_walking_forward_rough}~(c).
FTS-e2e shows a much smoother estimation signal than MOB-Net as shown in Figure~\ref{fig/simulation_walking_forward_rough}~(d). However, when the target value changes quickly, FTS-e2e can not estimate the sharp signal as shown in the dotted red circle in Figure~\ref{fig/simulation_walking_forward_rough}~(d).
This is because the estimated signals from the neural network are inherently smooth but MOB-Net utilizes both MOB and the neural networks and the sharp estimation signals of the MOB appear in the estimated values of MOB-Net. 

Walking data with a disturbance on the swing leg is tested to investigate the estimation performance for the unseen data. 
Unlike the external torque propagated from the ground reaction forces through the foot link, disturbance on the swing leg through other links except the foot link, generates unseen patterns from the training data.
Therefore, a step external joint torque of 30\,Nm is applied on the hip pitch (LL3) joint during the swing phase to mimic the collision on the knee.
As shown in Figure~\ref{fig/simulation_left_hip_pitch_dist_30nm}~(a), MOB residual signal (gray line) resembles the changes of ground truth (dotted line) even if it shows large offset and errors of 37.05\,Nm due to the uncertainty torque. 
MOB-Net estimates the external torque with relatively small errors (3.53\,Nm) which is larger than the estimation errors of the straight walking test by 2.22\,Nm.
However, FTS-e2e can not estimate the external torque and infers almost zero external torques for the disturbance resulting in large errors (16.32\,Nm). 
Even though MOB-Net and FTS-e2e use the same training data and the same network, the combination of the model-based method and the data-driven method (MOB-Net) shows robust performance for the unseen data compared with just the data-driven method (FTS-e2e) or model based-method (MOB).
The robustness of MOB-Net for the unseen data is possible due to the use of MOB as a base algorithm of MOB-Net.
Even if the GRU network causes large estimation errors for the unseen data, MOB can capture the changes of the external torque based on the dynamics of the robot and the estimation error is not significantly increased in MOB-Net unlike the end-to-end learning method, FTS-e2e. 
Note that the ground truth in Figure~\ref{fig/simulation_left_hip_pitch_dist_30nm}~(a) is the low pass filtered signal of the step torque with the same cutoff frequency of MOB because MOB estimates the low pass filtered external torque theoretically and it is fair to compare with the low pass filtered true value. 

% upper body random_motion1_test results
\begin{table*}[!t]
\centering
\caption{Simulation result of the upper body joints and virtual joints for the three test data sets: \emph{Collision free motion, Left hand load (10\,N), and Left hand load (30\,N)}. RMSE of the estimated external joint torque or force are summarized according to the uncertainty level, estimation method, and data set.}
\label{table/simulation_upperbody_rmse}
\resizebox{1.0\textwidth}{!}{%
\begin{tabular}{@{}cccccccccccccccccccccc@{}}
\toprule
\textbf{} &
  \textbf{} &
  \multicolumn{12}{c|}{\textbf{Upper body joint}} &
  \multicolumn{8}{c}{\textbf{Virtual joint}} \\ \midrule
\textbf{\begin{tabular}[c]{@{}c@{}}Uncertainty\\ level\end{tabular}} &
  \textbf{Method} &
  \textbf{W1} &
  \textbf{W2} &
  \textbf{W3} &
  \textbf{LA1} &
  \textbf{LA2} &
  \textbf{LA3} &
  \textbf{LA4} &
  \textbf{LA5} &
  \textbf{LA6} &
  \textbf{LA7} &
  \textbf{LA8} &
  \multicolumn{1}{c|}{\textbf{\begin{tabular}[c]{@{}c@{}}Avg \\ (W+LA)\end{tabular}}} &
  \textbf{\begin{tabular}[c]{@{}c@{}}V1\\ \text{[\emph{N}]}\end{tabular}} &
  \textbf{\begin{tabular}[c]{@{}c@{}}V2\\ \text{[\emph{N}]}\end{tabular}} &
  \textbf{\begin{tabular}[c]{@{}c@{}}V3\\ \text{[\emph{N}]}\end{tabular}} &
  \textbf{\begin{tabular}[c]{@{}c@{}}Avg\\ (V lin)\end{tabular}} &
  \textbf{V4} &
  \textbf{V5} &
  \textbf{V6} &
  \textbf{\begin{tabular}[c]{@{}c@{}}Avg\\ (V ang)\end{tabular}} \\ \midrule
\multicolumn{22}{c}{\textbf{Data Set 1: \emph{Collision free motion}}} \\ \midrule
\textbf{Ideal} &
  \textbf{MOB} &
  0.01 &
  0.02 &
  0.02 &
  0.00 &
  0.00 &
  0.00 &
  0.00 &
  0.00 &
  0.00 &
  0.00 &
  0.00 &
  \multicolumn{1}{c|}{\textbf{0.01}} &
  0.05 &
  0.08 &
  0.08 &
  \textbf{0.07} &
  0.02 &
  0.04 &
  0.01 &
  \textbf{0.03} \\ \midrule
\textbf{Sensor noise} &
  \textbf{MOB} &
  1.85 &
  2.09 &
  2.42 &
  0.58 &
  0.55 &
  0.49 &
  0.35 &
  0.26 &
  0.06 &
  0.10 &
  0.05 &
  \multicolumn{1}{c|}{\textbf{0.80}} &
  3.98 &
  3.37 &
  3.77 &
  \textbf{3.71} &
  2.46 &
  2.04 &
  1.59 &
  \textbf{2.03} \\ \midrule
 &
  \textbf{MOB} &
  9.43 &
  11.77 &
  7.91 &
  7.52 &
  9.95 &
  10.21 &
  10.31 &
  10.97 &
  12.70 &
  10.51 &
  10.28 &
  \multicolumn{1}{c|}{\textbf{10.14}} &
  5.45 &
  7.24 &
  109.21 &
  \textbf{40.63} &
  4.75 &
  18.72 &
  2.32 &
  \textbf{8.60} \\ \cmidrule(l){2-22} 
\multirow{-2.5}{*}{\textbf{All uncertainty}} &
  \textbf{MOB-Net} &
  \cellcolor[HTML]{C0C0C0}2.84 &
  \cellcolor[HTML]{C0C0C0}2.75 &
  \cellcolor[HTML]{C0C0C0}3.64 &
  \cellcolor[HTML]{C0C0C0}2.72 &
  \cellcolor[HTML]{C0C0C0}1.45 &
  \cellcolor[HTML]{C0C0C0}1.63 &
  \cellcolor[HTML]{C0C0C0}1.91 &
  \cellcolor[HTML]{C0C0C0}1.63 &
  \cellcolor[HTML]{C0C0C0}0.84 &
  \cellcolor[HTML]{C0C0C0}1.31 &
  \cellcolor[HTML]{C0C0C0}2.06 &
  \multicolumn{1}{c|}{\cellcolor[HTML]{C0C0C0}\textbf{2.07}} &
  \cellcolor[HTML]{C0C0C0}4.87 &
  \cellcolor[HTML]{C0C0C0}4.77 &
  \cellcolor[HTML]{C0C0C0}4.56 &
  \cellcolor[HTML]{C0C0C0}\textbf{4.73} &
  \cellcolor[HTML]{C0C0C0}3.41 &
  \cellcolor[HTML]{C0C0C0}3.34 &
  \cellcolor[HTML]{C0C0C0}1.95 &
  \cellcolor[HTML]{C0C0C0}\textbf{2.90} \\ \midrule
\multicolumn{22}{c}{\textbf{Data Set 2: \emph{Left hand load (10N)}}} \\ \midrule
 &
  \textbf{MOB} &
  9.34 &
  9.73 &
  8.02 &
  5.95 &
  10.71 &
  9.21 &
  9.99 &
  9.99 &
  12.18 &
  9.96 &
  8.96 &
  \multicolumn{1}{c|}{\textbf{9.46}} &
  7.77 &
  4.19 &
  110.44 &
  \textbf{40.80} &
  4.01 &
  16.82 &
  3.58 &
  \textbf{8.13} \\ \cmidrule(l){2-22} 
\multirow{-2.5}{*}{\textbf{All uncertainty}} &
  \textbf{MOB-Net} &
  \cellcolor[HTML]{C0C0C0}2.95 &
  \cellcolor[HTML]{C0C0C0}2.45 &
  \cellcolor[HTML]{C0C0C0}2.88 &
  \cellcolor[HTML]{C0C0C0}2.47 &
  \cellcolor[HTML]{C0C0C0}1.04 &
  \cellcolor[HTML]{C0C0C0}1.89 &
  \cellcolor[HTML]{C0C0C0}1.60 &
  \cellcolor[HTML]{C0C0C0}1.73 &
  \cellcolor[HTML]{C0C0C0}0.66 &
  \cellcolor[HTML]{C0C0C0}0.91 &
  \cellcolor[HTML]{C0C0C0}2.08 &
  \multicolumn{1}{c|}{\cellcolor[HTML]{C0C0C0}\textbf{1.88}} &
  \cellcolor[HTML]{C0C0C0}5.51 &
  \cellcolor[HTML]{C0C0C0}3.79 &
  \cellcolor[HTML]{C0C0C0}6.25 &
  \cellcolor[HTML]{C0C0C0}\textbf{5.18} &
  \cellcolor[HTML]{C0C0C0}2.93 &
  \cellcolor[HTML]{C0C0C0}2.44 &
  \cellcolor[HTML]{C0C0C0}1.81 &
  \cellcolor[HTML]{C0C0C0}\textbf{2.39} \\ \midrule
\multicolumn{22}{c}{\textbf{Data Set 3: \emph{Left hand load (30N)}}} \\ \midrule
 &
  \textbf{MOB} &
  11.44 &
  10.38 &
  11.88 &
  6.51 &
  8.33 &
  9.70 &
  9.74 &
  10.66 &
  10.17 &
  8.53 &
  6.97 &
  \multicolumn{1}{c|}{\textbf{9.48}} &
  5.99 &
  12.77 &
  109.66 &
  \textbf{42.81} &
  7.81 &
  16.75 &
  3.43 &
  \textbf{9.33} \\ \cmidrule(l){2-22} 
\multirow{-2.5}{*}{\textbf{All uncertainty}} &
  \textbf{MOB-Net} &
  \cellcolor[HTML]{C0C0C0}2.50 &
  \cellcolor[HTML]{C0C0C0}3.01 &
  \cellcolor[HTML]{C0C0C0}3.79 &
  \cellcolor[HTML]{C0C0C0}2.37 &
  \cellcolor[HTML]{C0C0C0}1.73 &
  \cellcolor[HTML]{C0C0C0}2.15 &
  \cellcolor[HTML]{C0C0C0}1.37 &
  \cellcolor[HTML]{C0C0C0}1.96 &
  \cellcolor[HTML]{C0C0C0}1.48 &
  \cellcolor[HTML]{C0C0C0}1.18 &
  \cellcolor[HTML]{C0C0C0}2.42 &
  \multicolumn{1}{c|}{\cellcolor[HTML]{C0C0C0}\textbf{2.18}} &
  \cellcolor[HTML]{C0C0C0}5.84 &
  \cellcolor[HTML]{C0C0C0}9.71 &
  \cellcolor[HTML]{C0C0C0}7.25 &
  \cellcolor[HTML]{C0C0C0}\textbf{7.60} &
  \cellcolor[HTML]{C0C0C0}5.21 &
  \cellcolor[HTML]{C0C0C0}3.64 &
  \cellcolor[HTML]{C0C0C0}2.68 &
  \cellcolor[HTML]{C0C0C0}\textbf{3.84} \\ \bottomrule
\end{tabular}%
}
\end{table*}

The other limb groups (LA, W, V) are also validated in the simulation. 
Unlike the legs, MOB-Net for the upper body limbs (LA, W) does not include the desired torque in the input vector and the training data only consists of collision free data.
The networks of the upper body are trained using random motion data explained in Section \ref{Subsection/Proposed Method/Data Collection}.
The test data consists of random upper body motion and random walking commands with 0.7\,s step duration.  
Three test data sets are tested with the same random motion: \emph{Collision free motion}, \emph{Left hand load (10\,N)}, and \emph{Left hand load (30\,N)}.
In \emph{Collision free motion} data set, the robot did not receive any external forces while walking and moving the upper body randomly, and this data set is for the baseline of the learning performance as an in-distribution data. \emph{Collision free motion} contains data from three different uncertainty levels. 
In two \emph{Left hand load} data sets, the humanoid robot receives a step downward force on the left hand along the gravitational direction for 5 seconds, [10, 15]\,s while walking and moving the upper body randomly, similar to the collision free motion data. 
The two \emph{Left hand load} data sets are tested to measure the estimation performance for the external force data, i.e., unseen data.
Note that the RMSE of left hand load tests is calculated only for the disturbance duration ([10, 15]\,s) whereas the RMSE of collision free motion is calculated for the entire test data ([0, 60]\,s).

% upper body lhand dist
\begin{figure}[!t]
\centering
\vspace*{0.0cm}
\includegraphics[width=0.995\linewidth]{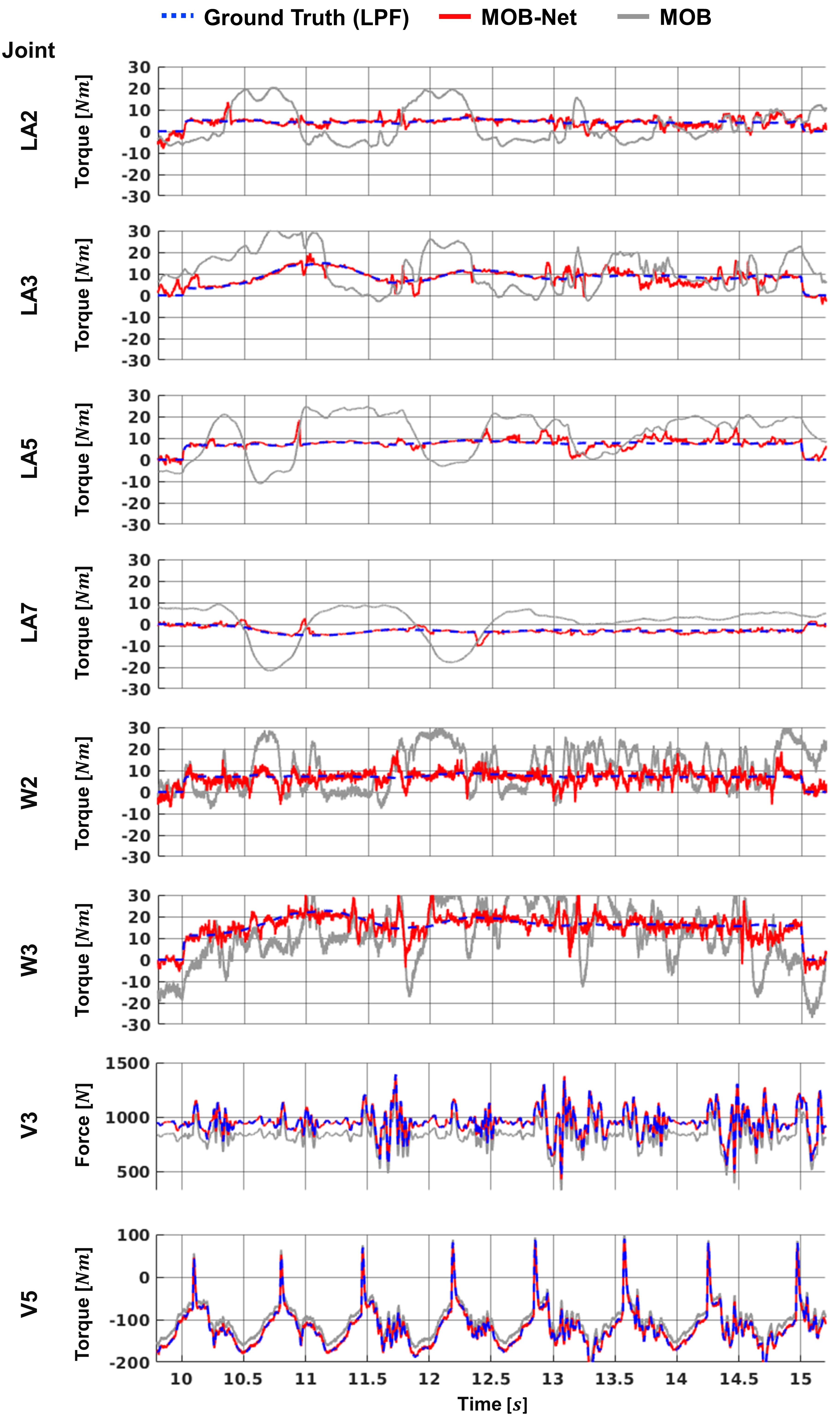}
\caption{\textbf{Simulation result of the partial upper body joints and the virtual joints for the \emph{Left hand load (30N)} data set.} A downward linear force of 30\,N is applied on the left hand during [10, 15]\,s while the robot is walking and moving the upper body.}
\label{fig/simulation_lhand_dist_30N_upper_body_joint}
\end{figure}

All the test results of the upper body joints and the virtual joints for the three data sets are summarized in Table \ref{table/simulation_upperbody_rmse}.
In the \emph{Collision free motion}, the results of MOB show a similar tendency to the results of the left leg across the uncertainty levels. For the ideal environment, the estimation error is almost zero, but the error increases to 0.80 and 10.14 \,Nm for the upper body joint on average as the uncertainty components are added in the sensor noise and all uncertainty, respectively.
Additionally, the estimation errors of MOB in virtual joints increase as the uncertainty components are added.
For the virtual joints, joint friction torque is not applied and the modeling error is significant in the estimation error.
MOB-Net successfully decreases the average estimation error for the upper body joints to 2.07\,Nm, and for the virtual joints, it reduces the error to an average of 4.73\,N in linear joints and 2.90\,Nm in angular joints.

In the \emph{Left hand load (10\,N)} scenario, the estimation errors of MOB-Net are comparable to those observed in the collision-free motion test. This indicates that MOB-Net is effectively generalizable to minor disturbances which is unseen data.
% In the \emph{Left hand load (10\,N)} results, all the methods have similar errors to the results of the collision free motion test, which means that MOB-Net for the upper body joints and virtual joints can be generalized to the small disturbance, 10\,N force on the -z-axis. 
In the \emph{Left hand load (30\,N)} results, compared to the results of 10\,N load, the estimation error of MOB-Net increases by 0.30\,Nm on average for the upper body, but the errors of the virtual joint network increase by 2.42\,N and 1.45\,Nm for the linear and angular virtual joints, respectively.
Similar or smaller amounts of increase in errors are also observed in the MOB results.
These results demonstrate the robust performance of MOB-Net over the external forces on the hand although the training data contains only collision-free motion. 

Figure~\ref{fig/simulation_lhand_dist_30N_upper_body_joint} shows the ground truth of external joint torque, the estimated value of MOB-Net, and the estimated value of MOB for the load of 30\,N on the left hand with all uncertainties. The load is applied on the left hand of the robot in a step function for 5 seconds ([10, 15]\,s) and the ground truth of the external joint torque is calculated using the Jacobian matrix of the hand link. In general, MOB-Net follows the ground truth but the estimated value of MOB-Net shows minor noisy signals due to the sensor noise whereas MOB has larger errors due to the uncertainty torque.
In the last two plots of Figure~\ref{fig/simulation_lhand_dist_30N_upper_body_joint}, two virtual joints, V3 (linear joint on the z-axis) and V5 (angular joint on the y-axis), are demonstrated. In the virtual joints, the estimated value of MOB-Net tracks the true value but MOB shows large bias signals due to the modeling error with a 10\% lighter nominal model.

\subsection{Comparison of External Joint Torque Estimation Performance in Real Humanoid Robot}
\label{Subsection/Experiment/Comparison of External Torque Estimation Performance in Real Humanoid Robot}
In this experiment, the estimation performance of MOB-Net is validated using a real humanoid robot.
Similar to the previous simulation section, the proposed method is compared with two methods (FTS-e2e, MOB), and an additional model-based method is also implemented for the comparison: \emph{MOB-fric}.
Proposed in \cite{lee2015sensorless}, \emph{MOB-fric} utilizes a friction model to estimate more accurate external joint torques. The friction model requires joint velocity and joint torque for the estimation of the joint friction. It consists of coulomb, static, and viscous friction, and the model parameters are regressed using the same training data of MOB-Net.

\begin{figure}[!t]
\centering
\vspace*{0.0cm}
\includegraphics[width=1.0\linewidth]{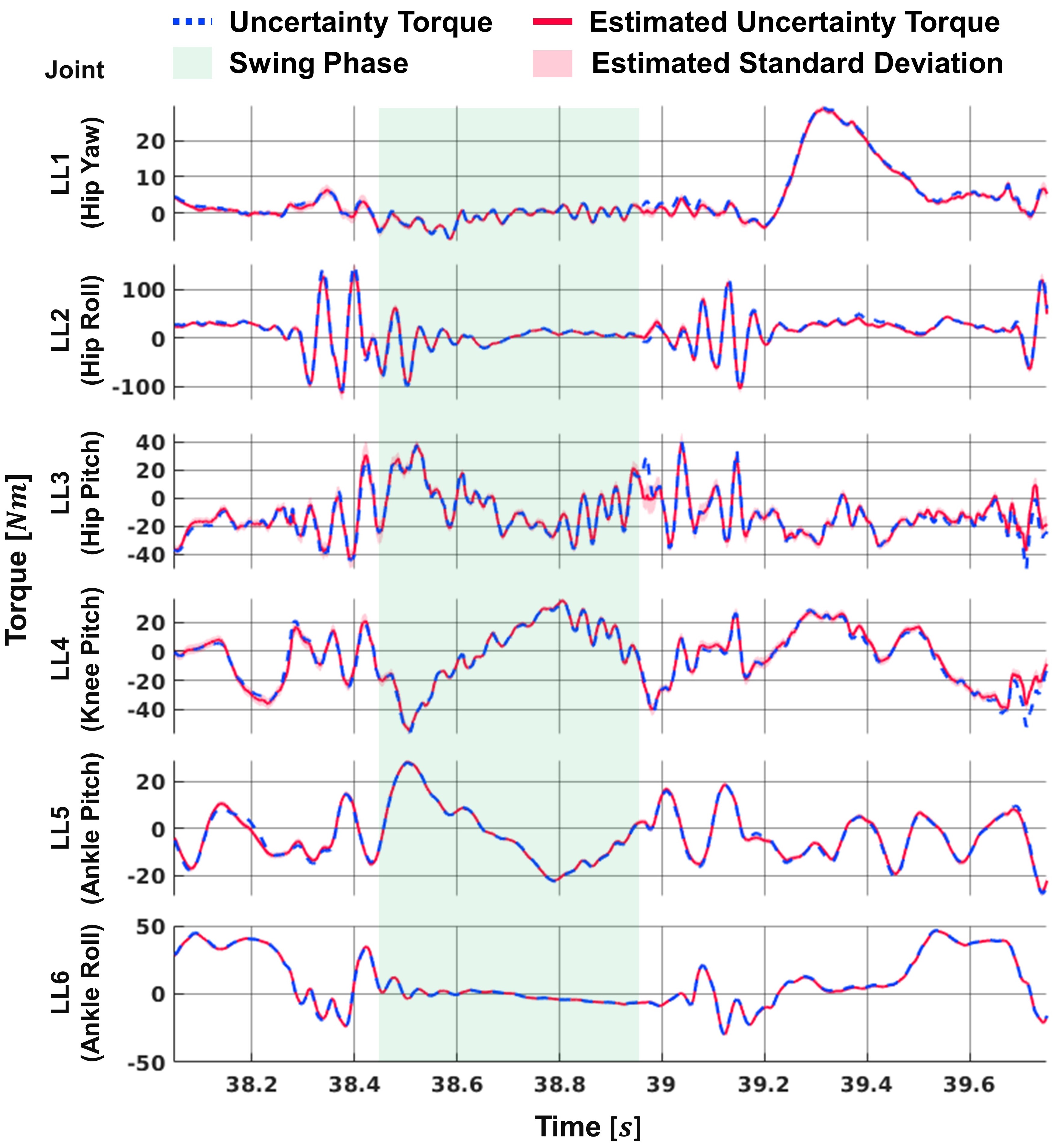}
\caption{\textbf{Uncertainty torque estimation of the joints in the left leg using MOB-Net and the corresponding target value.}}
\label{fig/uncertainty_torque_estimation_lleg}
\end{figure}

The estimation performance of the legs is tested first. 
The testing data contains the random walking data for one minute in the uneven terrain with a step duration of 0.7\,s and a maximum foot height of 5.5\,cm.
The estimated uncertainty torque of the left leg is plotted in Figure~\ref{fig/uncertainty_torque_estimation_lleg} with the target value which is calculated by subtracting the external joint torque measured by FTS on the foot from the residual vector of MOB.
The weight of the foot link is calibrated from the FTS measurements. 
The dotted blue line indicates the target value of the uncertainty torque learning, the red line indicates the estimated uncertainty torque from MOB-Net, and the red area around the estimated uncertainty torque shows the estimated standard deviation of the uncertainty torque from MOB-Net. 
This plot shows one cycle of the walking period consisting of the swing phase (green area) at first and the supporting phase following the swing phase. 
The estimated uncertainty torque follows the target value closely with RMSE of [0.980	6.266	4.191	2.731	0.917	0.598]\,Nm.
It is noted that in Figure~\ref{fig/uncertainty_torque_estimation_lleg} the estimation error of uncertainty torque increases when the robot's swing foot lands (right after the swing phase, the green area) and the estimated standard deviation of the uncertainty torque in the LL3 joint also increases accordingly due to the large sensor noise of the landing impact.

%%% uncertainty_torque_estimation_lleg

%%% comparison_of_external_torque_estimation
\begin{figure}[!t]
\centering
\vspace*{0.0cm}
\includegraphics[width=1.0\linewidth]{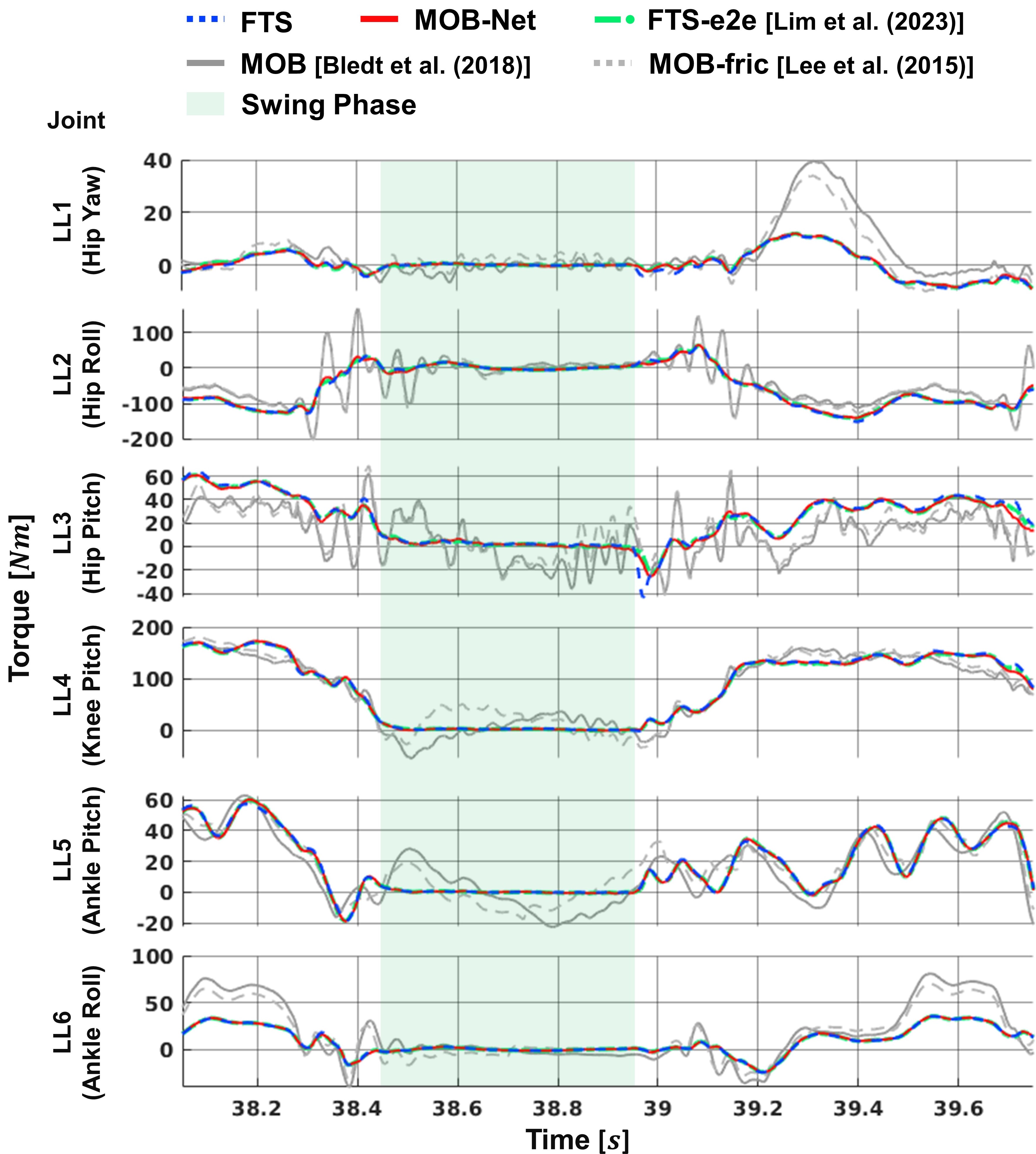}
\caption{\textbf{Comparison of external torque estimation according to the estimation methods in the left leg joints.}}
\label{fig/comparison_of_external_torque_estimation}
\end{figure}

The estimation result of the external joint torque according to each method is depicted in Figure~\ref{fig/comparison_of_external_torque_estimation}. 
The blue dotted line is the ground truth of the external joint torque measured by the FTS.
The red line indicates the estimated external torque of the proposed method, MOB-Net. 
The green line indicates the estimated external torque of our previous method, FTS-e2e.
Lastly, the gray line and the dotted gray line show the estimated external torques from MOB and MOB-fric, respectively.
As shown in Figure~\ref{fig/comparison_of_external_torque_estimation}, the two data-driven methods, MOB-Net and FTS-e2e, show more accurate estimation results following the target value closely than the two model-based methods, MOB and MOB-fric. 
During the swing phase (green area), MOB-Net and FTS-e2e estimate near zero torques whereas MOB and MOB-fric estimate non-zero torques due to the uncertainty torques. 
% During the step change around 9.4s, however, MOB-Net fluctuates and deviates from the target value while the estimated signal of FTS-e2e follows the target value more closely.

\begin{table}[!t]
\centering
\caption{RMSE and \emph{r}-RMSE of External Torque Estimation of Various Methods for Left Leg.}
\label{table/RMSE of external torque estimation of left leg}
\resizebox{\columnwidth}{!}{%
\begin{tabular}{@{}ccccccccc@{}}
\toprule
 &
   &
  \multicolumn{6}{c}{\textbf{Left leg joint}} &
  \textbf{} \\ \cmidrule(l){3-9} 
\multirow{-2.5}{*}{\textbf{Method}} &
  \multirow{-2.5}{*}{\textbf{Statistics}} &
  \textbf{LL1} &
  \textbf{LL2} &
  \textbf{LL3} &
  \textbf{LL4} &
  \textbf{LL5} &
  \textbf{LL6} &
  \textbf{avg} \\ \midrule
 &
  \textbf{RMSE [\emph{Nm}]} &
  8.589 &
  37.380 &
  25.673 &
  26.921 &
  15.238 &
  18.121 &
  \textbf{21.987} \\
\multirow{-2}{*}{\textbf{MOB}} &
  \textbf{r-RMSE [\%]} &
  33.76 &
  32.65 &
  18.94 &
  13.18 &
  15.24 &
  35.09 &
  - \\ \midrule
 &
  \textbf{RMSE [\emph{Nm}]} &
  6.439 &
  36.907 &
  23.960 &
  23.398 &
  11.929 &
  14.556 &
  \textbf{19.531} \\
\multirow{-2}{*}{\textbf{MOB-fric}} &
  \textbf{r-RMSE [\%]} &
  25.31 &
  32.24 &
  17.68 &
  11.45 &
  11.93 &
  28.18 &
  - \\ \midrule
 &
  \textbf{RMSE [\emph{Nm}]} &
  \cellcolor[HTML]{C0C0C0}0.908 &
  \cellcolor[HTML]{C0C0C0}6.266 &
  \cellcolor[HTML]{C0C0C0}4.191 &
  \cellcolor[HTML]{C0C0C0}2.731 &
  \cellcolor[HTML]{C0C0C0}0.917 &
  \cellcolor[HTML]{C0C0C0}0.598 &
  \cellcolor[HTML]{C0C0C0}\textbf{2.602} \\
\multirow{-2}{*}{\textbf{MOB-Net}} &
  \textbf{r-RMSE [\%]} &
  3.57 &
  5.47 &
  3.09 &
  1.34 &
  0.92 &
  1.16 &
  - \\ \hline
 &
  \textbf{RMSE [\emph{Nm}]} &
  \cellcolor[HTML]{C0C0C0}0.727 &
  \cellcolor[HTML]{C0C0C0}4.054 &
  \cellcolor[HTML]{C0C0C0}4.372 &
  \cellcolor[HTML]{C0C0C0}2.719 &
  \cellcolor[HTML]{C0C0C0}0.987 &
  \cellcolor[HTML]{C0C0C0}0.531 &
  \cellcolor[HTML]{C0C0C0}\textbf{2.232} \\
\multirow{-2}{*}{\textbf{FTS-e2e}} &
  \textbf{r-RMSE [\%]} &
  2.86 &
  3.54 &
  3.23 &
  1.33 &
  0.99 &
  1.03 &
  - \\ \bottomrule
\end{tabular}%
}
\end{table}

The RMSE and relative RMSE (r-RMSE) of the external torque estimation for each method are summarized in Table~\ref{table/RMSE of external torque estimation of left leg}. 
r-RMSE is the percentage of the RMSE relative to the maximum external joint torque measured by FTS for each joint.
The result shows that MOB-Net has an error of 2.602\,Nm on average.
Among the leg joints, the hip roll joint (LL2) has the largest estimation error whereas ankle joints (LL5 and LL6) have less than 1\,Nm errors.
In this random walking test data, FTS-e2e shows a similar estimation performance to MOB-Net resulting in 2.232\,Nm error on average.
In FTS-e2e, LL3 joint shows the largest error among the leg joints.
The two model-based methods, MOB and MOB-fric, show much higher estimation error than the data-driven methods due to the model uncertainty whereas the friction model reduces the estimation error by 2.456\,Nm on average.

\begin{figure*}[!ht]
\centering
\vspace*{0.0cm}
\includegraphics[width=1.0\linewidth]{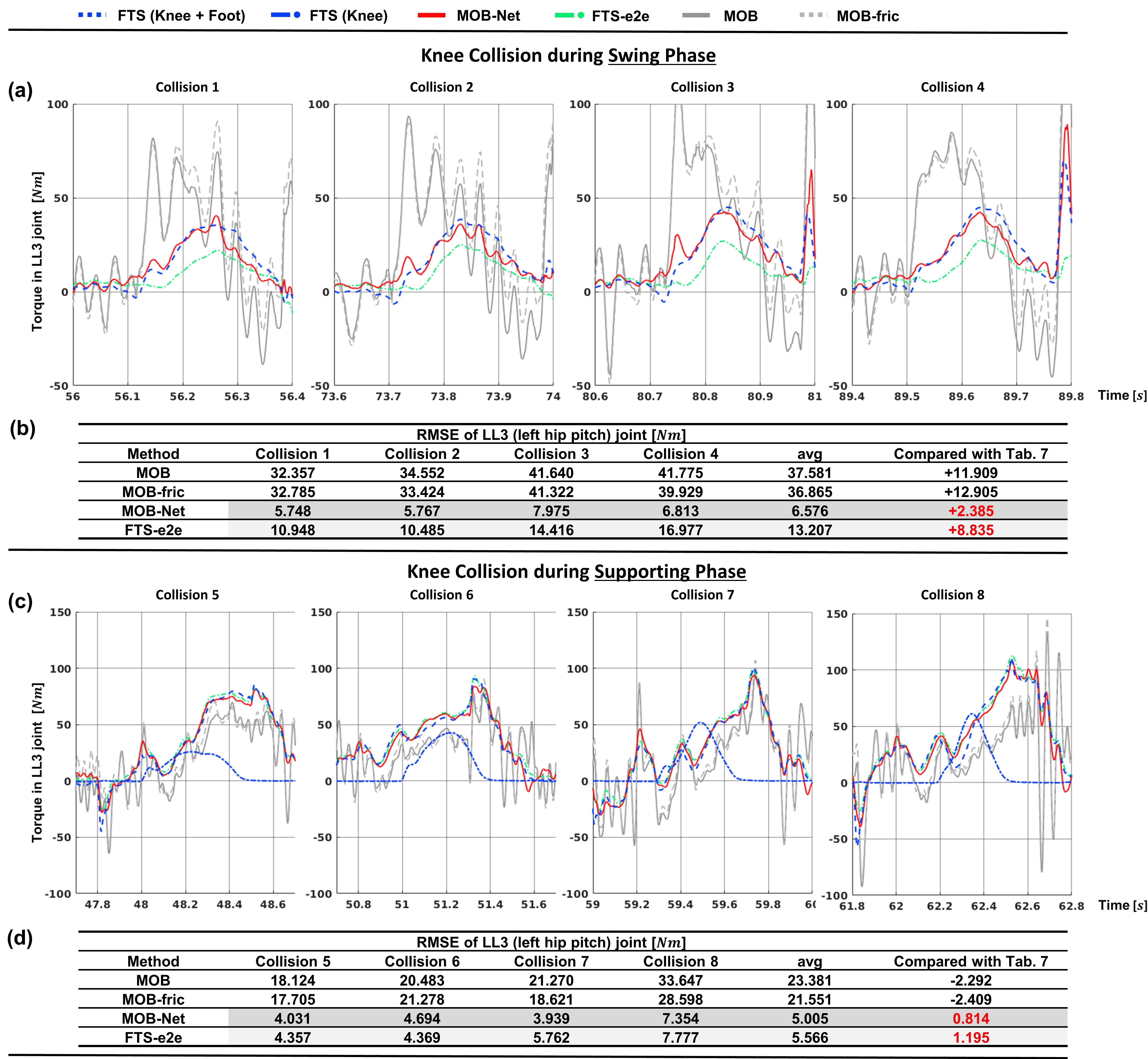}
\caption{\textbf{Comparison of external joint torque estimation performance for the knee collisions.} Collisions are made during both the swing and the supporting phase. (a), (c) The measured torque and the estimated torque of each method in the LL3 joint for 4 collision cases during the swing and supporting phase, respectively. (b), (d) RMSE of the estimation for each method and collision case. The changes in the average RMSE compared to the error of the random walking test are displayed in the last column.}
\label{fig/exp_knee_collision_plot}
\end{figure*}

\begin{figure}[!ht]
\centering
\vspace*{0.0cm}
\includegraphics[width=0.8\linewidth]{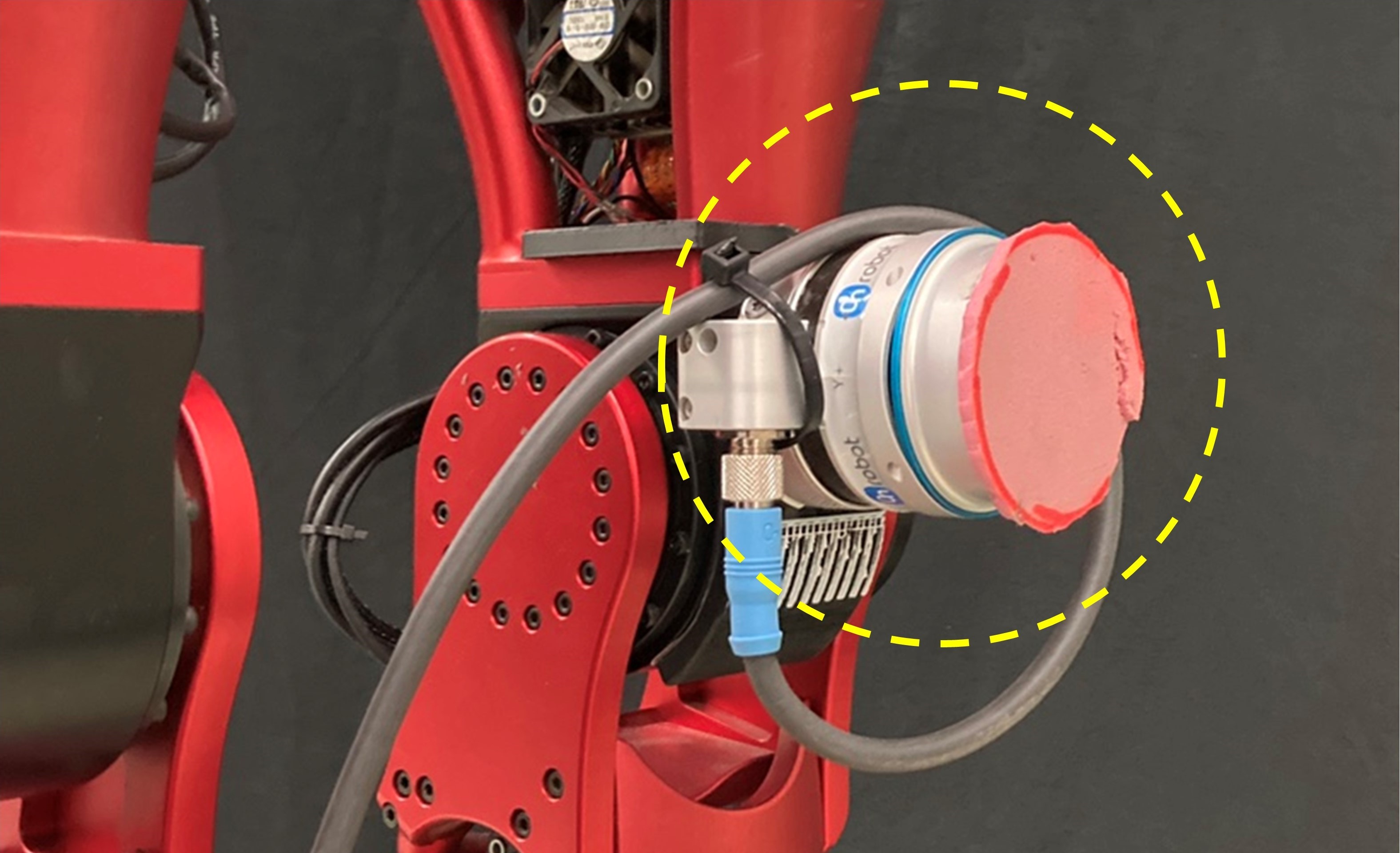}
\caption{\textbf{Experimental setup for the knee collision.} An FTS attached on the left knee link to measure the external torque due to knee collisions.}
\label{fig/knee_FTS_exp_setting}
\end{figure}

\begin{figure}[!ht]
\centering
\vspace*{0.0cm}
\includegraphics[width=1.0\linewidth]{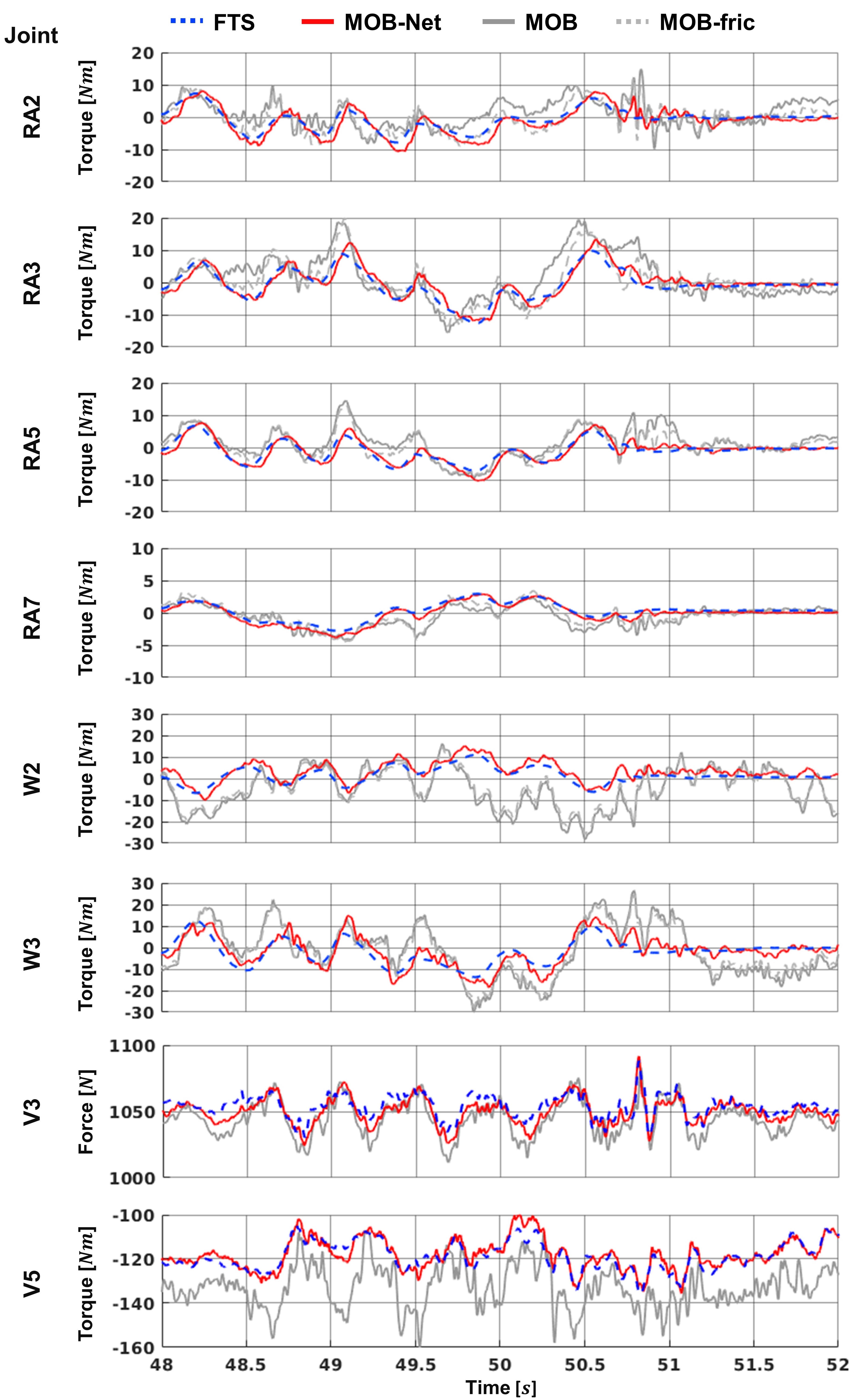}
\caption{\textbf{Real robot experimental result of the partial upper body joints and the virtual joints for the \emph{External wrench on the hand} data set.} A random external wrench is applied on the left hand.}
\label{fig/tocabi_0512_ext_wrench_upper_body_joint}
\end{figure}

To test the external joint torque estimation performance for the unseen data in the real humanoid robot, an external force is applied to the left knee link when the left leg is in the swing phase, while the robot is walking in place with a step duration of 0.7\,s.
The training data does not include such collision data on the swing leg.
To measure the external joint torque that occurred from the knee collision accurately, an FTS is attached to the left knee link of the robot using 3D-printed parts as shown in Figure~\ref{fig/knee_FTS_exp_setting}, and the external joint torque from the knee collision is calculated using the contact Jacobian of the FTS on the knee.
The experimental results of the four collision cases during the swing phase are plotted in Figure~\ref{fig/exp_knee_collision_plot}~(a).
These plots show the measured and estimated external joint torque in the LL3 (left hip pitch) joint which is mainly affected by the front knee collision.
The dotted blue line indicates the measured external joint torque using the FTS on the knee and the foot link to measure the collision force on the knee and the ground reaction force on the foot; Although the knee collision starts during the swing phase, it can last after the foot lands.

% upper body and virtual joint estimation results
\begin{table*}[!t]
\centering
\caption{Real robot experimental result of the upper body joints and the virtual joints for the two test data sets: \emph{Collision free motion} and \emph{External wrench on the hand}. RMSE of the estimated external joint torque or force are summarized according to the estimation method, and the data set.}
\label{table/realrobot_upper_body_rmse}
\resizebox{\textwidth}{!}{%
\begin{tabular}{@{}ccccccccccccccccccccc@{}}
\toprule
 &
  \multicolumn{12}{c|}{\textbf{Upper body joint}} &
  \multicolumn{8}{c}{\textbf{Virtual joint}} \\ \midrule
  \textbf{Method} &
  \textbf{W1} &
  \textbf{W2} &
  \textbf{W3} &
  \textbf{RA1} &
  \textbf{RA2} &
  \textbf{RA3} &
  \textbf{RA4} &
  \textbf{RA5} &
  \textbf{RA6} &
  \textbf{RA7} &
  \textbf{RA8} &
  \multicolumn{1}{c|}{\textbf{\begin{tabular}[c]{@{}c@{}}Avg\\ (W+RA)\end{tabular}}} &
  \textbf{\begin{tabular}[c]{@{}c@{}}V1\\ \text{[\emph{N}]}\end{tabular}} &
  \textbf{\begin{tabular}[c]{@{}c@{}}V2\\ \text{[\emph{N}]}\end{tabular}} &
  \textbf{\begin{tabular}[c]{@{}c@{}}V3\\ \text{[\emph{N}]}\end{tabular}} &
  \textbf{\begin{tabular}[c]{@{}c@{}}Avg\\(V lin)\end{tabular}} &
  \textbf{V4} &
  \textbf{V5} &
  \textbf{V6} &
  \textbf{\begin{tabular}[c]{@{}c@{}}Avg\\(V ang)\end{tabular}} \\ \midrule
\multicolumn{21}{c}{\textbf{Data set1: Collision free motion}} \\ \midrule
\textbf{MOB} &
  14.44 &
  14.35 &
  10.55 &
  3.14 &
  3.73 &
  4.31 &
  3.45 &
  3.27 &
  1.94 &
  1.31 &
  1.18 &
  \multicolumn{1}{c|}{\textbf{5.61}} &
  36.69 &
  55.31 &
  41.41 &
  \textbf{44.47} &
  44.32 &
  30.04 &
  10.64 &
  \textbf{28.34} \\ \midrule
\textbf{MOB-fric} &
  12.86 &
  13.38 &
  9.18 &
  2.68 &
  3.72 &
  3.12 &
  1.94 &
  4.03 &
  1.93 &
  1.40 &
  1.24 &
  \multicolumn{1}{c|}{\textbf{5.04}} &
  - &
  - &
  - &
  - &
  - &
  - &
  - &
  - \\ \midrule
\textbf{MOB-Net} &
  \cellcolor[HTML]{C0C0C0}5.40 &
  \cellcolor[HTML]{C0C0C0}2.75 &
  \cellcolor[HTML]{C0C0C0}4.46 &
  \cellcolor[HTML]{C0C0C0}1.06 &
  \cellcolor[HTML]{C0C0C0}1.07 &
  \cellcolor[HTML]{C0C0C0}1.09 &
  \cellcolor[HTML]{C0C0C0}0.80 &
  \cellcolor[HTML]{C0C0C0}0.65 &
  \cellcolor[HTML]{C0C0C0}0.37 &
  \cellcolor[HTML]{C0C0C0}0.19 &
  \cellcolor[HTML]{C0C0C0}0.27 &
  \multicolumn{1}{c|}{\cellcolor[HTML]{C0C0C0}\textbf{1.65}} &
  \cellcolor[HTML]{C0C0C0}14.18 &
  \cellcolor[HTML]{C0C0C0}17.12 &
  \cellcolor[HTML]{C0C0C0}22.58 &
  \cellcolor[HTML]{C0C0C0}\textbf{17.96} &
  \cellcolor[HTML]{C0C0C0}11.62 &
  \cellcolor[HTML]{C0C0C0}11.91 &
  \cellcolor[HTML]{C0C0C0}4.73 &
  \cellcolor[HTML]{C0C0C0}\textbf{9.42} \\ \midrule
\multicolumn{21}{c}{\textbf{Data set2: External wrench on the hand}} \\ \midrule
\textbf{MOB} &
  16.02 &
  12.93 &
  10.06 &
  4.58 &
  3.51 &
  4.82 &
  3.93 &
  3.94 &
  1.65 &
  1.10 &
  1.38 &
  \multicolumn{1}{c|}{\textbf{5.81}} &
  14.24 &
  20.11 &
  12.32 &
  \textbf{15.56} &
  12.42 &
  18.45 &
  10.44 &
  \textbf{13.77} \\ \midrule
\textbf{MOB-fric} &
  14.24 &
  11.08 &
  8.46 &
  3.96 &
  2.42 &
  3.17 &
  2.35 &
  3.04 &
  2.11 &
  0.96 &
  1.24 &
  \multicolumn{1}{c|}{\textbf{4.82}} &
  - &
  - &
  - &
  - &
  - &
  - &
  - &
  - \\ \midrule
\textbf{MOB-Net} &
  \cellcolor[HTML]{C0C0C0}4.26 &
  \cellcolor[HTML]{C0C0C0}3.77 &
  \cellcolor[HTML]{C0C0C0}3.18 &
  \cellcolor[HTML]{C0C0C0}2.39 &
  \cellcolor[HTML]{C0C0C0}1.71 &
  \cellcolor[HTML]{C0C0C0}1.47 &
  \cellcolor[HTML]{C0C0C0}1.36 &
  \cellcolor[HTML]{C0C0C0}1.03 &
  \cellcolor[HTML]{C0C0C0}0.31 &
  \cellcolor[HTML]{C0C0C0}0.49 &
  \cellcolor[HTML]{C0C0C0}0.37 &
  \multicolumn{1}{c|}{\cellcolor[HTML]{C0C0C0}\textbf{1.85}} &
  \cellcolor[HTML]{C0C0C0}5.94 &
  \cellcolor[HTML]{C0C0C0}7.79 &
  \cellcolor[HTML]{C0C0C0}7.62 &
  \cellcolor[HTML]{C0C0C0}\textbf{7.12} &
  \cellcolor[HTML]{C0C0C0}4.45 &
  \cellcolor[HTML]{C0C0C0}3.54 &
  \cellcolor[HTML]{C0C0C0}3.67 &
  \cellcolor[HTML]{C0C0C0}\textbf{3.89} \\ \bottomrule
\end{tabular}%
}
\end{table*}

As displayed in Figure~\ref{fig/exp_knee_collision_plot}~(a), MOB-Net shows the best estimation performance among the four methods.
Conversely, FTS-e2e estimates the external joint torque with delayed and lower values resulting in large errors.
The average errors of the estimation for the four cases are summarized in Figure~\ref{fig/exp_knee_collision_plot}~(b). 
The average error of MOB-Net for the knee collisions is 6.576\,Nm in the LL3 joint, which is 2.385\,Nm larger than the errors of the LL3 joint in the random walking test set (in-distribution data).
However, the average error of FTS-e2e significantly increases to 13.207\,Nm, which is 8.835\,Nm larger than the RMSE of the LL3 joint in the random walking test set.
This result validates the robust performance of MOB-Net for the knee collision that is unseen data, whereas the estimation of FTS-e2e deteriorates for the unseen data, which is similar to the result of hip pitch disturbance in the simulation. 
This is the well-known limitation of the deep learning method, but MOB-Net reduces the negative effect of the learning method for the unseen data by combining both the model-based observer and the deep learning together.

The knee collision experiment during the supporting phase is also performed to investigate the external torque estimation performance when the support leg is disturbed.
The left knee is pushed when it is in the supporting phase and the external torques of the left leg are measured using both FTS on the knee and the foot.
Figure~\ref{fig/exp_knee_collision_plot}~(c) displays the measured external torque and the estimation results of the LL3 joint for the four collision cases. 
The estimated torque of MOB-Net and FTS-e2e follows the measured external torque tightly even if the support leg is pushed. 
However, the two model-based methods still show large errors. 
Figure~\ref{fig/exp_knee_collision_plot}~(d) summarizes the estimation errors of each method. 
The estimation errors of MOB-Net and FTS-e2e are approximately 1\,Nm larger than the random walking test results of the LL3 joint in Table~\ref{table/RMSE of external torque estimation of left leg}.
The difference in errors between MOB-Net and FTS-e2 for the supporting knee collisions is smaller than the difference for the swing knee collisions. 
This is because the training data includes various disturbance patterns for the supporting leg obtained from the random walking data on uneven terrain. Therefore, the supporting knee collision can be regarded as in-distribution data, but the swing knee collision is not covered by the training data (out-of-distribution data).

The upper body and virtual joints are tested for two test data sets: \emph{Collision free motion} and \emph{External wrench on the hand}.  
\emph{Collision free motion} contains random upper body motion while the robot is walking randomly (in-distribution data).
In \emph{External wrench on the hand}, the experimenter applies a random external wrench to the right hand of the robot while the upper body is moving randomly.
All the test results are summarized in Table~\ref{table/realrobot_upper_body_rmse}. 
MOB-Net results in much smaller errors than the model-based methods across all joints and test sets. Although MOB-Net is trained only using collision free motion data, MOB-Net can estimate external torque from the external wrench on the hand as validated in the second data set (External wrench on the hand).
Figure~\ref{fig/tocabi_0512_ext_wrench_upper_body_joint} shows the measured and estimated external torques in the upper body and the virtual joints. 
The estimated torque from MOB-Net tracks the measured external torque by FTS whereas MOB residual shows large errors. 

\begin{table*}[!h]
\centering
\caption{The number of collision detection successes and collision delay of the left leg for detection Method, collision link, and impact direction}
\label{table/collision detection result}
\resizebox{\textwidth}{!}{%
\begin{tabular}{@{}ccccccccccccccc@{}}
\toprule
\multicolumn{1}{c|}{\textbf{}} &
  \multicolumn{7}{c|}{\textbf{Collision Detection   Success}} &
  \multicolumn{7}{c}{\textbf{Collision Detection   Delay [\emph{ms}]}} \\ \midrule
\multicolumn{1}{c|}{} &
  \multicolumn{2}{c}{\textbf{Foot}} &
  \multicolumn{2}{c}{\textbf{Ankle}} &
  \multicolumn{2}{c}{\textbf{Knee}} &
  \multicolumn{1}{c|}{} &
  \multicolumn{2}{c}{\textbf{Foot}} &
  \multicolumn{2}{c}{\textbf{Ankle}} &
  \multicolumn{2}{c}{\textbf{Knee}} &
   \\
\multicolumn{1}{c|}{\multirow{-2}{*}{\textbf{Method}}} &
  \textbf{Front} &
  \textbf{Side} &
  \textbf{Front} &
  \textbf{Side} &
  \textbf{Front} &
  \textbf{Side} &
  \multicolumn{1}{c|}{\multirow{-2}{*}{\textbf{Total}}} &
  \textbf{Front} &
  \textbf{Side} &
  \textbf{Front} &
  \textbf{Side} &
  \textbf{Front} &
  \textbf{Side} &
  \multirow{-2}{*}{\textbf{Avg}} \\ \midrule
\multicolumn{1}{c|}{MOB-Net-OR} &
  10 &
  10 &
  10 &
  10 &
  10 &
  10 &
  \multicolumn{1}{c|}{\cellcolor[HTML]{C0C0C0}60/60} &
  9.60 &
  6.30 &
  17.20 &
  14.60 &
  15.50 &
  8.20 &
  \cellcolor[HTML]{C0C0C0}11.90 \\
\multicolumn{1}{c|}{MOB-Net-mean} &
  10 &
  10 &
  10 &
  5 &
  8 &
  10 &
  \multicolumn{1}{c|}{53/60} &
  10.20 &
  6.30 &
  17.20 &
  14.00 &
  19.25 &
  8.20 &
  \cellcolor[HTML]{EFEFEF}12.53 \\
\multicolumn{1}{c|}{MOB-Net-sigma} &
  10 &
  10 &
  9 &
  10 &
  9 &
  10 &
  \multicolumn{1}{c|}{\cellcolor[HTML]{EFEFEF}58/60} &
  10.10 &
  6.60 &
  20.00 &
  15.20 &
  19.67 &
  10.50 &
  13.68 \\
\multicolumn{1}{c|}{FTS-e2e-mean} &
  10 &
  10 &
  10 &
  4 &
  9 &
  10 &
  \multicolumn{1}{c|}{53/60} &
  9.50 &
  7.70 &
  34.00 &
  11.50 &
  95.67 &
  11.60 &
  28.33 \\
\multicolumn{1}{c|}{FTS-e2e-sigma} &
  1 &
  0 &
  0 &
  0 &
  0 &
  0 &
  \multicolumn{1}{c|}{1/60} &
  261.00 &
  - &
  - &
  - &
  - &
  - &
  261.00 \\
\multicolumn{1}{c|}{MOB} &
  4 &
  10 &
  5 &
  10 &
  1 &
  9 &
  \multicolumn{1}{c|}{39/60} &
  21.50 &
  18.50 &
  19.00 &
  17.10 &
  31.00 &
  14.11 &
  20.20 \\
\multicolumn{1}{c|}{MOB-fric} &
  8 &
  10 &
  10 &
  10 &
  2 &
  10 &
  \multicolumn{1}{c|}{50/60} &
  17.63 &
  10.50 &
  17.70 &
  19.90 &
  29.50 &
  10.20 &
  17.57 \\
\multicolumn{1}{c|}{MOB-fric-BPF} &
  3 &
  1 &
  7 &
  1 &
  0 &
  1 &
  \multicolumn{1}{c|}{13/60} &
  11.67 &
  20.00 &
  13.57 &
  17.00 &
  X &
  19.00 &
  16.25 \\ \bottomrule
\end{tabular}%
}
\end{table*}

% In summary, MOB-Net has superior estimation performance than the two model-based methods for both in-distribution and out-of-distribution data and shows similar performance to FTS-e2e for the in-distribution data. However, for the out-of-distribution data (swing knee collisions), MOB-Net shows more robust estimation results than FTS-e2e.

\subsection{Comparison of Collision Detection Performance}
\label{Subsection/Experiment/Comparison of Collision Detection Performance}
In this experiment, the collision detection performance is compared with various methods to validate the superiority of the proposed method. 
Similar to the previous section, FTS-e2e, MOB, and MOB-fric are selected as comparison methods.
Collision is detected when the collision signal (normally estimated external joint torque) is over the threshold value in each joint, and collision detection is only performed for unexpected collision during the swing phase except for repetitive foot contacts while walking.
The collision threshold values were conservatively set to be 10 percent higher than the maximum error occurring during the robot's normal walking by each estimation method to compare each method fairly and prevent the occurrence of false positives which can be considered as zero.
The unexpected collision among repetitive foot contacts is classified based on several heuristic criteria.
Continuous filtering of 5\,ms time horizon and low pass filtering of 15\,Hz cut-off frequency are used for more robust collision detection performance.
The collision detection of the data-driven methods is performed using two collision signals, $\hat{\bm{\tau}}$ and $\hat{\bm{\sigma}}$ that are the mean and standard deviation in the network outputs.
Thus, three variants of MOB-Net are tested; \emph{MOB-Net-OR} uses both $\hat{\bm{\tau}}_{e}$ and $\hat{\bm{\sigma}}_{u}$ for collision detection using OR logic operation, \emph{MOB-Net-mean} uses only $\hat{\bm{\tau}}_{e}$, and \emph{MOB-Net-sigma} uses only $\hat{\bm{\sigma}}_{u}$ for collision detection.
Similarly, two variants of FTS-e2e are tested: \emph{FTS-e2e-mean} and \emph{FTS-e2e-sigma}.
\emph{FTS-e2e-OR} is not tested because \emph{FTS-e2e-sigma} shows poor collision detection performance.
A model-based method, \emph{MOB-fric-BPF}, is implemented additionally using band-pass-filter to suppress the error caused by modeling error as in \cite{song2013collision, van2022collision} in addition to the MOB and MOB-fric. The cutoff frequency of the band-pass filter is chosen as $[2, 15]$\,Hz.

\begin{figure}[!t]
\centering
\vspace*{0.0cm}
\includegraphics[width=1.0\linewidth]{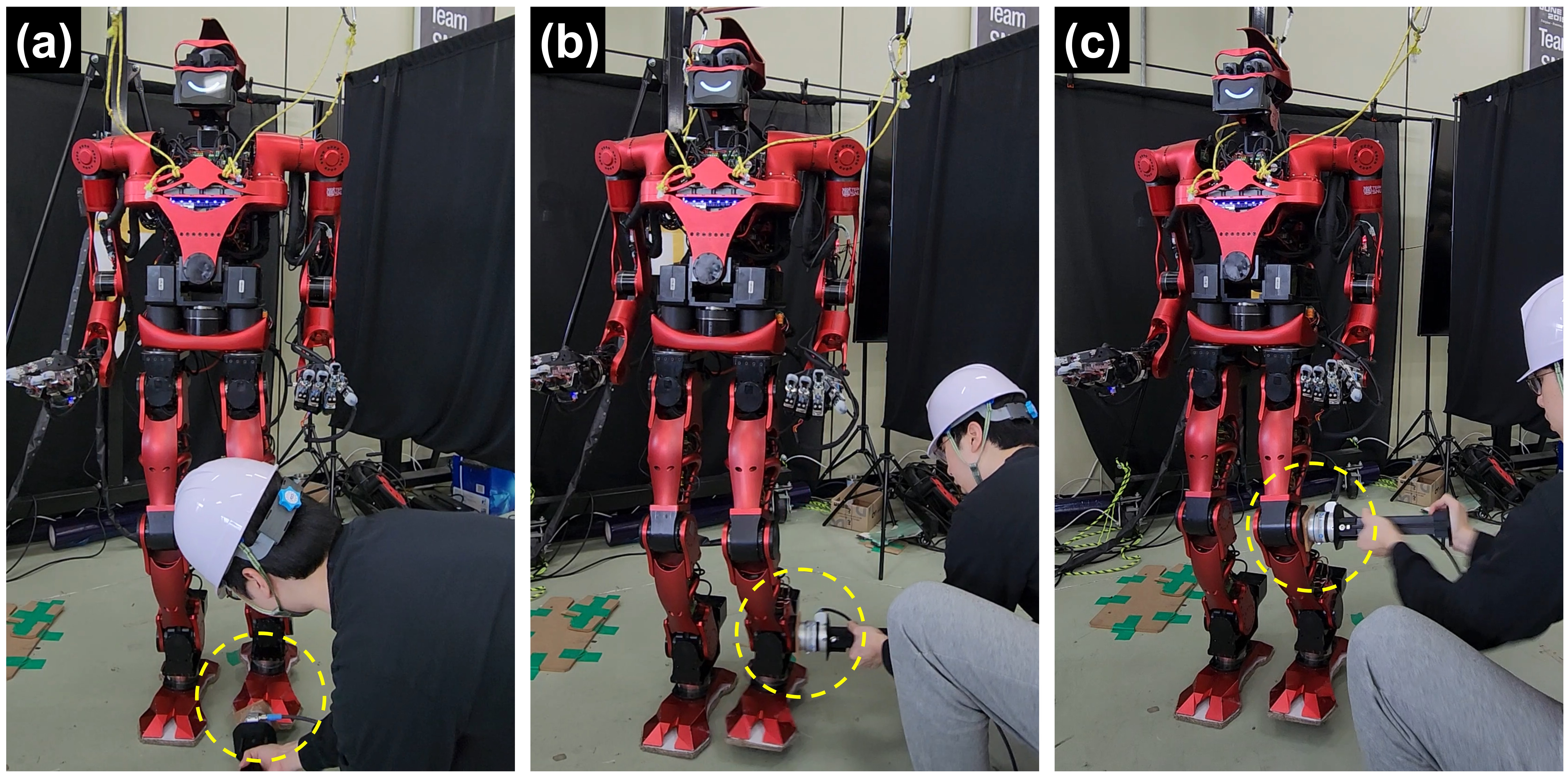}
\caption{\textbf{Example snapshots of collision detection experiment.} (a) A foot-front collision. (b) An ankle-side collision. (c) A knee-side collision.}
\label{fig/collision_detection_examples}
\end{figure}

\begin{figure}[!t]
\centering
\vspace*{0.0cm}
\includegraphics[width=1.0\linewidth]{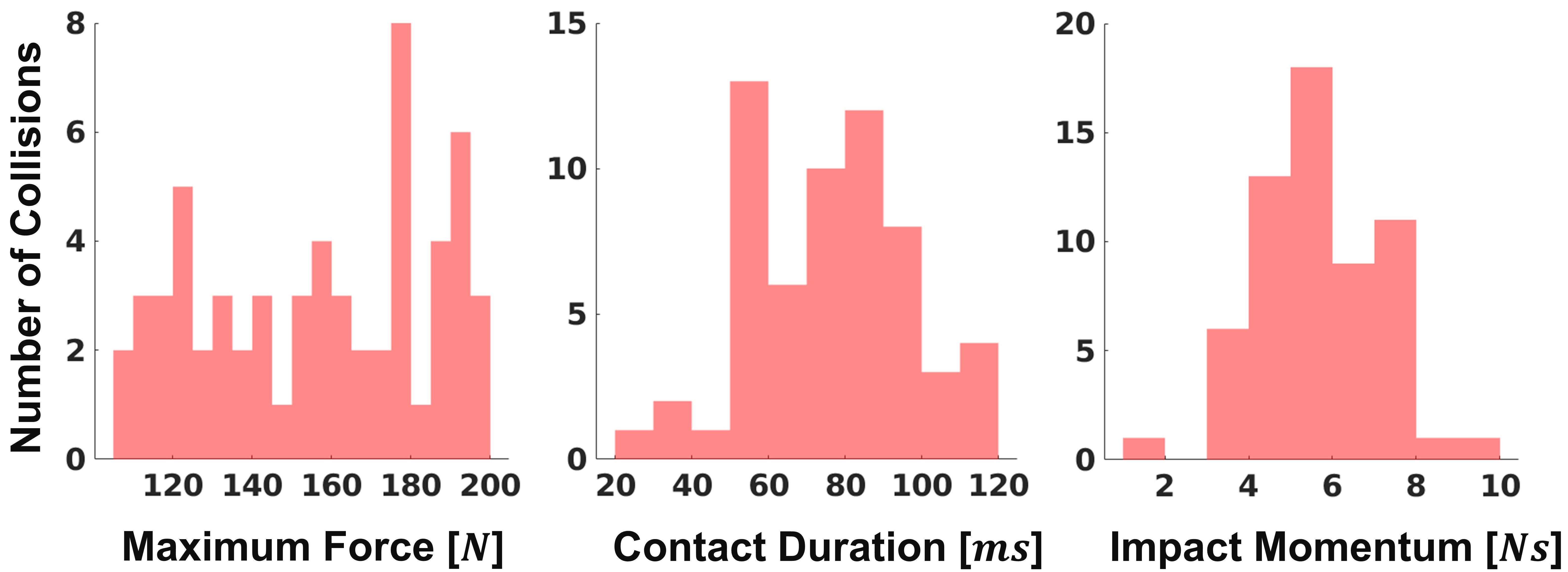}
\caption{\textbf{\textcolor{black}{Histograms of 60 collisions used in collision detection experiment.}}}
\label{fig/collision_detection_impact_histogram}
\end{figure}

The collisions were made by an experimenter using a collision tool equipped with FTS and a rudder pad on the collision side while the robot was walking in place. 10 collisions occurred at each collision point of the left leg with two different directions (front and side), and three different links (foot, ankle, and knee) resulting in a total of 60 collisions. Figure~\ref{fig/collision_detection_examples} shows three examples of the collision detection experiment. 
The maximum collision force ranges in [100, 200]\,N approximately and the average contact duration is 78\,ms as shown in \textcolor{black}{Figure~\ref{fig/collision_detection_impact_histogram}}. The average and maximum impact momentum of all collisions is 7\,Nm and 10\,Nm, respectively. \textcolor{black}{}

Table \ref{table/collision detection result} summarizes all the collision detection results.
MOB-Net-OR shows the best performance without detection failure and has the fastest detection time. 
Although both MOB-Net-mean and MOB-Net-sigma failed to detect collisions several times, each method complements the other's failure by combining both signals in MOB-Net-OR. 
FTS-e2e-mean resulted in similar detection accuracy to MOB-Net-mean for all collisions and similar detection delay for the foot collisions. However, the detection delay of FTS-e2e-mean increased for ankle and knee collisions which is the unseen data resulting in more than 6 times longer detection delay for the Knee-Front compared to the delay of MOB-Net-OR. 
This is because the external torque estimation of FTS-e2e shows delayed signals and its error increases for the knee impact (unseen data) as validated in Section \ref{Subsection/Experiment/Comparison of External Torque Estimation Performance in Real Humanoid Robot}.
FTS-e2e-sigma shows poor performance detecting a single foot-collision which is even too slow to be used for a safe collision reaction. 
MOB fails to detect 20 front collisions among 30 collisions whereas successful for the side collisions. The friction model improves the accuracy of the collision detection resulting in 50 successes over 60 collisions and decreases the collision detection delay from MOB. The BPF, however, was not helpful for collision detection unlike the result in \cite{van2022collision} because the modeling error is not in the low-frequency domain during dynamic motions such as locomotion which is different from the stationary task performed in the other studies.

\begin{figure}[!t]
\centering
\vspace*{0.0cm}
\includegraphics[width=1.0\linewidth]{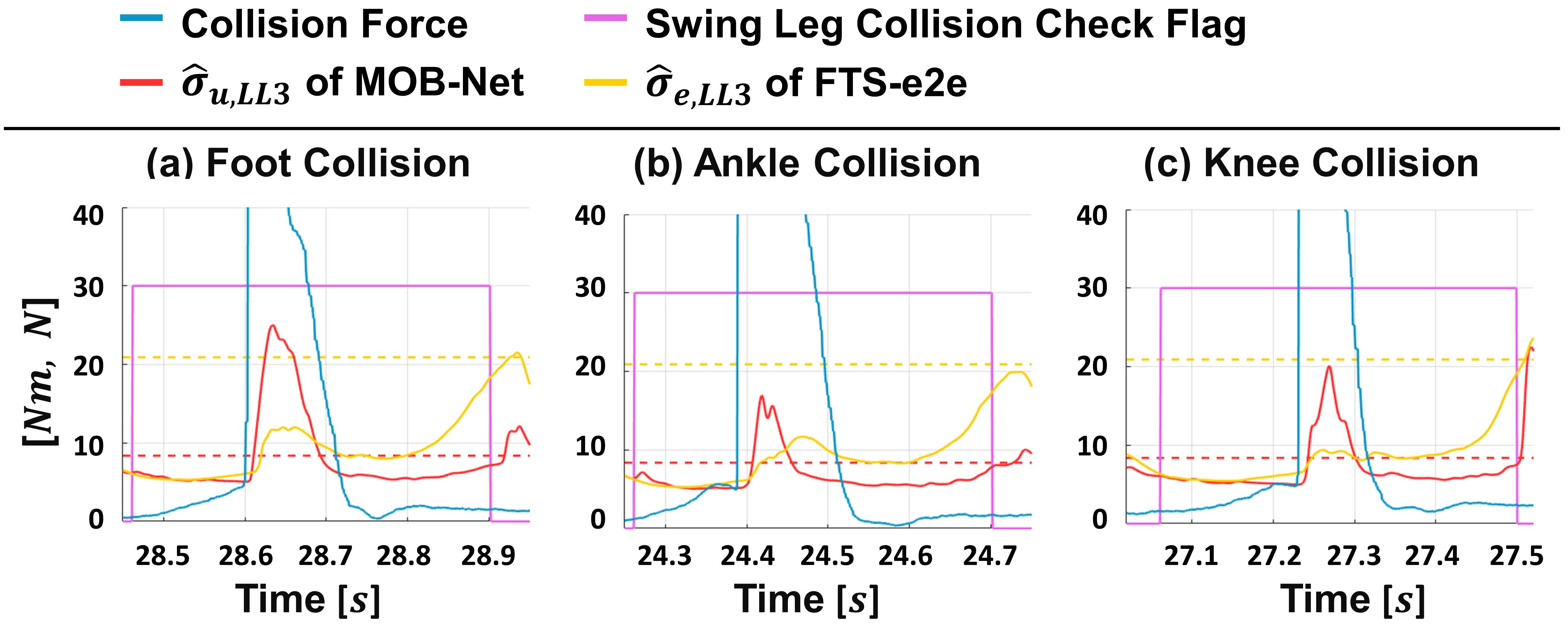}
\caption{\textbf{Examples of the estimated standard deviation of MOB-Net and FTS-e2e in LL3 joint for three different front collisions.} The dotted line shows the collision detection threshold of each method. (a) A foot collision. (b) An ankle collision. (c) A knee collision.}
\label{fig/estimated_variance_for_collision_detection}
\end{figure}

Figure~\ref{fig/estimated_variance_for_collision_detection} displays the behavior of the estimated standard deviation of two data-driven methods for the swing leg collisions. The estimated standard deviations of MOB-Net (red line) and FTS-e2e (yellow line) in the LL3 joint are plotted for the three front collisions in Figure~\ref{fig/estimated_variance_for_collision_detection}. The collision force (light blue line) is displayed to show the timing of the collision. The purple line indicates the swing leg collision check flag which is determined from the walking planner and heuristic conditions to detect unexpected collisions on the swing leg and to ignore repetitive ground contacts of the foot. The dotted line indicates the collision threshold for each method. As shown in the figure, $\hat{\bm{\sigma}}_{u,LL3}$ of MOB-Net surges according to the collision, which results in fast collision detection. Conversely, $\hat{\bm{\sigma}}_{e,LL3}$ of FTS-e2e fluctuates slightly but under the threshold value, failing to detect collisions. 

\subsection{\textcolor{black}{Collision Forces with Fixed Obstacles using Collision-Handling Strategy}}
\label{Subsection/Experiment/Sensitivity of Collision Detection and Reaction}

\textcolor{black}{
While MOB-Net-OR detected all 60 collision instances in the previous experiment (Section~\ref{Subsection/Experiment/Comparison of Collision Detection Performance}), collision detection in joint space varies in sensitivity depending on the contact point and the robot's configuration, making it challenging to measure the collision detection sensitivity and false negatives of the proposed method for various situations. Instead, an indirect estimation can be made through the collision detection threshold values which are summarized in Appendix B, Table~\ref{table/collision detection threshold}. Moreover, Figure~\ref{fig/collision_detection_impact_histogram} enables us to understand the maximum force and impact of detectable external forces, but more experiments are needed to determine the detection sensitivity for smaller forces.}

\begin{figure}[!t]
\centering
\vspace*{0.0cm}
\includegraphics[width=0.9\linewidth]{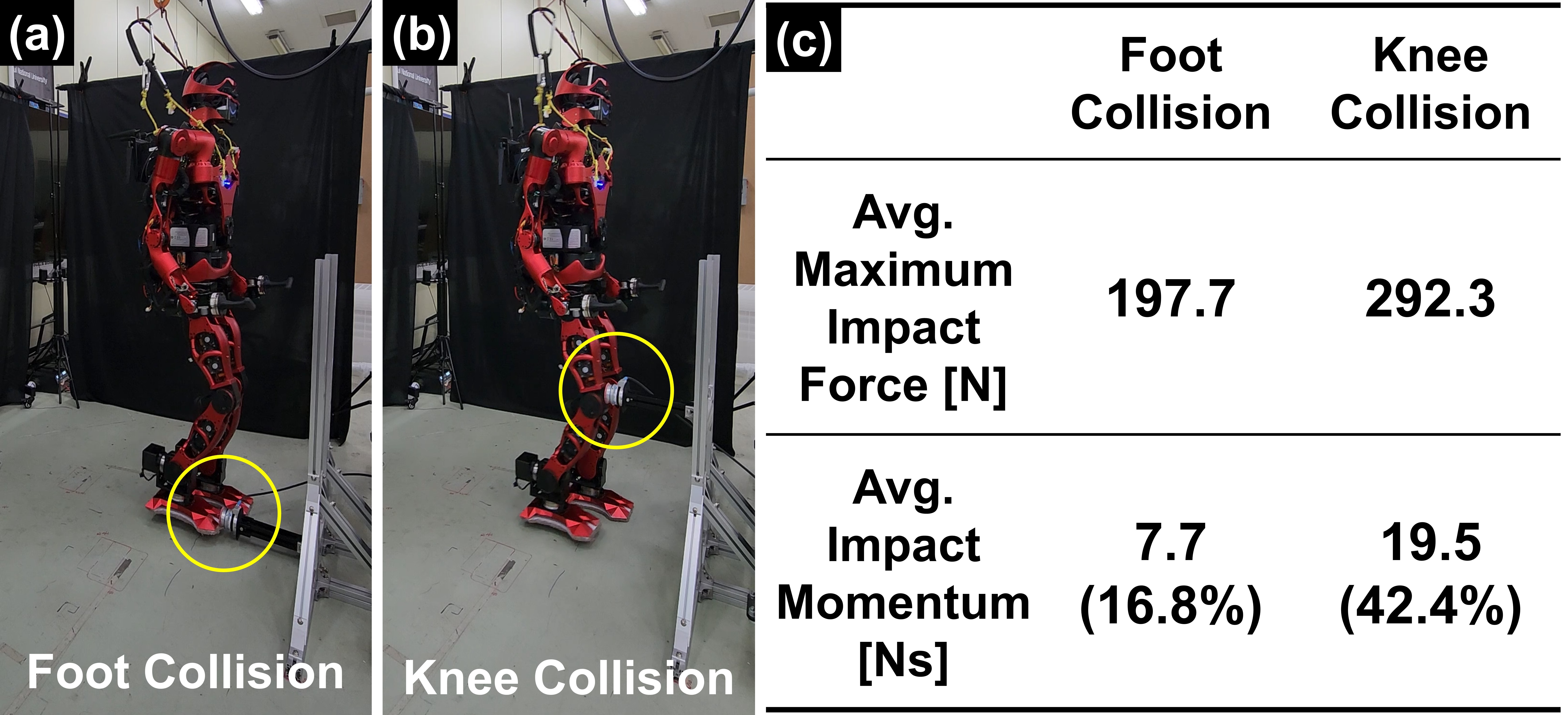}
\caption{\textcolor{black}{\textbf{Experimental results of collision detection and reaction with maximum impact force and impact momentum.} (a) The collision on the right foot. (b) The collision on the right knee. (c) The average maximum impact force and impact momentum for foot and knee collisions over five runs.}}
\label{fig/collision_detection_and_reaction_sensitivity}
\end{figure}

\textcolor{black}{
However, the purpose of the external torque estimation using MOB-Net and the collision-handling strategy proposed in this study is not to detect extremely sensitive contacts, but rather to prevent the humanoid robot from falling over due to collisions with fixed or heavy obstacles, thereby preventing further accidents.
In such situations, the collision force increases until the robot detects the collisions and appropriately reacts, which means the collisions are eventually detected.  
Therefore, it is rather important to measure how much the maximum impact force and impact momentum are generated using MOB-Net-OR and the proposed collision-handling pipeline because the maximum force and impact momentum indirectly represent how fast our method reacts to the collisions. Hence, in this experiment, as shown in Figure~\ref{fig/collision_detection_and_reaction_sensitivity}, we measured the maximum impact force and impact momentum generated during the collision handling process for the foot and knee collisions with a fixed obstacle while the robot is walking forward. The contact force was measured using the experimental structure we constructed and the FTS (Force Torque Sensor) mounted at the collision link, as indicated by the yellow circle in Figure~\ref{fig/collision_detection_and_reaction_sensitivity}~(a) and (b). The collision-handling pipeline used in this experiment is introduced in Section~\ref{Section/Application}.
}

\textcolor{black}{
Figure~\ref{fig/collision_detection_and_reaction_sensitivity}~(c) presents the averaged values of the maximum impact force and impact momentum from five experimental runs. In the case of foot collisions, the average maximum impact force between the robot and a fixed obstacle was 197.7\,N, with an impact momentum of 7.7\,Ns. It is noteworthy that this impact momentum represents only 16.8\% of the maximum endurable momentum in the -x direction of the latest controller developed by our research group introduced in \cite{kim2023model}, indicating that the value is sufficiently small for the robot to maintain balance. In knee collisions, the values were significantly higher, averaging 298.3\,N for the maximum impact force, and 19.5\,Ns for the impact momentum. This difference can be attributed to the fact that for the hip joint, the moment arm of foot contact is larger compared to one of knee contact. Consequently, even with the same impact force, the external torque applied to the hip joint is measured to be greater in foot collisions, allowing for more sensitive detection and quicker collision reaction. Nevertheless, the average impact momentum of 19.5\,Ns in the knee collisions also falls within the robot's maximum endurable impact momentum at 42.4\%, enabling the robot to maintain balance and react effectively to collisions. It is also important to note that in all collision experiments, the robot was able to detect all collisions with static objects using the proposed method.
}

\section{Ablation Study}
In this section, several ablation studies are provided to validate the network architecture of MOB-Net and the random torque exploration method.
\label{Section/Ablation Study}

\subsection{Network Structure Comparison}
\label{Subsection/Ablation Study/Network Structure Comparison}

\begin{figure}[ht]
\centering
\vspace*{0.0cm}
\includegraphics[width=1.0\linewidth]{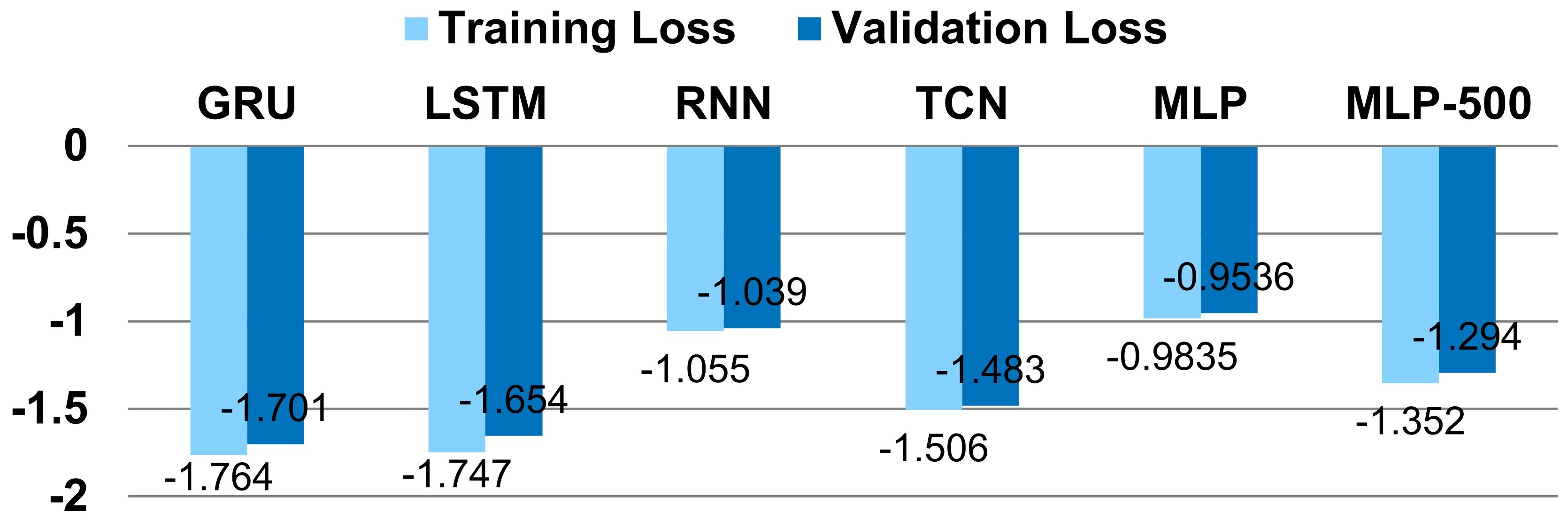}
\caption{\textbf{The training and the validation loss according to the kind of neural network.} GRU shows the lowest losses.}
\label{fig/abliation_networks}
\end{figure}

Five different kinds of neural networks are compared to validate the use of the GRU module including two other RNN-series networks. 
For this experiment, only MOB-Net for the left leg ($\bm{f}_{\theta}^{LL}$) is compared using the random walking data set, but other limbs show similar results. 
The five comparison networks are summarized below.
\begin{itemize}
    \item LSTM (Long Short-Term Memory), \cite{hochreiter1997long}: The GRU module in MOB-Net is replaced by an LSTM module. The network structure is LSTM(30, 128)-Linear(128, 12) making the learning parameter size (83468) similar to the parameter size of MOB-Net (83712).
    \item RNN: The GRU module in MOB-Net is replaced by a vanilla RNN. The network structure is RNN(30, 268)-Linear(268, 12). The network size is 83628.
    \item TCN (Temporal Convolutional Network), \cite{bai2018empirical}: TCN is a variant of 1D-CNN using casual and dilated convolutions, and is specialized in sequence modeling. TCN with dilation factors d = 2, 4, 8 and filter size k = 3 is used following the linear output layer. The hidden size is 32 resulting in a model size of 83956 and a dropout of 0.2 is used. 
    \item MLP: A simple 2-layer MLP network is used with a hidden size of 55, i.e., MLP(150, 55, 12). The model size is 83227.
    \item MLP-500: A larger 2-layer MLP network with a hidden size of 500 is also tested resulting in the model size of 756512 which is approximately 9 times larger than the other networks.
\end{itemize}

\begin{figure}[!t]
\centering
\vspace*{0.0cm}
\includegraphics[width=1.0\linewidth]{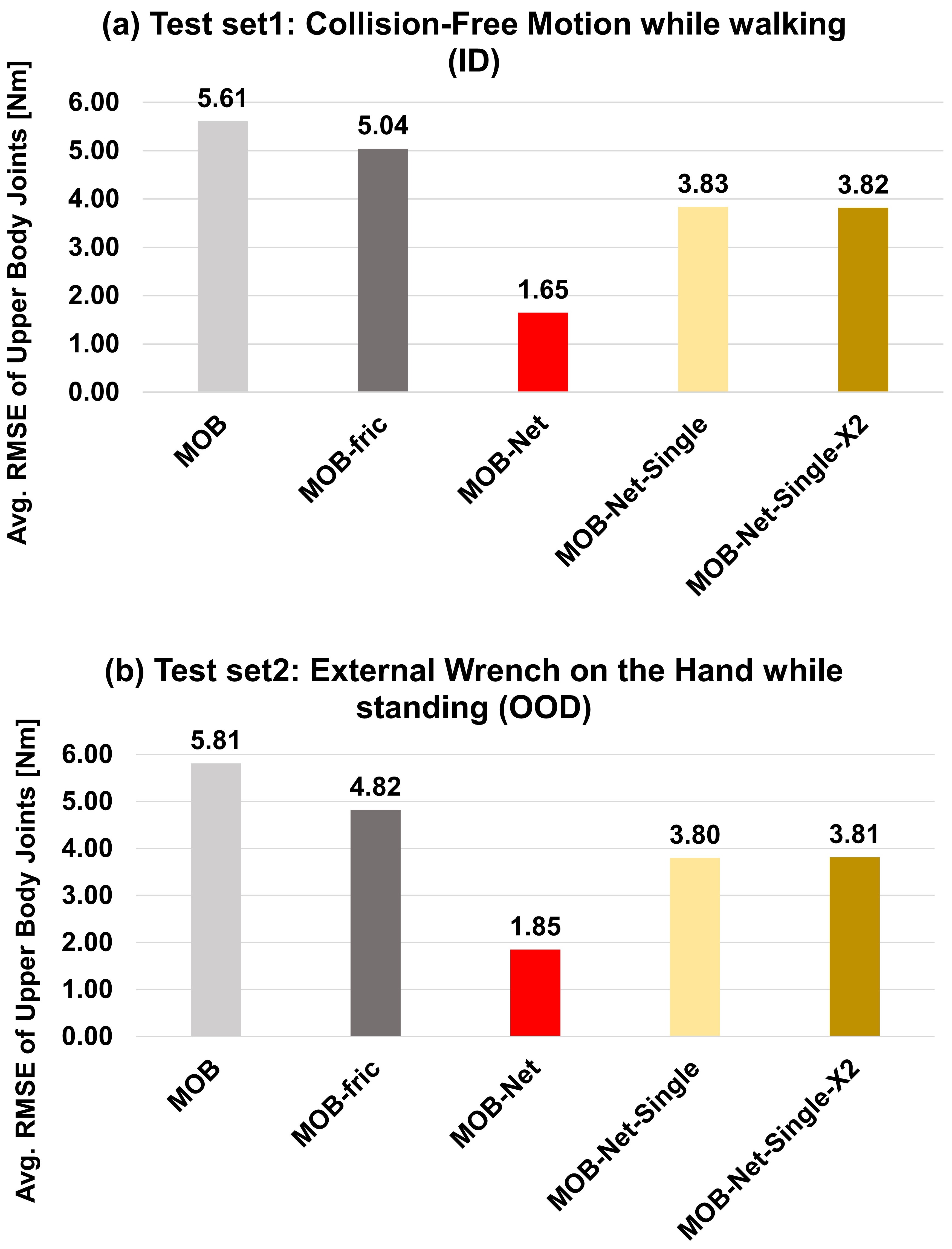}
\caption{\textcolor{black}{\textbf{Average RMSE of upper body joints for (a) the random motion test (ID) and (b) the random contract wrench test on the right hand (ODD) according to the different methods.} MOB-Net-Single shows larger errors than MOB-Net.}}
\label{fig/Ablation_single_network}
\end{figure}

The model size of each network is constrained to be similar to the one of MOB-Net to give a fair comparison of the architecture of each network except MLP-500. 
The average training and validation losses of five runs for each network are compared in Figure~\ref{fig/abliation_networks}. 
The results show that GRU has the lowest loss compared to other networks. LSTM shows a loss lower than that of the other comparison networks but higher than that of GRU. The vanilla RNN network shows poor performance compared to the other RNN-series networks (GRU and LSTM) and similar performance with the simple 2-layer MLP network which shows the worst performance. TCN ranked third showing better performance than the other simple architectures. When the model size increases, the performance gain is observed not only for MLP-500 but also for other networks. However, the gain in estimation accuracy was insignificant compared to the increased calculation time. Moreover, the network size can not be increased beyond its current size due to the computation time limit of real-time calculation at 1000hz. 
Therefore, the model size is compromised considering the learning result and the limited calculation time on the embedded computer of the robot.
We conjecture that the reason for the better performance of GRU over LSTM is that the uncertainty torque learning only requires a short history of the joint state whereas LSTM has an advantage in the task requiring long-term memory such as the natural language process.
Therefore, the GRU is efficient and suitable for uncertainty torque learning with constraints of network size.
\textcolor{black}{However, it is not always the case that GRU modules outperform LSTM.}

\textcolor{black}{We also validate the effectiveness of the limb-modularized structure of MOB-Net. A naive single network of MOB-Net that learns the uncertainty torque of whole body joints using all the input vectors is compared with the limb-modularized version of MOB-Net. The single MOB-Net is called 
\emph{MOB-Net-Single} and the network size of MOB-Net-single is designed to be similar to the total size of six sub-networks of the original MOB-Net. For training MOB-Net-Single, the random motion data is used as the training data. Figure~\ref{fig/Ablation_single_network}~(a) and \ref{fig/Ablation_single_network}~(b) show the test results of upper body joints for random motion data (ID) and random contact wrench data (ODD). The test results show that MOB-Net has smaller estimation errors compared to MOB-Net-Single, which validates the effectiveness of the limb-modularized architecture of MOB-Net. Furthermore, the double size of MOB-Net-Single rarely improves the estimation performance as shown in MOB-Net-Single-2X results in Figure~\ref{fig/Ablation_single_network}~(a) and \ref{fig/Ablation_single_network}~(b).}

\subsection{Effectiveness of Random Torque Exploration}
\label{Subsection/Ablation Study/Effectiveness of Random Torque Exploration}

% RTE ablation results
\begin{table}[!ht]
\centering
\caption{Random walking test results of left leg according to the use of random torque exploration.}
\label{table/comparison_of_RTE}
\resizebox{\columnwidth}{!}{%
\begin{tabular}{@{}ccccccccc@{}}
\toprule
                                                                                     &                    & \multicolumn{6}{c}{\textbf{Left leg joint}}   & \textbf{}      \\ \cmidrule(l){3-9} 
\multirow{-2}{*}{\textbf{Method}} &
  \multirow{-2}{*}{\textbf{Statistics}} &
  \textbf{LL1} &
  \textbf{LL2} &
  \textbf{LL3} &
  \textbf{LL4} &
  \textbf{LL5} &
  \textbf{LL6} &
  \textbf{avg} \\ \midrule
 &
  \textbf{RMSE [\emph{Nm}]} &
  \cellcolor[HTML]{C0C0C0}0.908 &
  \cellcolor[HTML]{C0C0C0}6.266 &
  \cellcolor[HTML]{C0C0C0}4.191 &
  \cellcolor[HTML]{C0C0C0}2.731 &
  \cellcolor[HTML]{C0C0C0}0.917 &
  \cellcolor[HTML]{C0C0C0}0.598 &
  \cellcolor[HTML]{C0C0C0}\textbf{2.602} \\
\multirow{-2}{*}{\textbf{\begin{tabular}[c]{@{}c@{}}MOB-Net\\ W/ RTE\end{tabular}}}  & \textbf{rRMSE [\%]} & 3.57  & 5.47  & 3.09  & 1.34  & 0.92  & 1.16  & -              \\ \midrule
                                                                                     & \textbf{RMSE [\emph{Nm}]} & 1.150 & 9.134 & 7.265 & 4.704 & 1.418 & 1.311 & \textbf{4.164} \\
\multirow{-2}{*}{\textbf{\begin{tabular}[c]{@{}c@{}}MOB-Net\\ WO/ RTE\end{tabular}}} & \textbf{rRMSE [\%]} & 4.52  & 7.98  & 5.36  & 2.30  & 1.42  & 2.54  & -              \\ \bottomrule
\end{tabular}%
}
\end{table}

% Please add the following required packages to your document preamble:
% \usepackage{booktabs}
% \usepackage{graphicx}
% \usepackage[table,xcdraw]{xcolor}
% If you use beamer only pass "xcolor=table" option, i.e. \documentclass[xcolor=table]{beamer}
\begin{table}[!ht]
\centering
\caption{Swing knee collision test results of the LL3 joint of the left leg according to the use of random torque exploration.}
\label{table/comparison_of_RTE_2}
\resizebox{0.85\columnwidth}{!}{%
\begin{tabular}{@{}cccccc@{}}
\toprule
\textbf{} &
  \multicolumn{5}{c}{\textbf{RMSE of LL3 joint [\emph{Nm}]}} \\ \midrule
\textbf{Method} &
  \textbf{Collsion 1} &
  \textbf{Collsion 2} &
  \textbf{Collsion 3} &
  \textbf{Collsion 4} &
  \textbf{avg} \\ \midrule
\textbf{\begin{tabular}[c]{@{}c@{}}MOB-Net\\ W/ RTE\end{tabular}} & 5.748
 & 5.767 & 7.975 & 6.813 & \cellcolor[HTML]{C0C0C0}\textbf{6.576} \\ \midrule
\textbf{\begin{tabular}[c]{@{}c@{}}MOB-Net\\ WO/ RTE\end{tabular}} &
  13.293 &
  12.039 &
  15.173 &
  13.218 &
  \textbf{13.431} \\ \bottomrule
\end{tabular}%
}
\end{table}

The random torque exploration (RTE) method is validated in this study.
During the data collection process, additional random torques are added to the robot to obtain widely distributed training data.
To validate the effect of RTE on the training results, we prepared two training data sets with and without RTE.
Table~\ref{table/comparison_of_RTE} represents the test results according to the use of RTE.
Not applying RTE in data collection for all joints increases estimation errors.
On average, the estimation error with RTE is less than the result without RTE by 1.562\,Nm.
We also tested the trained network without RTE for the unseen data.
Table~\ref{table/comparison_of_RTE_2} represents the RMSE of the LL3 joint for the four knee collisions during the swing phase that is the same experiment in \ref{Subsection/Experiment/Comparison of External Torque Estimation Performance in Real Humanoid Robot}. 
The results show that the RMSE increases to 13.431\,Nm when the RTE is not used during data collection.
The error without RTE is 6.855\,Nm larger than the RMSE with RTE, which means that RTE is effective for both in-distribution and out-of-distribution tasks.

\begin{figure}[!ht]
\centering
\vspace*{0.0cm}
\includegraphics[width=1.0\linewidth]{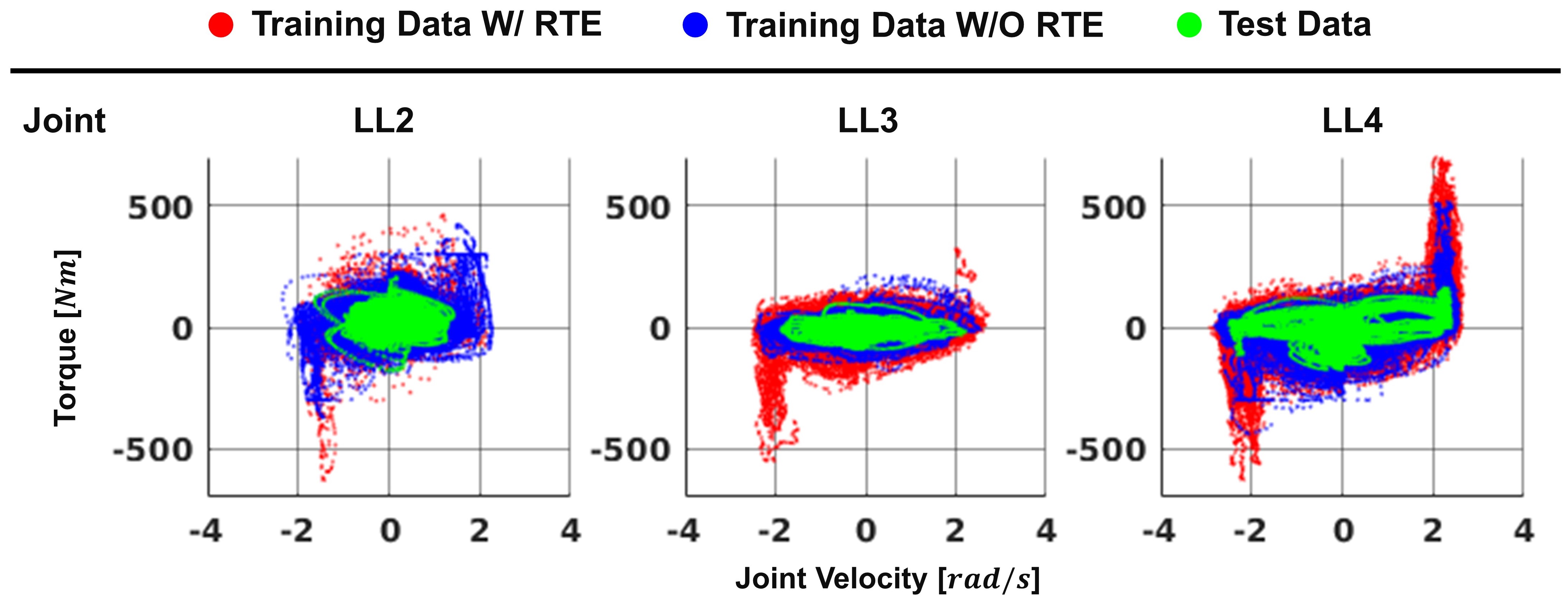}
\caption{\textbf{Joint velocity-torque plots of training data and test data.} Only three joints of the left leg are displayed.}
\label{fig/TrainingData_distirbution_RTE}
\end{figure}

To visualize the data distribution according to the use of RTE, the joint velocity-torque graphs for three joints of the left leg are displayed in Figure~\ref{fig/TrainingData_distirbution_RTE}.
Because it is difficult to visualize the entire multi-dimensional input vectors, only a subset of input features (joint velocity and torque) are displayed. 
As shown in Figure~\ref{fig/TrainingData_distirbution_RTE}, the use of RTE expands the distribution of training data, which in turn results in better estimation performance.

\subsection{Effectiveness of Network Size and Training Data Size}
\label{Subsection/Ablation Study/Effectiveness of Network Size and Training Data Size}

\begin{figure}[!ht]
\centering
\vspace*{0.0cm}
\includegraphics[width=1.0\linewidth]{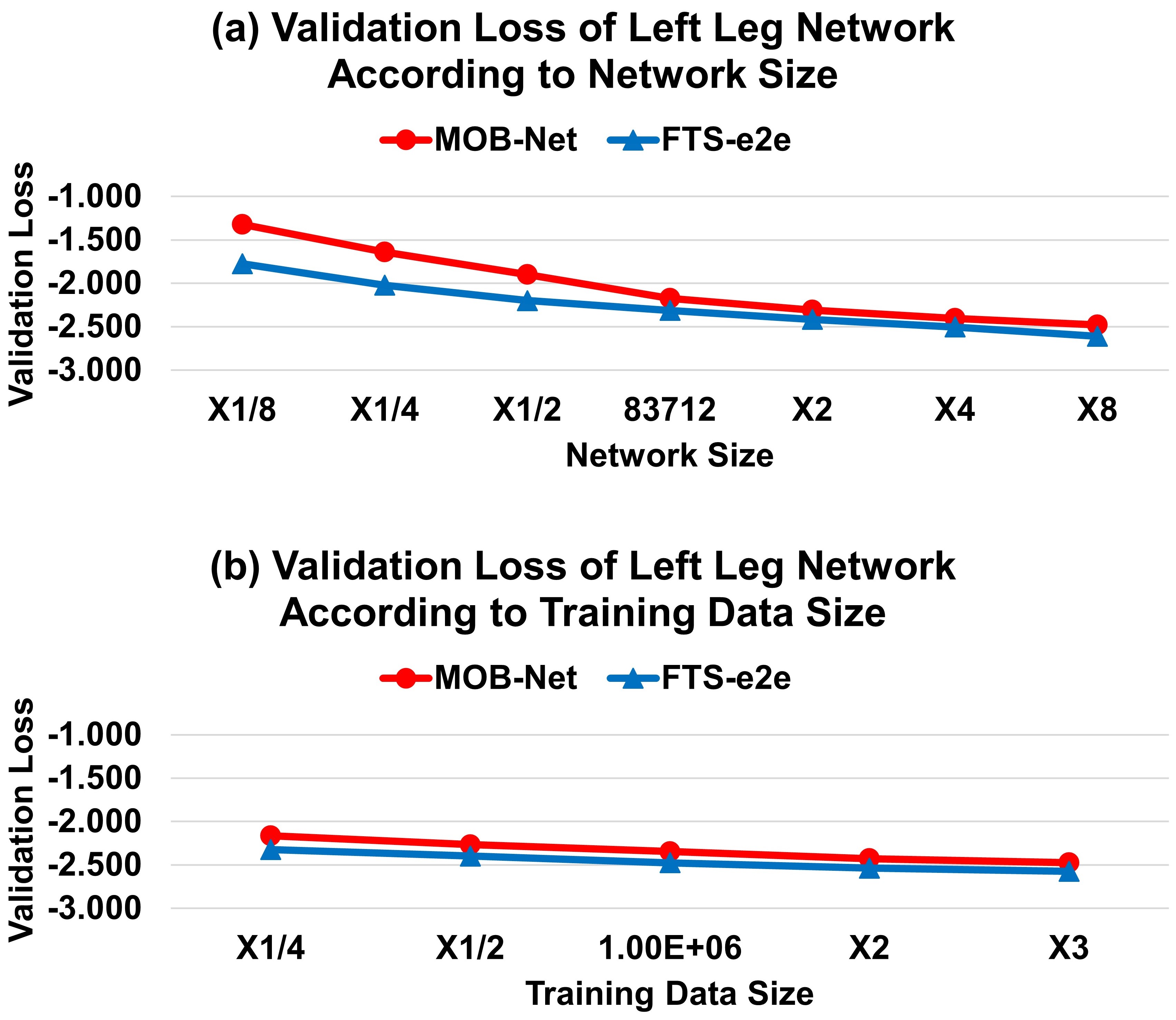}
\caption{\textcolor{black}{\textbf{Validation loss of MOB-Net and FTS-e2e according to the (a) network size and (b) training data size.}}}
\label{fig/Ablation_network_data_size}
\end{figure}

\textcolor{black}{
The ablation studies of MOB-Net and FTS-e2e for the network size and training data size are provided in this section.
To observe the effects of network size and training data size on the training results, the left leg networks are only trained and the other hyperparameters are kept the same as the original version shown in Table~\ref{Table/Network Architecture}.
For the network size, the network sizes are varied from an 8 times smaller model to an 8 times larger one.
As shown in Figure~\ref{fig/Ablation_network_data_size}~(a), the validation loss of both networks decreases as the network size increases.
However, as the network size increases, the computation time of inference also increases.
Therefore, the network size should be compromised considering the computation budget and the training performance. \textcolor{black}{For our robot with 37 DoF, the network size is determined to realize the real-time calculation of 1\,ms on an onboard computer (Intel Core i7-10700K).}
}

\textcolor{black}{
For the size of the training data, we make the training data by randomly sampling the data from the existing random walking training data (a total of 3.6M samples). Note that the training performance between the data collection of only 1M data and using 1M random samples from the 3.6M data would be different because the latter has more distributed training data. Although it is ambiguous how the subset of the training data should be selected for a fair comparison, the random sampling method is adopted.
As shown in Figure~\ref{fig/Ablation_network_data_size}~(b), the validation loss of both networks decreases as the number of training data samples increases.
However, the margin of training performance can be regarded as minor considering the increased training time with a larger data set.
Therefore, it is important to balance between the training performance and the training time. 
}

%%%%%%%%%%%%%%%%%%%%%%%%% APPLICATION %%%%%%%%%%%%%%%%%%%%%%%%%%%%%%%%%
\section{Application: Sensorless Locomotion, Collision Detection, and Collision Reaction Scenarios}
\label{Section/Application}
\begin{figure*}[!t]
\centering
\vspace*{0.0cm}
\includegraphics[width=1.0\linewidth]{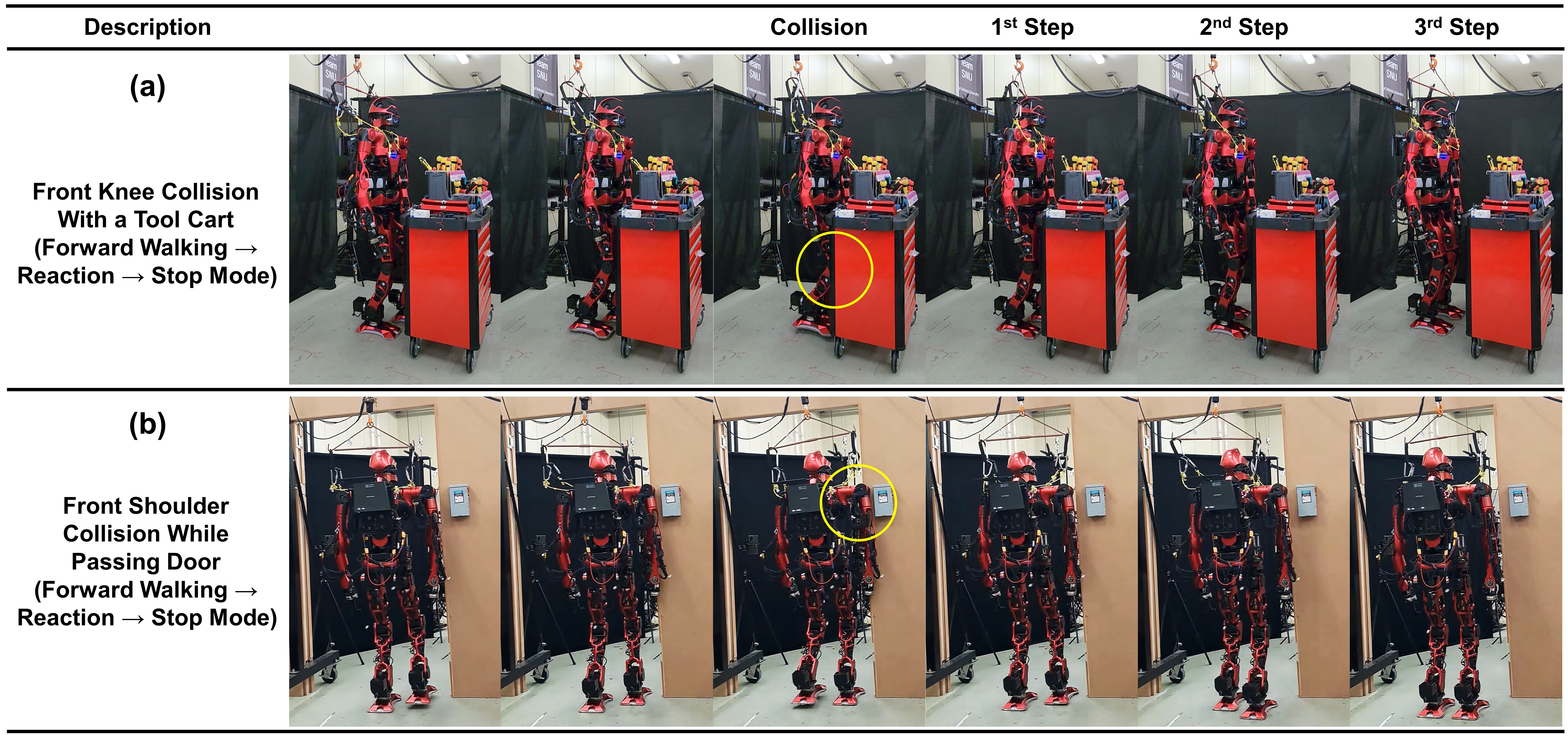}
\caption{\textbf{Collision detection and reaction scenarios of humanoids using the estimated external torque of MOB-Net (see also \href{https://youtu.be/ZpnMEjvGsaQ}{Extension 2}).} (a) The front knee of the humanoid collides with a tool cart fixed on the ground while walking forward. (b) The right shoulder of the humanoid robot collides with a door while passing the door.}
\label{fig/collision_detection_to_reaction_scenarios}
\end{figure*}

\textcolor{black}{In this section, our heuristic collision-handling strategy using MOB-Net is introduced.}
Furthermore, we present two realistic collision handling scenarios as shown in Figure~\ref{fig/collision_detection_to_reaction_scenarios} (\href{https://youtu.be/ZpnMEjvGsaQ}{Extension 2}). Three additional scenarios can be found in \href{https://youtu.be/ZpnMEjvGsaQ}{Extension 2}.
These scenarios present the practical application and highlight the robust estimation performance of the proposed method for various collision cases.
Note that, in these scenarios, only proprioceptive sensors of the robot (encoders and IMU) are used for walking control and collision handling. The walking controller using the estimated external joint torque is the same as in \cite{lim2023proprioceptive}.

\textcolor{black}{
The implemented collision-handling pipeline for humanoid robots consists of several steps as defined in \cite{haddadin2017robot}: collision detection, isolation, identification, classification, and reaction. 
\begin{itemize}
    \item Collision Detection: The detection method utilizes the estimated external torque and standard deviation signals from MOB-Net. It employs thresholding and signal filtering techniques to effectively monitor and detect collisions.
    \item Collision Isolation: This stage simplifies the process by adopting a lumped limb-wise approach for collision isolation. It does not require the estimation of exact contact points, thereby simplifying the process.
    \item Collision Identification: Here, the strategy leverages the estimated external torques and contact Jacobian matrices. It operates under the assumption that intentional contacts are confined to designated end-effectors, specifically the robot's feet and hands.
    \item Collision Classification: The classification process distinguishes between intentional and unintentional contacts. It employs criteria based on the robot's walking planner, estimated external wrenches, and friction cones to make this distinction.
\end{itemize}
Each of these stages collectively forms a robust framework, which is integral to the development of a sophisticated collision reaction strategy. This strategy is pivotal in enhancing the safety and effectiveness of robotic operations.
}

\begin{figure}[!ht]
\centering
\vspace*{0.0cm}
\includegraphics[width=1.0\linewidth]{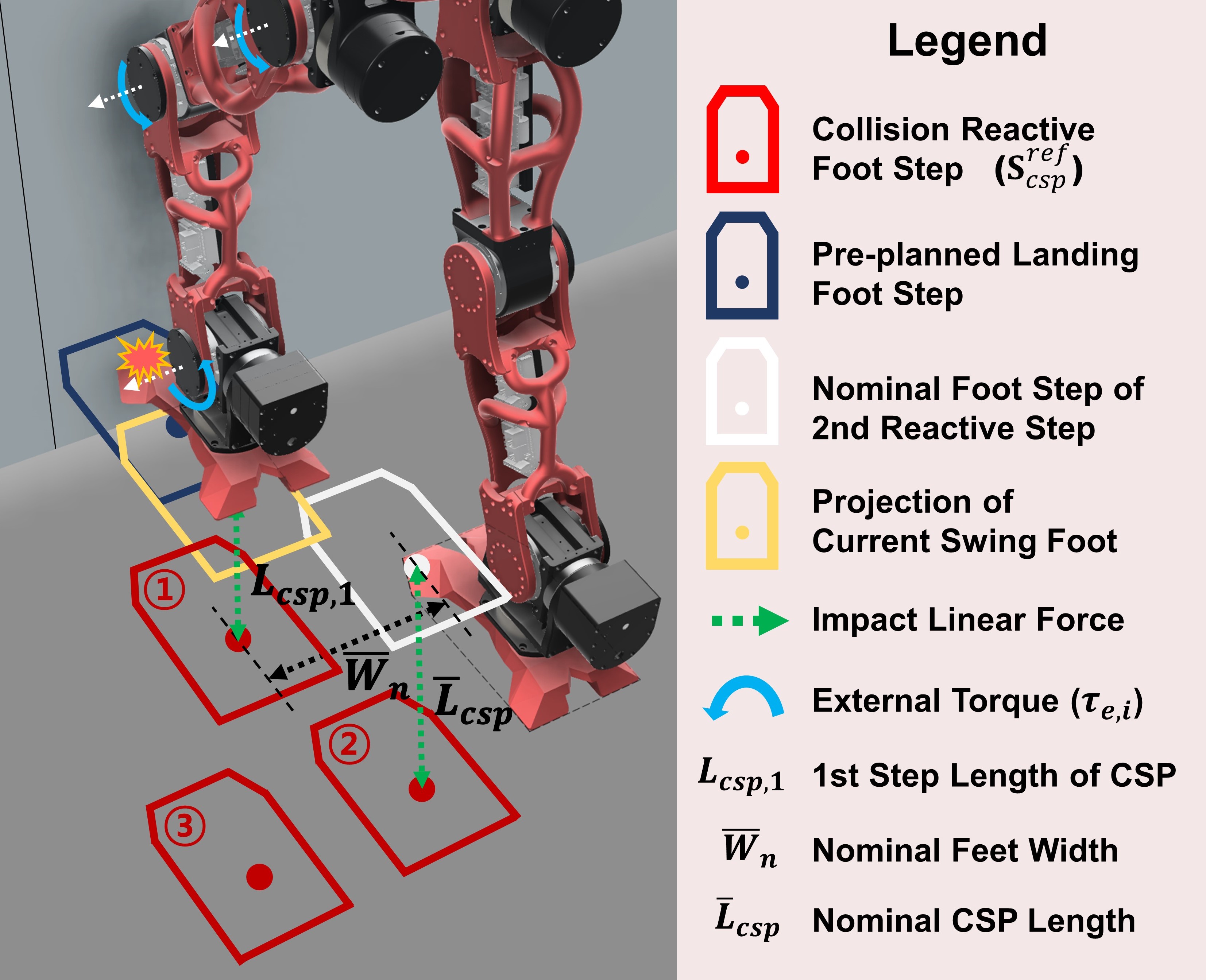}
\caption{\textcolor{black}{\textbf{An illustration of the proposed Compliant Stepping Planning (CSP) method.} When the Humanoid's swing foot collides with the wall while walking forward, CSP generates three reactive footsteps along with the impact force observed in the pelvis frame.}}
\label{fig/CSP}
\end{figure}

\textcolor{black}{
In response to unexpected collisions, a collision reaction strategy for humanoid robots is developed, emphasizing the need for appropriate responses to prevent falls and damage. This strategy includes Compliant Stepping Planning (CSP) and Rewind Stepping Planning (RSP), tailored for different environmental contexts. The CSP, illustrated in Figure~\ref{fig/CSP}, generates reactive footsteps $\bm{S}^{ref}_{csp,i} \in SE(3)$ along the direction of the ground-projected unexpected external wrench on the pelvis link, $\hat{\bm{F}}_{u}^{fb}\in \mathbb{R}^{6}$ to avoid further collision with the unknown obstacle and to achieve compliant behavior.
The unexpected collision wrench of the pelvis link, $\hat{\bm{F}}_{u}^{fb}$, is calculated by subtracting the anticipated contact wrench of the supporting foot or manipulating hands from the estimated external wrench of the pelvis frame as 
\begin{equation}
    \hat{\bm{F}}_{u}^{fb} = \hat{\bm{F}}_{e}^{fb} - \sum_{l\in L}^{n_{cl}} {}^{fb}_{l}\bm{X} \hat{\bm{F}}_{e}^{l}, \ L = \{RL, LL, RA, LA\}
\end{equation}
where $L$ is the limb index in which contacts can be expected, $n_{cl}$ is the number of limbs where the expected contact occurs, and ${}^{A}_{B}\bm{X} \in \mathbb{R}^{6 \times 6}$ is the coordinate transformation matrix from the coordinate frame \emph{B} to the coordinate frame \emph{A}.
Also, the first reactive step length, $L_{csp,1}$, is determined according to the height of the swing foot and the remaining step time to prevent fast swing foot motion.
In response to collisions in confined spaces, the RSP strategy is employed, which safely retraces the robot's previous steps stored in a footstep buffer, adapting the next steps based on the current swing foot phase. The fundamental idea under RSP is that following the footsteps that the robot has stepped before is safe because the robot already passed without any accident. Thanks to our robust walking control framework, sudden changes in the future footsteps can be accomplished stably.
}

These scenarios exemplify typical and representative collision situations that are likely to happen in human-centered industrial sites. 
In Figure~\ref{fig/collision_detection_to_reaction_scenarios}~(a), the humanoid robot walks forward and its right knee collides with a fixed tool cart. Then, the collision is detected in the hip-pitch joint of the right leg and the robot reacts to the collision by walking three steps along the external force direction (backward) to avoid additional collisions with an obstacle. After three steps backward, the robot entered the stop mode safely. \textcolor{black}{In this case, CSP is utilized for reaction.} 
Figure~\ref{fig/collision_detection_to_reaction_scenarios}~(b) demonstrates a door collision scenario. In this scenario, the robot moved forward to get through the door, but the right shoulder of the robot collided with the door frame. The proposed method can also detect such collision on the upper body and enable the robot to react to the collision by returning to the way it passed, \textcolor{black}{i.e., RSP.} This collision reaction prevents the robot from making further collisions with the door and falling. Note that these collisions can not be detected using FTS on the end-effectors of the humanoid.

\section{Discussion}
\label{Section/Discussion}
\subsection{\textcolor{black}{Requirement of FTS for Data Collection}}
\textcolor{black}{Although MOB-Net only requires encoder and IMU measurements for network inference, the use of FTS on the feet is essential during the random walking data collection phase for legs to calculate the external joint torque. This calculation is crucial for determining the uncertainty torque on the legs and virtual joints in the walking data. For research groups possessing only a few humanoid robots, this requirement for FTS during data collection could diminish the anticipated cost reduction benefits. However, it's important to highlight that MOB-Net extends the collision detection capability beyond the foot link to other links of the humanoid robot, a functionality that is challenging to achieve with FTS solely on the foot. Furthermore, we believe that even though FTS is necessary for data collection, a small number of these sensors can be used to generate data for multiple humanoid robots. Consequently, in mass production scenarios, the proposed method can still lead to significant cost reductions in manufacturing humanoid robots. However, the specific manufacturing process and learning process using FTS in large-scale production would necessitate further research. 
}

\subsection{\textcolor{black}{Quality of Training Data and Generalization for Unseen Data}}
\textcolor{black}{
Another limitation of the proposed method is that the robust estimation performance for the unseen data is not theoretically guaranteed but validated experimentally, and the estimation performance is closely related to the quality of the training data.
MOB-Net, unlike FTS-e2e, employs both MOB and GRU networks. This architecture enables collision detection even when network estimation errors increase due to unseen collisions, as MOB can detect the change of external torques in its residual signals. MOB-Net also experiences increased external joint torque estimation errors for unseen collisions compared to the in-distribution data. However, these errors are significantly lesser than those in FTS-e2e, making it reliable for collision detection. Additionally, the standard deviation of uncertainty torque increases during collisions, which is utilized to enhance collision detection performance.
However, this limitation is still an open problem in the field of deep learning research.}

\textcolor{black}{
The unseen data tested in this thesis deviates from the distribution of the training data in the dimension of torque while the robot moves in the distribution of the training data in the dimension of joint position and velocity.
Thus, the robust estimation performance for the unseen data in the dimension of joint position and velocity is not validated, and it is also not guaranteed that the proposed data collection method can cover all the possible motions of the robot although many efforts are devoted to collect diverse and representative data such as RTE and random walking commands on uneven terrain.
Nonetheless, the training data covers almost all possible motions of the robot in the mechanical limitations, and if the robot is operated for unseen motions when the motion planning algorithm or control method is changed, fine-tuning with a small amount of additional data as introduced in \cite{kim2021transferable} could solve the problem.
}

\subsection{\textcolor{black}{Contact Point Localization}}
\textcolor{black}{In this study, rather than estimating the precise contact location, we focus on determining which limb has been impacted by a collision. Contact point localization methods have been proposed in \cite{manuelli2016localizing, popov2021real, liang2021contact}. The accurate external torque estimated by MOB-Net can be applied to these existing contact location estimation methods. However, there are anticipated challenges. Firstly, real-time computation may become difficult when applying these methods to humanoid robots with many DoF and a complex surface mesh topology. Secondly, even when utilizing MOB-Net, the accuracy of the estimated external torque is expected to be lower compared to collaborative robots using JTS, such as KUKA iiwa and Franka Panda, which may result in lower accuracy of contact point estimation.}

\subsection{\textcolor{black}{Multi Contact Identification and Rank Deficit of Contact Jacobian}}
\textcolor{black}{MOB-Net, similar to conventional momentum observers (MOB) and other disturbance observers, observes disturbances in the generalized coordinate, i.e., joint space. External wrenches acting on the robot are propagated to the generalized coordinates via the transpose of the contact Jacobian, $\bm{J}^{T}_{c}\bm{F}_{e}$. During this process, depending on the rank of the Jacobian, information loss can occur, potentially preventing the restoration of the actual external wrenches. Specifically, under the assumption that the external force is a 3-dimensional linear force with an unknown contact point, the information that must be restored for each external force includes the 3D linear force vector and the 3D position of the contact, totaling 6 ranks.}

\textcolor{black}{
For instance, if no external forces act upon the upper body of a robot while the robot is in the single support phase with a left leg, the right swing foot's Jacobian, including the virtual joint, possesses 12 ranks. Therefore, it is possible to estimate and restore up to two external forces—one on the swing foot link and another on a different right leg link. However, it is impossible to perfectly identify more than 3 multiple contacts on the swing leg due to the rank deficit of the stack of contact Jacobian. The theoretical analysis and simulation results related to multi-contact detection depending on the Jacobian rank are well presented in \cite{vorndamme2017collision}.}

\section{Conclusion}
\label{Section/Conclusion}
In this paper, the limb-modularized uncertainty torque learning method, MOB-Net, has been presented. By utilizing the sparsity of the dynamics of the humanoid robot as an inductive bias, an efficient and effective network architecture was designed. The integration of a model-based method (MOB) with deep learning techniques leads to improved accuracy in estimating normal walking data (in-distribution data) as well as robust estimation for unseen collision data.
Furthermore, the estimated standard deviation of the uncertainty torque from the Gaussian output feature contributes to the fast and robust collision detection performance.
Extensive simulations and experiments were conducted to compare the proposed method with end-to-end learning and other model-based observers, demonstrating superior estimation performance. 
\textcolor{black}{Finally, the sensorless collision handling method is implemented and various collision handling scenarios are presented using the real humanoid robot to show the versatility of MOB-Net.} This is made possible due to MOB-Net, the core technology: accurate external joint torque estimation from MOB-Net in all joints including virtual joints, and sensitive collision detection using MOB-Net's output. 
We conclude that MOB-Net can enhance the safety of humanoids by implementing the whole-body collision handling method along with the balancing controller. It can also reduce the manufacturing cost of humanoids by replacing FTS with MOB-Net in mass production although FTS on the foot is required for data collection.

\subsection{\textcolor{black}{Future Works}}
\label{Subsection/conclusion/Limitation and Future Work}
One possible direction for future work is to use the estimated uncertainty torque and the external joint torque to improve control performance or enable compliant force control in humanoid robots. Similar to the work for aerial vehicles in \cite{o2022neural}, the learned disturbances can be incorporated into the controllers and improve the control performance.
As another possible direction of future work, the estimated uncertainty torque from MOB-Net can be used to narrow the sim-to-real gap of deep reinforcement learning. As in the study of \cite{hwangbo2019learning}, the learned uncertainty torque from the real robot could be utilized to simulate the virtual robot similar to the real robot.
\textcolor{black}{Finally, the contact point localization can be conducted using the estimated external joint torques from the MOB-Net and previous methods in \cite{manuelli2016localizing, popov2021real, liang2021contact}.}

\begin{acks}
The authors would like to thank JuneWhee Ahn and Kwanwoo Lee for their gracious help in the real robot experiments.
\end{acks}

\begin{dci}
The author(s) declared no potential conflicts of interest with respect to the research, authorship, and/or publication of this article.
\end{dci}

\begin{funding}
The author(s) disclosed receipt of the following financial support for the research, authorship, and/or publication of this article: This work was supported by the National Research Foundation of Korea (NRF) grant funded by the Korea government (MSIT) (No. 2021R1A2C3005914).
\end{funding}

% \theendnotes

\bibliographystyle{Bibliography/SageH}
\bibliography{Bibliography/IJRR_MOB_Net}

\section*{Appendix}
\label{Section/Appendix A}

\subsection*{A. Index to Multimedia Extensions}
\begin{table}[!ht]
\centering
\caption{}
\label{table/multimedia}
\resizebox{\columnwidth}{!}{%
\begin{tabular}{@{}ccl@{}}
\toprule
Extension & Media Type & Description                                                                                                                                                            \\ \midrule
1         & Video      & \begin{tabular}[c]{@{}l@{}l@{}}Training data procedure including random\\walking data, random motion data, and\\random torque exploration.\end{tabular} \\ \midrule
2         & Video      & \begin{tabular}[c]{@{}l@{}}Five different collision detection and reaction\\scenarios of a humanoid robot, TOCABI.\end{tabular}                    \\ \bottomrule
\end{tabular}%
}
\end{table}

\subsection*{B. Threshold Values for Collision Detection}
\label{Section/Appendix B}
\begin{table}[!t]
\centering
\caption{Threshold Values of Left Leg Joints for Collision Detection According to Estimation Method}
\label{table/collision detection threshold}
\resizebox{\columnwidth}{!}{%
\begin{tabular}{@{}ccccccc@{}}
\toprule
\multirow{2.5}{*}{\textbf{Method}} & \multicolumn{6}{c}{\textbf{Left leg joint [$Nm$]}}                                             \\ \cmidrule(l){2-7} 
                                 & \textbf{LL1} & \textbf{LL2} & \textbf{LL3} & \textbf{LL4} & \textbf{LL5} & \textbf{LL6} \\ \midrule
\textbf{MOB-Net-mean}            & 2.42         & 49.72        & 19.69        & 17.27        & 1.65         & 4.84         \\
\textbf{MOB-Net-sigma}           & 0.99         & 3.41         & 4.18         & 2.20         & 0.99         & 0.44         \\
\textbf{FTS-e2e-mean}            & 3.74         & 35.53        & 13.20        & 18.15        & 4.73         & 5.28         \\
\textbf{FTS-e2e-sigma}           & 1.65         & 8.69         & 10.45        & 3.96         & 2.09         & 0.77         \\
\textbf{MOB}                     & 19.25        & 60.28        & 114.62       & 111.98       & 38.94        & 15.62        \\
\textbf{MOB-fric}                & 15.95        & 84.26        & 88.33        & 81.29        & 26.62        & 15.40        \\
\textbf{MOB-fric-BPF}            & 31.00        & 97.80        & 65.80        & 68.40        & 38.90        & 25.00        \\ \bottomrule
\end{tabular}%
}
\end{table}
                                  
\end{document}